\documentclass[sn-nature]{sn-jnl}%

\usepackage{tabularx}
\usepackage{graphicx}%
\usepackage{multirow}%
\usepackage{amsmath,amssymb,amsfonts}%
\usepackage{amsthm}%
\usepackage{mathrsfs}%
\usepackage[title]{appendix}%
\usepackage{xcolor}%
\usepackage{textcomp}%
\usepackage{manyfoot}%
\usepackage{booktabs}%
\usepackage{algorithm}%
\usepackage{algorithmicx}%
\usepackage{algpseudocode}%
\usepackage{listings}%
\usepackage{subcaption}%
\usepackage{bibunits}
\defaultbibliography{references,lit}

\usepackage{titletoc}

\theoremstyle{thmstyleone}%

\theoremstyle{thmstyletwo}%

\theoremstyle{thmstylethree}%

\usepackage[colorinlistoftodos,linecolor=blue,backgroundcolor=white,bordercolor=blue,textsize=tiny,textwidth=28mm]
{todonotes}
\newcounter{task}

\newcounter{fs}
\newcounter{leo}
\newcounter{ig}
\newcounter{rc}
\newcounter{zy}

\newcommand{\rev}[1]{\textcolor{black}{#1}}

\newcommand{\SI}{Supplementary Information section}
\newcommand{\si}{Supplementary Information}

\usepackage{xspace}
\newcommand{\ourname}{GenFocal\xspace}

\begin{document}
\begin{bibunit}

\title[Title]{Regional climate risk assessment from climate models using probabilistic machine learning}

\author[1]{\fnm{Zhong Yi} \sur{Wan}}\email{wanzy@google.com}
\equalcont{These authors contributed equally to this work.}

\author[1]{\fnm{Ignacio} \sur{Lopez-Gomez}}\email{ilopezgp@google.com}
\equalcont{These authors contributed equally to this work.}

\author[1]{\fnm{Robert} \sur{Carver}}\email{carver@google.com}

\author[1,2]{\fnm{Tapio} \sur{Schneider}}\email{tapio@google.com}

\author[1,3]{\fnm{John} \sur{Anderson}}\email{janders@alum.mit.edu}

\author*[1,4]{\fnm{Fei} \sur{Sha}}\email{feisha@meta.com}

\author*[1]{\fnm{Leonardo} \sur{Zepeda-N\'u\~nez}}\email{lzepedanunez@google.com}

\affil[1]{\orgname{Google Research}, \orgaddress{\city{Mountain View}, \state{CA}, \country{USA}}}

\affil[2]{\orgname{California Institute of Technology}, \orgaddress{\city{Pasadena}, \state{CA}, \country{USA}}}
\affil[3]{\orgname{General Motors Sunnyvale Tech Center}, \orgaddress{\city{Sunnyvale}, \state{CA}, \country{USA}}}
\affil[4]{\orgname{Meta}, \orgaddress{\city{Menlo Park}, \state{CA}, \country{USA}}}

\keywords{climate downscaling, risk assessment, generative modeling}

\maketitle
\setcounter{page}{1}

\section*{Abstract}

Effective climate risk assessment is hindered by the resolution gap between coarse global climate models and the fine-scale information needed for regional decisions. We introduce GenFocal, an AI framework that generates statistically accurate, fine-scale weather from coarse climate projections, without requiring paired simulated and observed events during training. GenFocal synthesizes complex and long-lived hazards, such as heat waves and tropical cyclones, even when they are not well represented in the coarse climate projections. It also samples high-impact, rare events more accurately than leading methods. By translating large-scale climate projections into actionable, localized information, GenFocal provides a powerful new paradigm to improve climate adaptation and resilience strategies.

\section*{Main}
\noindent

Effective economic and societal planning over horizons from years to decades is increasingly hinging on understanding and predicting how climate changes. While global climate models (GCMs) provide robust projections of large-scale trends, their coarse resolution creates a critical information gap for local and regional decision-making~\cite{Voosen2025Local}. This gap limits risk assessments for infrastructure design~\cite{pmp}, energy system planning~\cite{Qiu2024}, flood forecasting~\cite{Nearing2024}, and financial services such as insurance pricing \cite{Mills2005}. The challenge is particularly acute when assessing compound events, such as concurrent heat and drought, where complex spatiotemporal correlations are not resolved by GCMs, yet drive the most severe impacts~\cite{Zscheischler2018,Bevacqua23a,Gettelman2025}. Bridging this scale gap is a grand computational and scientific challenge, and the primary obstacle to translating climate science into actionable intelligence for climate adaptation and resilience strategies.

This global-to-regional translation is fundamentally a problem of characterizing the statistics of fine-scale physical processes from coarse-grained climate projections. The large computational cost of GCMs restricts them to coarse grids, which inherently limits their fidelity and introduces biases. Furthermore, the chaotic nature of the climate system means that directly aligning coarse climate simulations with local weather observations at fine temporal scales is impossible.
Historically, this lack of alignment has seriously hampered the development of a general statistical mapping from GCM outputs to consistent local weather that reliably preserves the complex correlation structures needed for accurate regional climate risk assessment. In particular, this consistency with local weather is essential to near-term climate risk assessment where the statistical characterization must be grounded in past observations~\cite{Gettelman2025}. 

Existing efforts to overcome this downscaling challenge aim to correct biases and add coherent, fine-scale detail to coarse GCM outputs. Physics-based dynamical downscaling uses regional climate models (RCMs) forced by GCMs to simulate climate processes at higher resolution. They capture the rich spatiotemporal correlations of climate processes, but their computational expense limits the uncertainty quantification and robust assessment of extreme events, which requires a large number of samples~\cite{Giorgi2019, Goldenson2023}.
Statistical downscaling methods are a computationally efficient alternative, but they often lack the flexibility to capture the full range of spatiotemporal correlations that define complex weather and compound events~\cite{chandel2024state}.
Recent advances in machine learning (ML) have spurred a new generation of ML-based downscaling methods which have shown great promise in weather forecasting and regional climate model emulation~\cite{mardani2023generative,lockwood2024generative,rampal2024enhancing}. However, these methods predominantly rely on a supervised learning paradigm that requires temporally-aligned pairs of low- and high-resolution data for training. This requirement poses a significant challenge for downscaling climate projections, for which corresponding high-resolution ground truth data is unavailable.

We propose \ourname, a computationally efficient, general-purpose generative AI framework that addresses this grand challenge.
\ourname rapidly generates large ensembles of weather realizations that are physically realistic and statistically consistent with conditioning coarse climate projections. It is capable of elucidating complex, fine-scale weather phenomena from coarse climate data sources where those phenomena are poorly resolved. 

\ourname yields higher-fidelity local climate information than well-established statistical methods, including those used in leading reports such as the U.S. Fifth National Climate Assessment. By modeling spatial and temporal correlations, \ourname is able to capture the risk and evolution of spatiotemporal extremes—such as heatwaves and tropical cyclones (TCs)—in a unified framework,
a capability that is well beyond current statistical models.

\section*{\ourname}
\label{sec:genfocal}

\begin{figure*}[t]
\centering
\includegraphics[width=0.9\textwidth]{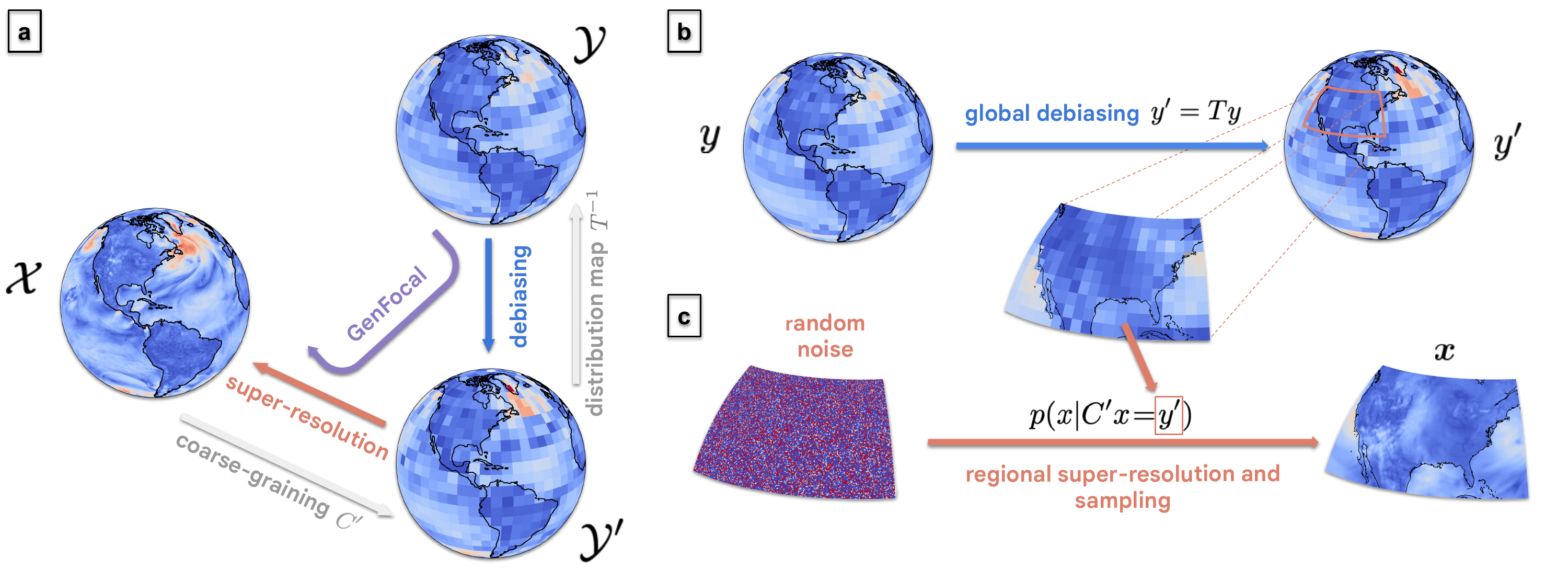}
  \caption{ 
  \textbf{\ourname is an end-to-end generative AI framework for climate downscaling, transforming coarse climate projections into actionable, fine-scale information for local regions, thereby enabling climate risk assessment and adaptation.} 
  \textbf{a}. Two-stage process. First, a coarse climate simulation from the space $\mathcal{Y}$ is bias corrected into the low-resolution space $\mathcal{Y}'$ in the same \rev{spatio-temporal grid (daily-mean at $1.5^{\circ}$)}. A super-resolution step then increases the resolution from  $\mathcal{Y'}$ to the target weather-state space $\mathcal{X}$.  \textbf{b}. The debiasing operator $T$ is instantiated as a rectified flow~\cite{liu2023flow} to match the distributions of the global low-resolution climate and a latent low-resolution weather space. \textbf{c}. The super-resolution step, $p(x|y')$, uses a conditional diffusion model~\cite{song2020score} to statistically invert the coarse-graining map $C'$. This process adds fine-grained spatiotemporal details, increasing temporal resolution from daily to 2-hourly within a local patch. At inference, a domain decomposition technique ensures temporal consistency across long sequences (see \SI~\ref{si:multi-diffusion} and Fig.~\ref{fig:dfn_sampling_schematic}).}
  \label{fig:framework_diagram}
\end{figure*}

Figure~\ref{fig:framework_diagram} provides a schematic of the algorithmic pipeline for \ourname\!, highlighting its key innovations (see \SI~\ref{si:method} for complete details). \ourname has three main features: two distinct generative AI techniques for bias correction and super-resolution, and the explicit modeling of temporal sequences of consecutive weather states at both the coarse and fine scale levels. To the best of our knowledge, \ourname is the first climate model downscaler capable of probabilistically mapping coarse climate projections to fine-grained weather fields with strong temporal coherence, enabling the accurate characterization of complex, multi-day events  such as prolonged heat waves or tropical cyclones (TCs).

In this work, \ourname downscales coarse climate states ($\mathcal{Y}$ in Fig.~\ref{fig:framework_diagram}) from $1.5^{^{\circ}}$ resolution to fine-grained weather states ($\mathcal{X}$ \rev{in Fig.~\ref{fig:framework_diagram}}) at $0.25^{^{\circ}}$ resolution, \rev{using an intermediate latent space ($\mathcal{Y}'$ in Fig.~\ref{fig:framework_diagram}) of coarse-grained weather states matching the resolution of $\mathcal{Y}$} . The input features 10 daily-averaged variables, while the output contains 4 variables sampled 2-hourly (Table \ref{table:modeled_vars_debiasing}). \ourname is trained on 20 years (1980-1999) of data, using the publicly available ERA5 reanalysis~\cite{hersbach2020era5} as the high-resolution target and the Community Earth System Model Version 2 (CESM2) Large Ensemble (LENS2)~\cite{LENS2} as the coarse-resolution source. Hyperparameter tuning was performed using the period 2000-2009, and the final evaluation is reported for the 10-year period 2010-2019, downscaling the full LENS2 ensemble. This chronological split for training, validation, and testing aligns with the intended application of assessing climate risk over the next few decades~\cite{Gettelman2025}. Details can be found in \SI~\ref{si:data} (including the data used) and \SI~\ref{si:method} (including details on the frameworks, neural architectures, training, and hyper-parameter tuning).

Catalyzed by the recent emergence of generative AI models, a few recent studies have explored generative modeling in weather forecasting \cite{Price2025,Li2024} and downscaling~\cite{mardani2023generative}. We provide a detailed review of these modern ML techniques, alongside traditional methods, in \SI~\ref{si:related}. A key limitation of existing approaches is their reliance on temporally aligned source and target data, generally unavailable when bridging coarse climate simulations with historical weather records. \ourname is the first generative AI framework designed to overcome the scientific challenge of correlational modeling of chaotic systems by operating without this alignment assumption. It offers the scalability needed for real-world datasets and has undergone thorough methodological verification.

\section*{Realistic genesis and evolution of tropical cyclones}
\label{sec:tc}

\begin{figure}[t]
\centering
\includegraphics[width=\textwidth]{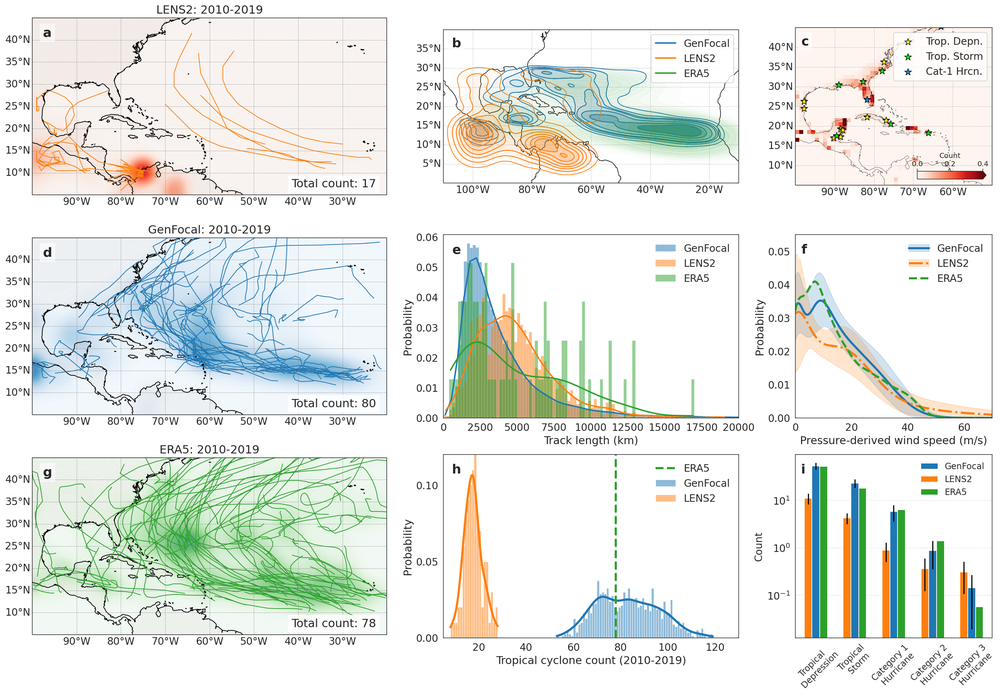}
\caption{\textbf{\ourname accurately reproduces the statistics of tropical cyclones in the North Atlantic in the time period 2010-2019, in terms of cyclogenesis, intensity and morphological features.} \textbf{a}, \textbf{d}. Ensemble track density and tracks for a single member from LENS2 and the downscaled high-resolution member generated by \ourname. \textbf{g}. Tracks and density map from the historical ERA5 reanalysis. \textbf{b}. Contours of cyclogenesis locations. \textbf{e}. Length of the tracks, characterizing their morphology. \textbf{c} Expected landfall count overlaid with ERA5 observations. \textbf{f}. Distribution of pressure-derived wind speed with 95\% confidence intervals. \textbf{h}, \textbf{i}. TC count and their Saffir-Simpson scale distributions. For LENS2 and \ourname, we use 100 and 800 members respectively to compute error bars and confidence intervals, shown in the plots.
}\label{fig:tc}
\end{figure}

Tropical cyclones (TCs) are exceptionally destructive natural hazards responsible for thousands of deaths and tens of billions of dollars in damages every year. The success of mitigation strategies depends heavily on reliable projections of TC frequency, intensity and tracks under different climate scenarios. 
High-fidelity simulation of fine-grained physical processes is necessary for driving TC genesis and evolution, requiring much higher resolutions than those afforded by current global climate models. 
Physics-based dynamical downscaling via RCMs can accurately capture the evolution of individual TCs, but it remains too expensive to generate the vast amount of data necessary to assess regional TC risk~\cite{Jing2021}. Studying future TC risk with statistical downscaling is possible, but only through bespoke methods that do not capture their interaction with the environment, and emulate TCs with reduced order systems that are partially coherent with their underlying physics~\cite{Jing2024}.

In contrast, \ourname is able to capture the full life cycle of TCs, from genesis to maturity (cf. Fig.~\ref{si:fig:tctrack} in \si), \emph{without} specifically targeting these emergent extreme phenomena in our model design and training. As shown in Fig.~\ref{fig:tc} for the North Atlantic basin, \ourname is able to generate TCs based on the input's large-scale conditions, even when these storms are largely absent from the input climate projections (see Methods and \SI~\ref{si:evaluation_protocol}). This ability crucially broadens \ourname\!'s applicability compared to methods reliant on input data at resolutions beyond those routinely available from climate models~\cite{Jing2024,lockwood2024generative}.

\ourname generates TCs with tracks (Fig.~\ref{fig:tc}b-c), cyclogenesis locations (Fig.~\ref{fig:tc}d), landfalls (Fig.~\ref{fig:tc}g), frequency (Fig.~\ref{fig:tc}f), intensity (Fig.~\ref{fig:tc}h,i), and morphology (Fig.~\ref{fig:tc}e, and Fig.~\ref{fig:si_sinuosity_idx_distribution} in \si) consistent with the ERA5 reanalysis and the target resolution~\cite{davis_2018} over the test period 2010-2019. This is in stark contrast with the statistics and tracks identified in the coarse LENS2, which exhibit both a lower frequency and excessively long durations (Fig.~\ref{fig:tc}a,e).

\section*{Accurate assessment of compound climate risk}

\begin{figure}[t]
\centering
    \includegraphics[width=0.95\textwidth]{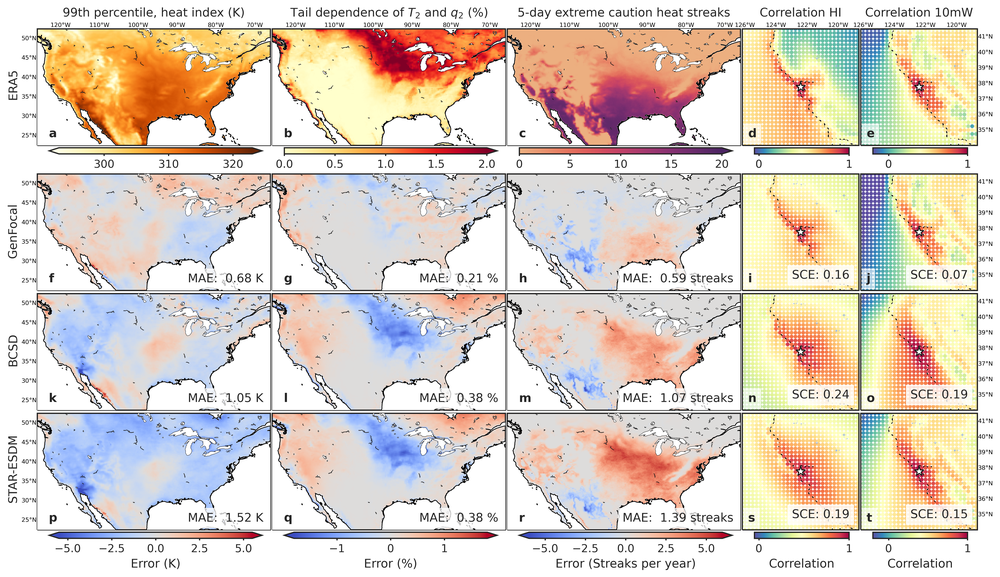}
\caption{\textbf{\ourname accurately assesses projected compound heat extremes over the Conterminous United States (CONUS) during the summer (June-August) of the evaluation period 2010-2019.} \ourname outperforms competing approaches, the Bias Correction and Spatial Disaggregation (BCSD)~\cite{wood2002long,wood2004hydrologic}, a method routinely used for downscaling ensembles from the Coupled Model Intercomparison Project (CMIP)~\cite{Thrasher2022} and the Seasonal Trends and Analysis of Residuals Empirical-Statistical Downscaling Model (STAR-ESDM)~\cite{Hayhoe2024}, a state-of-the-art method recommended for use in the US Fifth National Climate Assessment~\cite{osti_2202926}. For details about those two methods, see \SI~\ref{si:baselines_and_ablations}.
\textbf{a}. Heat index 99$^{th}$ percentile. \textbf{b}. Tail dependence of 2-meter temperature and specific humidity extremes. \textbf{c}. Number of 5-day streaks with ``Extreme Caution'' heat advisory per year. Errors in downscaled estimates are shown for \ourname (\textbf{f}-\textbf{h}), BCSD (\textbf{k}-\textbf{m}), and STAR-ESDM (\textbf{p}-\textbf{r}). \textbf{d,i,n} and \textbf{s}. Spatial correlation of the heat index of San Francisco and its surroundings, evaluated at 18Z for ERA5, \ourname, BCSD, and STAR-ESDM. \textbf{e,j,o} and \textbf{t}. Spatial correlation of the 10m wind speed. Insets show the mean absolute error (MAE) and spatial correlation error (SCE) of the downscaled results.
}\label{fig:conus}
\end{figure}

The risk of compound extremes arises from the cumulative effect of interacting physical processes, such as wildfires fueled by dry vegetation and fanned by strong winds. This type of interdependency is often underestimated by downscaling methods that neglect correlations between hazards and their timescales~\cite{Lopez-Gomez2025,Zscheischler2018}. Humid heatwaves, characterized by prolonged periods of high temperature and humidity, are among the most frequent and impactful of such events, straining human health and power grids. We evaluate the ability of \ourname to represent humid heatwaves by analyzing the risk of summer heat index extremes in the Conterminous United States (CONUS) across timescales (Fig.~\ref{fig:conus}). The physical and spatial structure of heatwaves is further examined in terms of the tail dependence of temperature and humidity extremes and the spatial autocorrelation, respectively. (For the definitions of these metrics, see \SI~\ref{si:evaluation_metrics}.)

\ourname yields accurate estimates of the $99^{\text{th}}$ percentile of the heat index during the summer months, with an average bias reduction over $35\%$ with respect to the statistical downscaling baselines, which systematically underestimate risk (Fig.~\ref{fig:conus}f,k,p). Furthermore, the tail dependence of temperature and humidity extremes demonstrates its superior ability to capture concurrent hazards, with notable improvements across the Midwest and the Western US (Fig.~\ref{fig:conus}g,l,q). These improvements amount to an average error reduction of 44\% with respect to STAR-ESDM and BCSD. \ourname also reproduces the spatial structure of weather patterns, which is strongly affected by fine-scale processes characteristic of regions with diverse topography like California. The spatial correlations of the heat index and wind speed over this region with respect to San Francisco are shown in Fig.~\ref{fig:conus}d-e, evaluated from the ERA5 reanalysis data.
\ourname captures the summertime decorrelation in the heat index between San Francisco and inland California driven by the coastal cooling effect of sea breeze, which increases with inland temperatures (Fig.~\ref{fig:conus}i)~\cite{Lebassi2009}. \ourname also reproduces the complex spatial correlations of wind speed modulated by changes in topography (Fig.~\ref{fig:conus}j). Downscaling methods that do not model spatial correlations explicitly, such as BCSD and STAR-ESDM, typically fail to identify the rich spatial correlation structure from the coarse climate simulation (Fig.~\ref{fig:conus}n,o,s,t).

Heat-related mortality increases with heatwave duration~\cite{Brooke2011}, highlighting the importance of estimating the risk of extended periods of extreme heat. Capturing persistent events requires adequate representation of the temporal coherence of climate fields, which \ourname models explicitly. We assess the skill at predicting extended heatwaves by estimating the risk of 5-day streaks with daily maximum heat indices exceeding 305$~\mathrm{K}$. This threshold corresponds to the ``extreme caution'' heat advisory of the National Oceanic and Atmospheric Administration (NOAA). \ourname provides largely unbiased estimates of 5-day extreme caution heat streaks across the East Coast and the Midwest, compared to the statistical downscaling methods, which tend to overestimate risk in these regions (Fig.~\ref{fig:conus}h,m,r). Overall, \ourname reduces average bias by $44\%$ and $57\%$ compared to BCSD and STAR-ESDM, respectively.

The skillful estimation of compound climate risks by \ourname, demonstrated here for heat waves and previously for tropical cyclones, stems in great measure from its ability to capture correlations across meteorological fields, space, and time. Additionally, the risk estimates provided by \ourname benefit from a more accurate representation of the 
marginal distribution of directly modeled fields than other methods. For example, \ourname reduces the bias of the $99^{\text{th}}$ percentile of near-surface temperature and humidity by more than $20\%$ and $22\%$, respectively (Fig.~\ref{fig:si_conus_99} in \si). Additional results are presented in \SI~\ref{si:conus}.

\section*{Future climate risk assessment}

The design of critical infrastructure with expected lifetimes of decades to centuries requires an assessment of future climate risk. In order to provide reliable assessments, downscaling methods must not only preserve trends projected by the input coarse climate data but also capture the effects of those changes to weather phenomena unresolved by the original projections. Preserving climate change signals can be challenging for statistical downscaling methods trained to \emph{correct biases} over a reference historical period, due to the distortion of climate change trends in the debiasing process~\cite{chandel2024state}. 

We assess the ability of \ourname to evaluate future climate risk by analyzing projected changes in summer heat extremes across cities in the western United States, and trends in tropical cyclone activity in the North Atlantic basin.

\subsection*{Changes in summer heat extremes in the western United States}

\begin{figure}[t]
\centering
\includegraphics[width=0.5\textwidth]{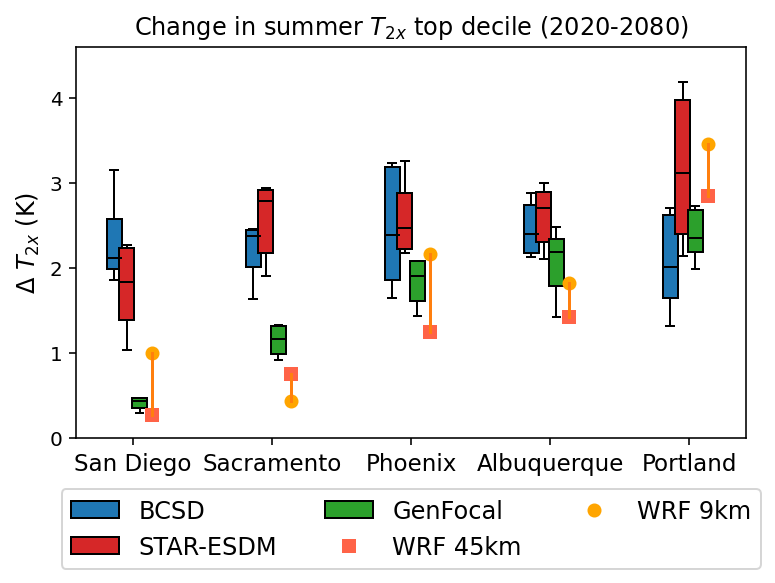}
\caption{\textbf{\ourname projects future regional warming trends from coarse climate simulations, capturing the change patterns in extreme heat across cities in the western U.S. (2020–2080) more consistently than other methods.} Shown is the top decile of daily maximum near-surface temperature. Results are computed as the average over $1^\circ\times 1^\circ$ regions, and 7 summers (June-August) centered around 2020 and 2080. Boxes for BCSD, STAR-ESDM, and \ourname show the interquartile range of an ensemble of 8 projections, and whiskers represent the $12.5\%$ and $87.5\%$ quantiles.
}\label{fig:western_us_clim_change}
\end{figure}

The western United States is expected to experience a substantial increase in extreme heat severity in the coming decades~\cite{meehl_tebaldi_2004}. We evaluate the climate change response of summer temperature extremes projected by \ourname by comparing them to dynamically downscaled climate projections from the Western United States Dynamically Downscaled Dataset~\cite{Rahimi2024}. Although dynamical downscaling is also subject to model errors, its reliance on physics-based modeling relaxes stationarity assumptions and ensures physically consistent climate change patterns~\cite{walton2020}. The dynamical downscaling simulations considered use the Weather Research and Forecasting (WRF) model and take as input data from the same climate model, CESM2, debiased {\it a priori} using the ERA5 reanalysis. We report results for dynamically downscaled projections at $45~\mathrm{km}$ and $9~\mathrm{km}$ resolution, after interpolation to the resolution of \ourname\!, to illustrate variability due to fine-scale processes.

Fig.~\ref{fig:western_us_clim_change} evaluates changes in the top decile of daily maximum temperature across different cities of the Western United States over the period 2020-2080, a complex statistic that requires spatiotemporal downscaling of the input daily-averaged climate data. Results for additional cities and statistics are included in \SI~\ref{si:wus}. \ourname exhibits similar regional warming trends to WRF, with relatively weak warming in coastal San Diego and much stronger warming trends in inland cities such as Albuquerque, Phoenix, and Portland. BCSD and STAR-ESDM fail to capture this modulation of climate change by regional processes, predicting quasi-uniform warming across regions.

\subsection*{Projecting future tropical cyclone risk}

GenFocal demonstrates the ability to realize detailed tropical cyclone activity driven by climate change, based on the underlying large-scale conditions, even when these specific events are not explicitly resolved by the input coarse climate simulations. To show this, we evaluate trends from 2010-2019 to 2050-2059 by producing downscaled results covering 8000 August-October seasons representative of each period with \ourname\!: we downscale 10-year trajectories from the LENS2 ensemble with 8 samples per trajectory.

\begin{figure}[t]
\centering
\includegraphics[width=.8\textwidth]{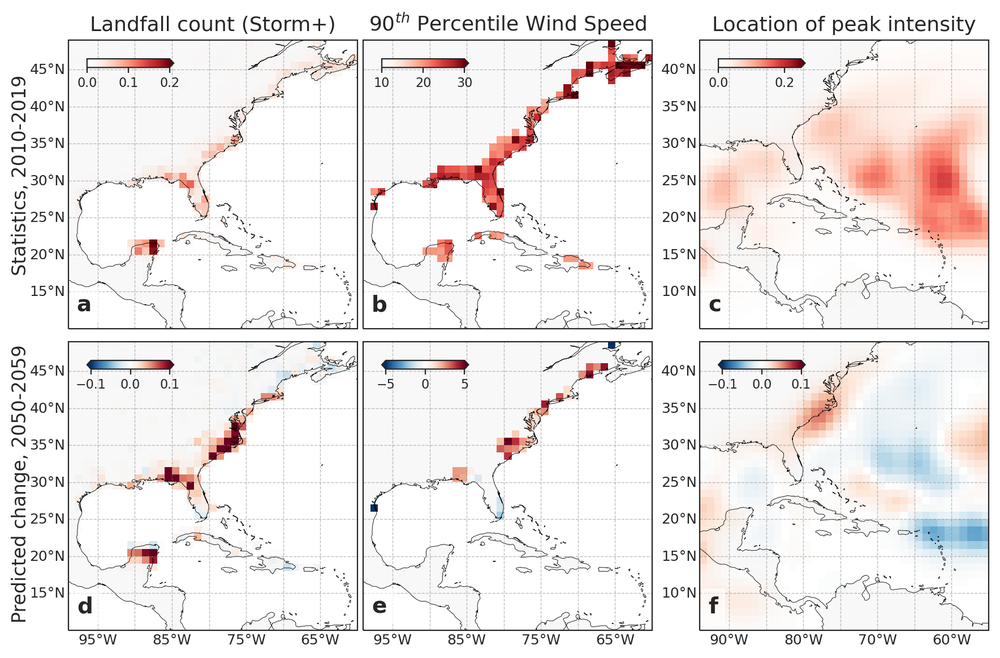}
\caption{\textbf{\ourname projects trends in TC landfall frequency and intensity over the first half of the 21$^{st}$ century consistent with well-established methods~\cite{balaguru_2023} and recent trends~\cite{Kossin2014,Studholme2022}. } 
\textbf{a, d}. Number of tropical storm and hurricane landfalls during the August-October season of years 2010-2019 and its projected change by 2050-2059, respectively.
 \textbf{b, e}. $90^{\text{th}}$ percentile of maximum pressure-derived wind speed (m/s) of landfalling TCs and its projected change over the same periods, respectively. \textbf{c, f}. Spatial distribution of lifetime TC peak intensity and its projected change over the same periods, respectively. All results are computed as the average over 800 downscaled climate projections, hence fractional counts. Changes are displayed only if they are statistically significant ($p < 0.05$ in a two-tailed Mann-Whitney U test) and set to zero otherwise.
}
\label{tc-change}
\end{figure}

Over the first half of the $21^{\text{st}}$ century, \ourname projects an increase in the number of tropical storms and hurricanes making landfall over the U.S. East Coast (Fig.~\ref{tc-change}a,d). This projection aligns with forecasts from other downscaled climate projections, such as the Risk Analysis Framework for Tropical Cyclones (RAFT) model~\cite{balaguru_2023}. These findings contribute to the ongoing scientific investigation and refinement of understanding regarding North Atlantic tropical cyclone landfall trends. \ourname also predicts subtropical intensification and tropical weakening of TCs over the North Atlantic basin (Fig.~\ref{tc-change}e,f), consistent with the observed poleward migration of the location of TC maximum intensity~\cite{Kossin2014, Studholme2022}. The projected TC intensification is largest over the Carolinas and the Mid-Atlantic, with the most intense TCs projected to strengthen at a faster pace (Fig.~\ref{si:fig:tc-change} in \si).

\section*{Discussion}
\label{sDiscuss}

\ourname represents a paradigm shift in climate downscaling, leveraging generative AI to overcome the limitations of traditional methods. 
It is trained directly from coarse climate simulations and weather reanalysis data, without requiring costly RCM simulations to establish temporal alignment between them. It is designed to capture the spatiotemporal, multivariate statistics of climate data accurately, addressing key limitations of statistical downscaling methods, such as BCSD and STAR-ESDM, which are incapable of modeling complex interdependency of multiple variables.
This enables \ourname to quantify the uncertainty of TCs and other compound extreme events robustly. Such tasks have traditionally required narrowly focused, bespoke statistical emulators or computationally expensive dynamical downscaling. 

The practical implications of our method are significant for downstream applications that demand physically-consistent localized climate data.
For instance, accurate spatial correlation modeling can improve system-wide energy grid planning~\cite{Qiu2024}, or by estimating the risk of concurrent heat extremes that increase energy demand and vulnerability of power lines~\cite{dumas2019extreme}.
Additionally, the ability to capture inter-variable correlations, such as those between temperature and humidity, is essential for predicting the heat index, which has direct applications in public health, food production~\cite{Wang2020}, energy demand forecasting~\cite{damiani2024,dumas2019extreme}, and disaster preparedness~\cite{Goss2020}.
Furthermore, directly modeling temporal correlations improves risk estimates for extended extreme events, such as prolonged heat waves and TCs, offering more reliable insights for resilience policies \cite{Delworth1999-yf,Dahl2019-ed}. By providing a full probabilistic characterization of future climate impacts, \ourname enables assessing risks associated with compound hazards involving any number of meteorological extremes interacting across space and time.

Finally, \ourname opens the way for downscaling efficiently large ensembles of climate projections, a computationally intractable task for physics-based downscaling approaches. This is a crucial capability for future risk assessments of regional extremes and rare events, such as tropical cyclones, particularly as advances in AI-accelerated climate simulation \cite{brenowitz2025climate, Watt-Meyer2025} continue expanding our ability to sample global climate uncertainty.

\section*{Methods}
\label{sMethod}
\subsection*{Generative models used by \ourname}

\ourname is a two-step framework: first, a temporal sequence of consecutive climate states, $y\in \mathcal{Y}$, which is coarse in scale and biased, is debiased into an intermediate sequence on the manifold $\mathcal{Y'}$ that is consistent with a sequence of coarse-grained weather states $C'x$ with $x\in \mathcal{X}$, the high-resolution weather manifold. A subsequent super-resolution step increases the spatiotemporal resolution of the debiased sequence while preserving temporal coherence.
This two-staged design decouples learning the debiasing and the super-resolution operations, enabling ``drop-in'' replacement of alternative debiasing operations, as explored in \SI~\ref{si:ablation_models}. 

\paragraph{Super-resolution} We construct $C'$ as a coarsening operation by downsampling the ERA5 data from 2-hourly and $0.25^{^{\circ}}$ to daily and $1.5^{^{\circ}}$, thus forming pairs of aligned data samples $(y_i'= C'x_i, x_i)$. To learn the super-resolution operation, i.e., the inverse of the downsampling, we use a conditional diffusion model~\cite{song2020improved,song2020score}, popularized by latest advances in image and video generation. We take advantage of the prior knowledge that a spatially-interpolated linear mapping $\mathcal{I}(y')$ already contains a strong approximation of the mean statistics of $x$ by modeling the residual $r :=  x - \mathcal{I}(y')$. As such we use the conditional diffusion model to sample from $p(r | y')$ and then add the sampled residual back to $\mathcal{I}(y')$ to obtain the final output of the super-resolution.

The conditional diffusion model learns a neural network based denoiser to iteratively refine a noisy version of the residual $r+\varepsilon\sigma$ to its clean version $r$. The noise is controlled by a scaled Gaussian variable $\varepsilon \sim \mathcal{N}(0,1)$ where the scale $\sigma$ is sampled from a refinement scheduling distribution $\mathcal{Q}$. The denoiser $D_\theta$ is thus trained to minimize the loss function between the refined and the clean residuals: 
\begin{equation}
    \ell(\theta) = \mathbb{E}_{x \in \mu_x} \mathbb{E}_{\sigma\sim\mathcal{Q}(\sigma)} \mathbb{E}_{\varepsilon\sim\mathcal{N}(0, 1)} \| D_\theta( r +\epsilon\sigma, \sigma, y') - r \|^2. 
\end{equation}
Once learned, the denoiser $D_\theta$ is used to  construct a stochastic differential equation (SDE)-based sampler that refines a Gaussian noise signal 
into a clean residual: 
\begin{equation}
    dr_{\tau} = - 2\dot{\sigma}_\tau\sigma_\tau D_\theta\left(r_\tau, \sigma_\tau, y' \right)\;d\tau + \sqrt{2\dot{\sigma}_\tau\sigma_\tau}\;d\omega_\tau,
\end{equation}
in diffusion time $\tau$ from $\tau = \tau_{\text{max}}$ to $0$, and initial condition $r_{\tau_{\text{max}}}\sim\mathcal{N}(0, \sigma^2_{\tau_{\text{max}}}I)$, where $\sigma_\tau:\mathbb{R}\rightarrow\mathbb{R}$ is the diffusion-time dependent noise schedule, controlled by $\mathcal{Q}(\sigma)$.
A more comprehensive description of the diffusion model is included in \SI~\ref{si:diffusion}, along with implementation details. 

While the model $D_\theta$ is trained on short sequences such as one to a few days, we employ an inference procedure to sample extended temporal sequences (spanning multiple months, for example). The procedure achieves temporal coherence through domain decomposition, where each shorter temporal period is a domain and overlapping domains are guided to coherence and contiguity. Details are provided in \SI~\ref{si:multi-diffusion}.  

\paragraph{Debiasing}

Due to the lack of alignment between data sampled from $\mathcal{Y}$ and $\mathcal{Y'}$, we seek a map between their sample distributions. This is a weaker notion than the sample-to-sample correspondence offered by physics-based downscaling methods. However, as demonstrated in this work, achieving a statistical distribution match can effectively debias while remaining computationally advantageous and generating plausible sampled states. 

We leverage the idea of rectified flows~\cite{liu2022rectified} by constructing the debiasing map $T$ as the solution map of an ordinary differential equation (ODE) given by
\begin{equation} \label{eq:reflow_ode}
    \frac{d y}{d \tau} = v_{\phi}(y, \tau) \qquad \text{for } \tau \in [0, 1],
\end{equation}
whose the vector field $v_{\phi}(x, \tau)$ is parametrized by a neural network (see \SI~\ref{si:debiased_architecture} for further details). By identifying the input of the map as the initial condition $y_0 = y(\tau=0)$, we have the solution as the mapping $T(y) := y(\tau = 1)$.  We train $v_{\phi}$ by minimizing loss 
\begin{equation} \label{eq:reflow_ode_estimation}
    \ell(\phi) = \mathbb{E}_{\tau \sim \mathcal{U}[0,1]} \mathbb{E}_{(y_0, y_1) \sim \pi \in \Pi(\mu_y, \mu_{y'})} \| (y_1 - y_0) - v_{\phi}(y_{\tau}, \tau) \|^2, 
\end{equation}
where $y_{\tau} = \tau y_1 + (1-\tau) y_0$.  $\Pi(\mu_y, \mu_{y'})$ is the set of couplings observing the marginal distributions of $\mathcal{Y}$ and $\mathcal{Y'}$ respectively. Once $v_{\phi}$ is learned, we debias any given $y$ by solving \eqref{eq:reflow_ode} from $\tau=0$ to $\tau=1$ using the $4^{\text{th}}$-order Runge-Kutta ODE solver.

Analogous to super-resolution, we also learn a debiasing map that takes into consideration a temporal sequence of climate variables. In \SI~\ref{si:debias}, we describe a simple way to achieve this as well as other important implementation details, such as selection of the coupling $\Pi(\mu_y, \mu_{y'})$ \rev{(see Supplementary Information section \ref{si:ablations_coupling} for an ablation study of different choices)} and parametrization of $v_\phi$ with various neural architecture choices.

\subsection*{Evaluation protocols and metrics}

The downscaling methods are evaluated in two categories of metrics. The first set of metrics evaluates the discrepancy between the distributions of the downscaled climate data and the corresponding ERA5 weather data. Three types of discrepancies are measured. The first measures the univariate differences at each site, which are averaged in space to give rise to mean absolute bias (MAB), Wasserstein distance (WD) and percentile mean absolute error (MAE). The second measures spatial correlation and temporal spectrum errors. The last type measures correlation discrepancies among different variables such as tail dependence, an important quantity for compound extremes. \SI~\ref{si:evaluation_metrics} provides detailed definitions of the evaluation metrics. 

The second category of metrics is application-specific. In this work, we focus on North Atlantic tropical cyclones and severe and prolonged heat events over CONUS. In either case, nontrivial processing is performed on the output variables to compute composite variables (such as heat indices, the number of heat streak days) and TC occurrences and tracks. Evaluation metrics vary and we describe them in detail in \SI~\ref{si:evaluation_protocol}.

\section*{Data availability} 
The data for training the models, pretrained model weights, as well as debiased and downscaled forecasts produced by \ourname, are available on Google Cloud (\url{https://console.cloud.google.com/storage/browser/genfocal}). Dynamically downscaled projections from the WUS-D3 dataset are available at \url{https://registry.opendata.aws/wrf-cmip6}.

\section*{Code availability}
Source code for our models and evaluation protocols can be found on GitHub \url{https://github.com/google-research/swirl-dynamics/tree/main/swirl_dynamics/projects/genfocal}.
\section*{Author contribution}

L.Z.N., F.S.,  Z.Y.W. and I.L.G. conceptualized the work.  F.S. and L.Z.N. managed the project. R.C., Z.Y.W., and I.L.G. curated the data. L.Z.N., Z.Y.W.
and F.S. developed the model and algorithms. Z.Y.W and L.Z.N. wrote the modeling codes. I.L.G. and R.C. supplemented with additional modeling and analysis codes. Z.Y.W and L.Z.N. conducted the modeling experiments. Z.Y.W., L.Z.N., I.L.G. and R.C. performed analysis, visualization and evaluation. I.L.G. and R.C. investigated literature and contextualized the results.  I.L.G., L.Z.N., Z.Y.W., and F.S. wrote the original draft. L.Z.N., Z.Y.W and F.S. led the subsequent revisions in additional experiments and writings. J. A. and T.S. advised the project and provided disciplinary science expertise.  All reviewed and edited the paper.

\section*{Declaration}
The authors declare no competing interests.

\section*{Acknowledgement}

We thank Lizao Li and Stephan Hoyer for productive discussions, and Daniel Worrall and John Platt for feedback on the manuscript. For the LENS2 dataset, we
acknowledge the CESM2 Large Ensemble Community Project and the supercomputing resources provided by the IBS Center for Climate Physics in South Korea.
ERA5 data \cite{Hersbach2023-pp} were downloaded from the Copernicus Climate Change Service\cite{Copernicus-Climate-Change-Service-Climate-Data-Store2023-rm}.  
The results contain modified Copernicus Climate Change Service information. Neither the European Commission nor ECMWF is responsible for any
use that may be made of the Copernicus information or data it contains. We thank Tyler Russell for managing data acquisition and other internal business processes. We thank referees for their invaluable comments and advices.

\putbib
\end{bibunit}

\newpage

\cleardoublepage
\renewcommand{\thesection}{\Alph{section}}
\setcounter{page}{1}
\setcounter{section}{0}
\setcounter{equation}{0}

\begin{titlepage}
\begin{center}
\vfill
{\large Supplementary Information}\\[2em]
Regional Climate Risk Assessment from Climate
Models \\ Using Probabilistic Machine Learning\\[1em]
Z.Y.~Wan, I.~Lopez-Gomez, R.~Carver, T.~Schneider, J.~Anderson,\\ F.~Sha, L.~Zepeda-N\'u\~nez\\
\today
\vfill
\end{center}
\end{titlepage}

\startlist{toc}
\printlist{toc}{}{\section*{Table of Contents}}
\newpage
\begin{bibunit}

\section{Related work}
\label{si:related}

\begin{figure*}[h!]
    \centering
    \hfill
    \includegraphics[height=4.4cm, trim={0cm 0cm 0cm 0cm}, clip]{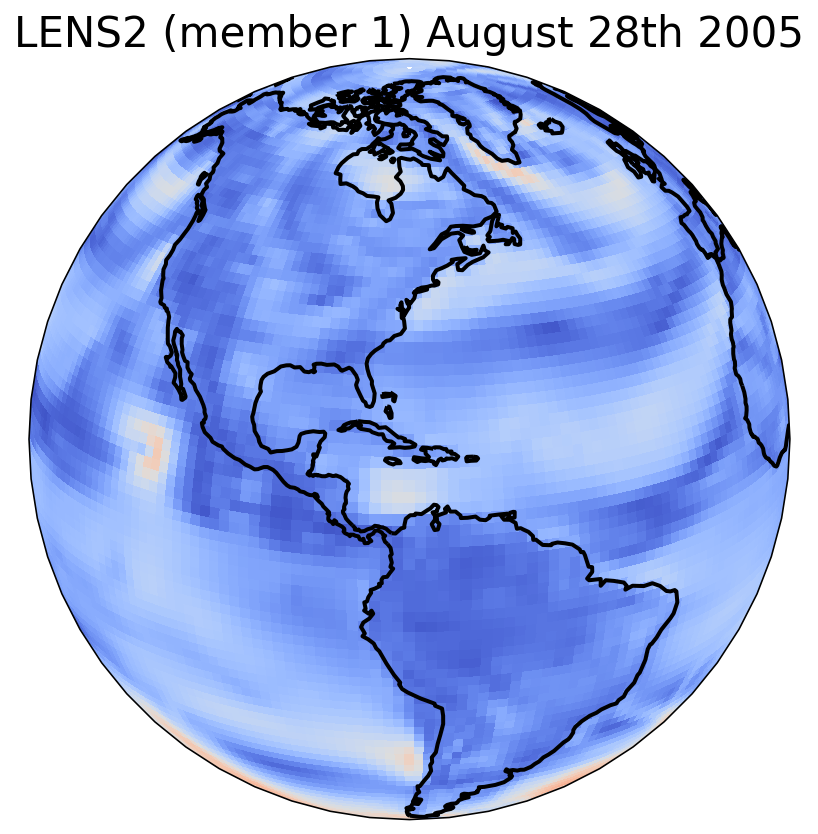}
    \hfill
    \includegraphics[height=4.4cm, trim={0cm 0cm 0cm 0cm}, clip]{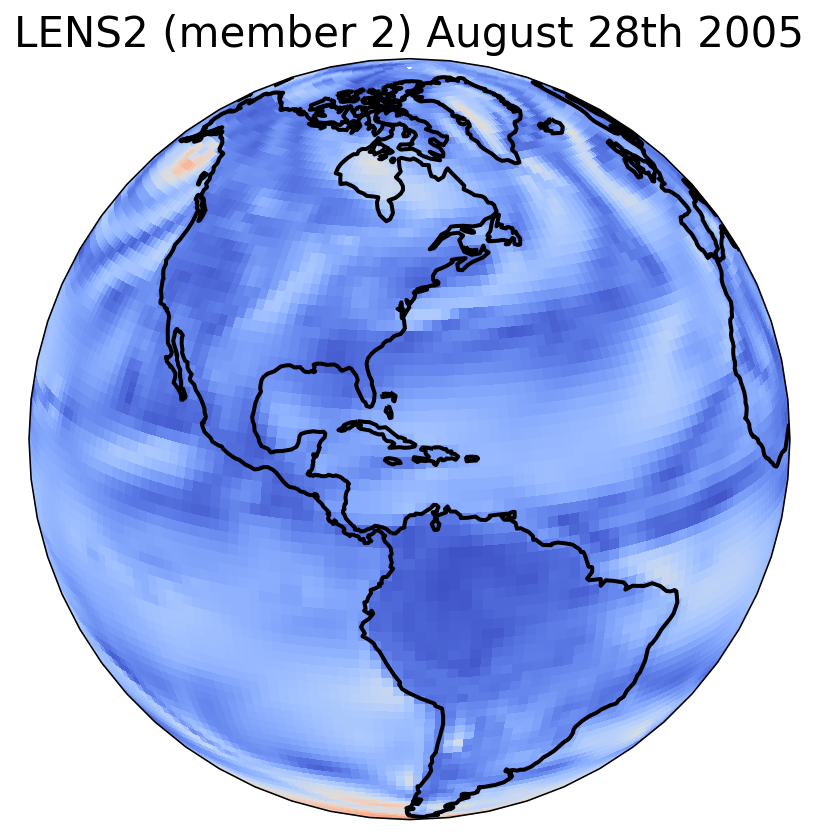}
    \hfill
    \includegraphics[height=4.4cm, trim={0cm 0cm 0cm 0cm}, clip]{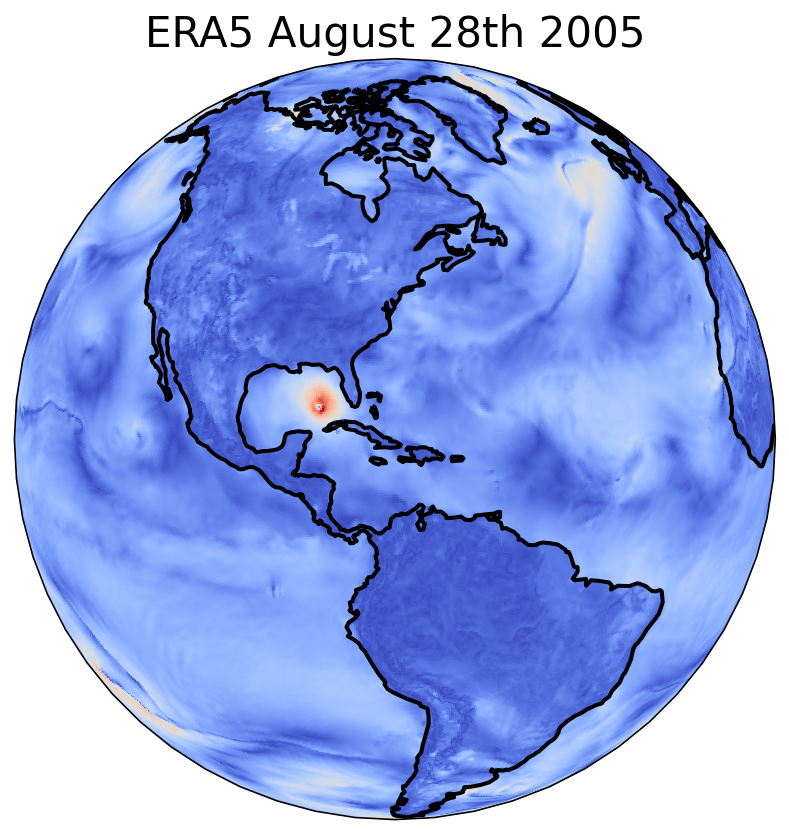}
    \hfill
    \caption{\textbf{Ten meter wind speed for August 28, 2005.} Daily average of Global 10 meter wind speed for the day of August 28 2005, from two ensemble members of LENS2 and ERA5, where we can observe the substantial differences between the different samples, particularly, as hurricane Katrina (a red blob next to Florida in the ERA5 sample) is absent from the climate samples.}
  \label{fig:unpaired_samples}
\end{figure*}

Existing ML downscaling methods predominantly rely on temporal alignment between low- and high-resolution training data~\cite{mardani2023generative,Lopez-Gomez2025,merizzi2024wind,xiao2024clouddiff,yi2025efficient,lyu2024downscaling,watt2024generative,srivastava2024precipitation}, which enables treating the problem as a supervised learning task. This framework has proved effective for downscaling weather forecasts ~\cite{mardani2023generative} and regional climate model emulation~\cite{Lopez-Gomez2025}. However, this supervised learning approach cannot be used to learn a downscaling model for free-running climate projections to be consistent with observation assimilated historical datasets such as ERA5~\cite{bischoff2022:unpaired}, since there is no temporal alignment between the coarse projections and the high-resolution data (e.g., see Fig.~\ref{fig:unpaired_samples}). \ourname adopts an original learning framework to address this challenge of mapping between unpaired data.

\ourname also stands in stark contrast to the traditional dynamical downscaling paradigm of using regional climate models (RCMs) to generate high-resolution climate data ~\cite{Giorgi2019}. Due to its high computational cost (even at regional scale, sacrificing spatial coverage), the use of dynamical downscaling is limited to small climate-projection ensembles, thus compromising their ability to capture the risk of climate extremes~\cite{Goldenson2023}. Alternative statistical downscaling approaches are more efficient~\cite{Pierce_2014_LOCA,Hayhoe2024} but come at the cost of highly customized designs for specific use cases~\cite{Pierce_2023} or fail to  capture the full range of spatiotemporal correlations between meteorological fields that characterize climate~\cite{chandel2024state}. Those  methods are highly bespoke: methods used in hydrology~\cite{wood2002long} are markedly different from those used for tropical cyclone analysis~\cite{Jing2021}. This inflexibility limits the value they add to coarse climate projections, compared to the more flexible but expensive physics-based approaches.

In what follows, we review the existing literature on downscaling. In the interest of broad coverage, we also briefly review supervised learning methods for downscaling on temporally aligned data, namely, super-resolution, see Tables \ref{table:ml_downscaling_studies} and \ref{table:ml_downscaling_studies_deterministic} for a summary.

\begin{table}[t!]
\centering
\caption{Summary of dowscaling methods with aligned data.}
\label{table:ml_downscaling_studies}
\begin{tabular}{|l|p{3cm}|p{3cm}|p{4cm}|} %
\hline
\textbf{Study} & \textbf{Technique Used} & \textbf{Resolution (Low $\rightarrow$ High)} & \textbf{Variables Downscaled} \\
\hline
\cite{watt2024generative} & Diffusion Model & 2$^{\circ}$ (Coarsened ERA5) $\rightarrow$ 0.25$^{\circ}$ (ERA5) & 2m Temperature, 10m Winds \\
\hline
\cite{mardani2023generative} & Corrector Diffusion Model & 25 km (ERA5) $\rightarrow$ 2 km (WRF Model) & 2m Temperature, 10m Winds, Radar Reflectivity \\
\hline
\cite{pathak2024kilometer} & Generative Diffusion Model for CAM Emulation & $\sim$ 28 km (ERA5) $\rightarrow$ 3 km (HRRR Model) & Full atmospheric state \\
\hline
\cite{merizzi2024wind} & Diffusion model & 25 km (ERA5) $\rightarrow$ 5.5 km (CERRA) & Surface Temperature, Wind Speed, Geopotential Height \\
\hline
\cite{srivastava2024precipitation} & Video Diffusion & 200 km (FV3GFS \cite{zhou2019toward}) $\rightarrow$ 25 km (FV3GFS \cite{zhou2019toward}) & Precipitation \\
\hline
\cite{xiao2024clouddiff} & Generative Diffusion Model (CloudDiff) & 2 km (Himawari-8 AHI \cite{okuyama2015preliminary}) $\rightarrow$ 1 km (MODIS) & Cloud related variables \\
\hline
\cite{tomasi2025can} & Latent Diffusion Model & 16 km (ERA5) $\rightarrow$ 2 km (COSMO\_CLM) & 2m Temperature, 10m Wind \\
\hline
\cite{yi2025efficient} & Wavelet Diffusion Model & 10 km (MRMS) $\rightarrow$ 1 km (MRMS) & Precipitation (Composite Reflectivity) \\
\hline
\cite{Vandal_2017:DeepDS} & Super-Resolution CNN & $\sim $ 1.25 $^{\circ}$ (MERRA-2) $\rightarrow$ 0.25$^{\circ}$ (CPC Obs.) & Daily Precipitation \\
\hline
\cite{bano2022downscaling} & CNN (DeepESD) under Perfect Prognosis & 2$^{\circ}$ (ERA-Interim) $\rightarrow$ 0.5$^{\circ}$ (E-OBS) & Daily Temperature \& Precipitation \\
\hline
\cite{jha2025deep} & ResNets (VDSR, EDSR) vs. SRCNN & 2.5$^{\circ}$ (ERA5) $\rightarrow$ 0.25$^{\circ}$ (ERA5) & 2m Temperature \\
\hline
\cite{koldunov2406emerging} & AI-NWP (Pangu-Weather) & 250 km (CMIP6) $\rightarrow$ 31 km (ERA5-like) & Full atmospheric state (focus on 2m Temperature) \\
\hline
\cite{harder2023hard} & Hard-Constrained Deep Learning & 9 km (WRF) $\rightarrow$ 3 km (WRF) & Total Column Water, Temperature, Water Vapor, Liquid Water \\
\hline
\end{tabular}
\end{table}

\begin{table}[h!]
\centering
\footnotesize
\caption{Summary of downscaling methods for unaligned data.}
\label{table:ml_downscaling_studies_deterministic}
\begin{tabular}{|l|p{3cm}|p{3cm}|p{4cm}|} %
\hline
\textbf{Study} & \textbf{Technique Used} & \textbf{Resolution (Low $\rightarrow$ High)} & \textbf{Variables Downscaled} \\
\hline
\ourname & Rectified Flow + Diffusion Model & 1.5$^{\circ}$ (LENS2) $\rightarrow$ 0.25$^{\circ}$ (ERA5) & 2m Temperature, Surface Humidity, Sea-level Pressure, 10m Winds \\
\hline
\cite{climalign:2021} & Nearest neighbor interpolation + Normalizing flows + CycleGAN loss & 1$^{\circ}$ (NOAA20CR~\cite{compo2011twentieth}) $\rightarrow$ 0.125$^{\circ}$ (Livneh Obs \cite{livneh2013long}) & Precipitation, Daily max temperature\\
\hline
\cite{bischoff2022:unpaired} & Filtered-weighted nearest neighbor + Diffusion Bridge & 2D
forced turbulent fluid and a supersaturation tracer &  Vorticity, Super-saturation\\
\hline
\end{tabular}
\end{table}

\subsection{Supervised learning for super-resolution}
\label{si_supervised}

Supervised learning is the most direct approach for downscaling aligned data, where a model learns a mapping from low- to high-resolution data using paired samples~\cite{mardani2023generative,tomasi2025can}. Recent works using this approach are summarized in Table~\ref{table:ml_downscaling_studies}. These supervised methods can be broadly categorized as either deterministic or probabilistic. Most deterministic models operate under the ``perfect prognosis'' assumption~\cite{Vandal_2017:DeepDS}, which assumes the low-resolution samples are unbiased (see \cite{rampal2024enhancing} for an extensive review). These models seek to capture correlations between large-scale meteorological fields and local observations by training on time-aligned samples from data-assimilated simulations (e.g., ERA5) and local observational data~\cite{bano2020configuration,bano2021suitability,bano2022downscaling}. The tendency of deterministic models to smoothen outputs and dampen extremes by collapsing to the conditional mean~\cite{molinaro2024generative} has motivated the development of probabilistic methods, which typically employ generative models such as diffusion models~\cite{merizzi2024wind,xiao2024clouddiff,yi2025efficient,lyu2024downscaling,watt2024generative} or GANs~\cite{harris2022generative,price2022increasing}.

One way to leverage these supervised learning approaches for climate downscaling is to enforce temporal alignment between the climate models and the high-resolution data. This can achieved by dynamical downscaling~\cite{Lopez-Gomez2025,tomasi2025can}, or by nudging the coarse climate model of interest to follow the historical weather record~\cite{charalampopoulos2023statistics,dixon2016evaluating}. These approaches come with their own limitations: they require access to a well-calibrated regional climate model in the case of dynamical downscaling, and modifying the input climate model in the case of nudging climate projection. Apart from these technical barriers, both methods require running additional climate simulations that are typically computationally expensive.
\ourname overcomes the restriction of temporal alignment and addresses head-on the challenge of learning to downscale coarse climate projections to be consistent with historical weather data.

\subsection{Empirical statistical downscaling methods}

The unpaired data assumption has been a staple of weather and climate science for decades, with many methods developed to address it \cite{Wilby_1998:downscaling,panofsky1968some,hayhoe2024star,Pierce_2014_LOCA}.
In the weather and climate literature (see~\cite{vandal2017intercomparison,burger2012downscaling} for extensive overviews), prior knowledge can be exploited to downscale specific variables~\cite{Wilby_1998:downscaling,panofsky1968some}.
Two of the most predominant methods of this type are bias correction and spatial disaggregation (BCSD)~\cite{wood2002long,wood2004hydrologic}, which combines traditional spline interpolation with a quantile matching bias correction~\cite{Maraun2013:quantile_matching} and a disaggregation step—and linear models~\cite{Hessami_2008:linear_models_statistical_dowscaling}.
More recent methods extend this type of methodology by adding climate signal corrections with different timescales to the climatology~\cite{hayhoe2024star}, or by adjusting historical weather analogs to follow an evolving climate~\cite{Pierce_2014_LOCA}. The statistical simplicity of these methods comes with limitations: they do not explicitly model spatial or inter-variable correlations, which hinders their use for risk analysis of derived weather variables, such as the heat index.

\subsection{Generative modeling approaches}
\label{si:related_ot}

The bias between the low-resolution and the high-resolution one can be seen as a distribution mismatch from characterizing differently the same underlying system. Thus, removing the bias can be framed as an instance of unsupervised domain adaptation \cite{gong2012geodesic}, a topic popularly studied in computer vision. Recent work has used generative models such as GANs and diffusion models to bridge the gap between two domains, usually under the umbrella of image-to-image translation or neural style transfer~\cite{bortoli2021diffusion,meng2021sdedit,pan2021learning,park2020contrastive,sasaki2021:UNIT-DDPM,su2023:dual_diffusion,wu2022unifying,zhao2022egsde}.

Similar to ours, several recent work have adopted this view to perform debiasing. However, unlike ours, those methods increase resolution \emph{first}, then debias. \ourname instead debiases first, then upsample. This deliberate design is methodologically sound and avoid several pitfalls:
\begin{itemize}
    \item Upsampling techniques such as interpolation, may incur aliasing~\cite{climalign:2021,bischoff2022:unpaired} and introduce compounded errors when debiasing.
    \item Upsamling requires the low-frequency power spectra of the two datasets match~\cite{bischoff2022:unpaired}, precisely the problem of bias that we need to solve.
\end{itemize}

Another notable innovation is that \ourname performs effectively temporal super-resolution with temporal coherence. In contrast, other approaches only sample snapshots and thus do not impose temporal coherence in the resulting sequences, a necessary requirement to capture accurately statistics of extended events such as tropical storms and extreme heat waves~\cite{bischoff2022:unpaired,climalign:2021,wan2023debias}.

In practice, debiasing first at low-resolution is computationally less expensive. As \ourname leverages diffusion models to perform super-resolution, we avoid well-known issues associated with GANs~\cite{tian2022generative,ballard2022contrastive} and normalized-flows based methods~\cite{lugmayr2020srflow}, which include over-smoothing, mode collapse, and large model footprints~\cite{dhariwal2021diffusion,Li2022:SRDiff}. 

Those methodological and computational advantages allow us to downscale large sequence of high-resolution (in both space and time) samples with ease. This task is beyond the capabilities of the methods mentioned above. (In our earlier work, we have compared a version of \ourname, which outperforms those methods on flow problems that are feasible for them. For details, please see~\cite{wan2023debias}.)

\section{\ourname identifies accurately tropical cyclones (TCs) in the North Atlantic} \label{si:nao}

\subsection{Physically plausible TC tracks and structures}

\begin{figure}[tbh!]
\centering
    \includegraphics[width=0.99\textwidth]{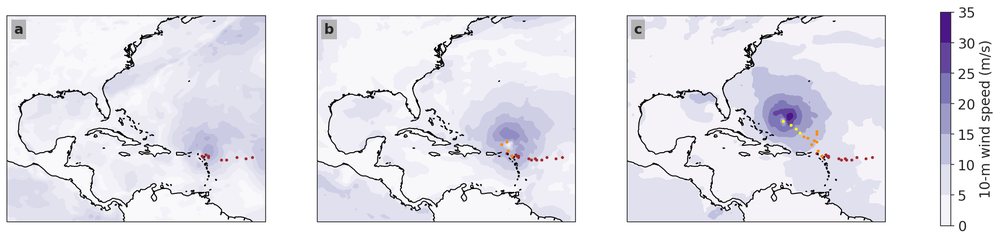}
    \caption{\textbf{Evolution of a hurricane downscaled by \ourname\!.} Plots of 10-meter wind speed at 60-hour intervals for a Category 1 hurricane projected by \ourname. Colored dots track the tropical cyclone eye and its intensity in the Saffir-Simpson scale. The tropical cyclone evolves from a depression (brown) to a storm (orange) and ultimately a Category 1 hurricane (yellow).}
    \label{si:fig:tctrack}
\end{figure}

Fig.~\ref{si:fig:tctrack} shows the wind speed at 60-hour intervals for a Category 1 hurricane from a climate projection downscaled with \ourname. The track shows the TC moving westerly, north of the Lesser Antilles, before recurving towards the north. The plots of 10-meter wind speed show that the strongest winds are in the right, front quadrant of the tropical cyclone. Both of these features are characteristics consistent with tropical cyclones in the North Atlantic basin.

\subsection{Track density}

\begin{figure}[tbh!]
\centering
\includegraphics[width=0.99\textwidth]{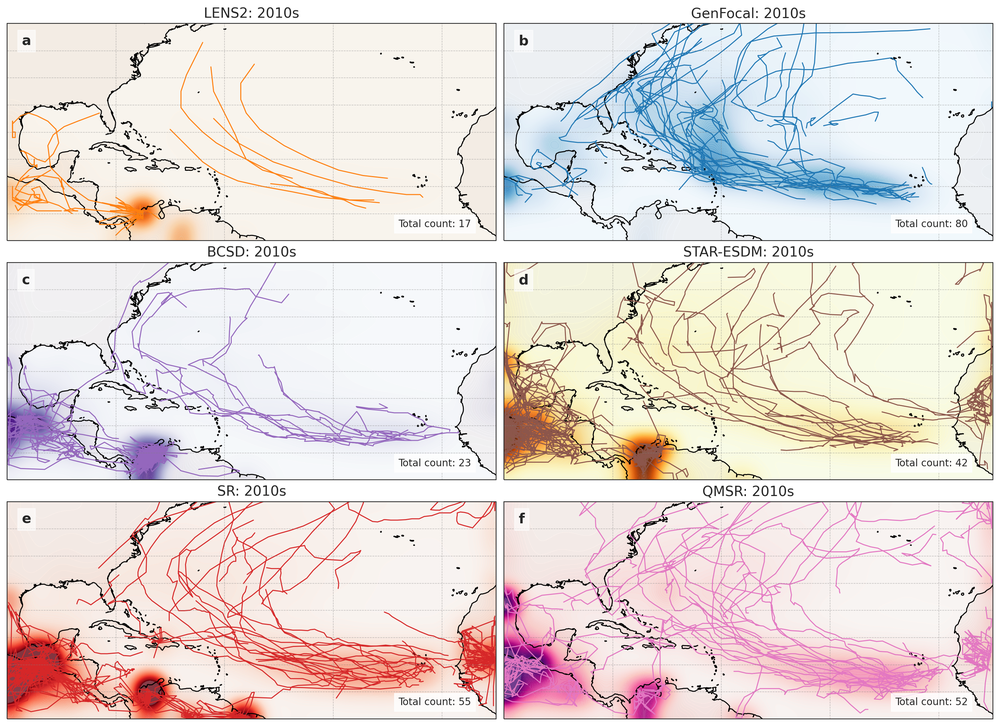}
\caption{\textbf{TC tracks and their density.} Tracks and their density for a LENS2 member in the North Atlantic in the time period 2010-2019 (\textbf{a}), for the same member we show a sample generated by 
 \ourname (\textbf{b}), BCSD (\textbf{c}), STAR-ESDM (\textbf{d}), SR (\textbf{e}) and QMSR (\textbf{f}). 
 The observed tracks from the ERA5 reanalysis are shown in Fig. \ref{fig:tc}\textbf{c}. Note other methods (including the original LENS2) have the unrealistic concentration of TCs over Venezuela and the Pacific Coast of Mexico as well as the unphysical tracks over the Sahara desert. 
}\label{fig:si_tracks}
\end{figure}

Fig.~\ref{fig:si_tracks} shows the TC tracks and densities in the original CESM2 Large Ensemble (LENS2) data and corresponding downscaled ensembles.  \ourname\!, shown in Fig.~\ref{fig:si_tracks}b, produces the most realistic TCs with a density that is remarkably close to the observed one in the ERA5 reanalysis, shown in Fig.~\ref{fig:tc}c. The other models shown in Fig.~\ref{fig:tc}c-f underestimate the number of TCs, overestimate TC track length, and project an unrealistic concentration of TCs over Venezuela and the Pacific Coast of Mexico. In addition, state-of-the-art (SoTA) statistical downscaling methods such as BCSD and STAR-ESDM (\ref{si:baselines_and_ablations}), as well as GenFocal variants without the generative debiasing component such as SR and QMSR (\ref{si:ablation_models}) predict unphysical tracks over the Sahara desert. 

\subsection{Counts and intensities of TCs}
\begin{figure}[tbh!]
\centering
\includegraphics[width=0.75\textwidth]{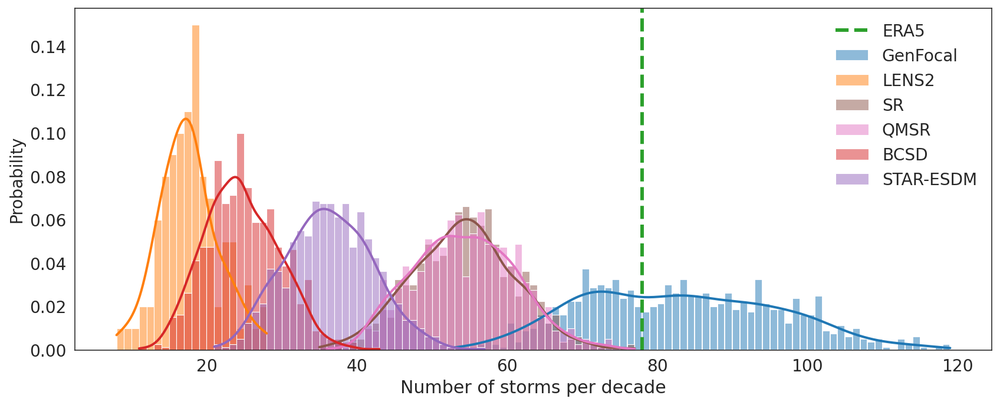}
\caption{\textbf{Distribution of TC counts.} Distributions of North Atlantic TC counts in the August-September-October season of 2010-2019 for the raw and downscaled LENS2 ensemble (100 members), using the different methods considered, and the ERA5 ground truth.
}\label{fig:si_counts_distribution}
\end{figure}

\begin{figure}[tbh!]
\centering
\includegraphics[width=0.75\textwidth]{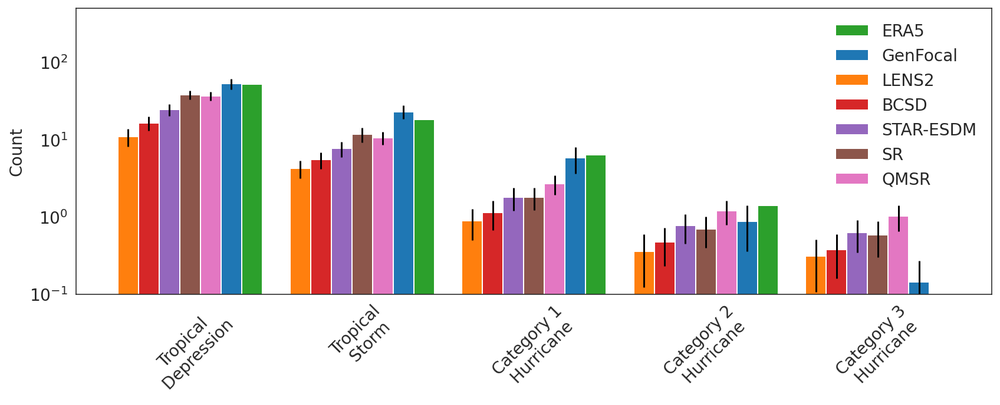}
\caption{\textbf{Distributions of TC intensity.} Distributions of intensity (the Saffir-Simpson Hurricane Wind Scale) of detected tropical cyclones in the North Atlantic in the August-September-October period during 2010-2019.
}\label{fig:si_intensity_distribution}
\end{figure}

\begin{figure}[tbh!]
\centering
\includegraphics[width=0.75\textwidth]{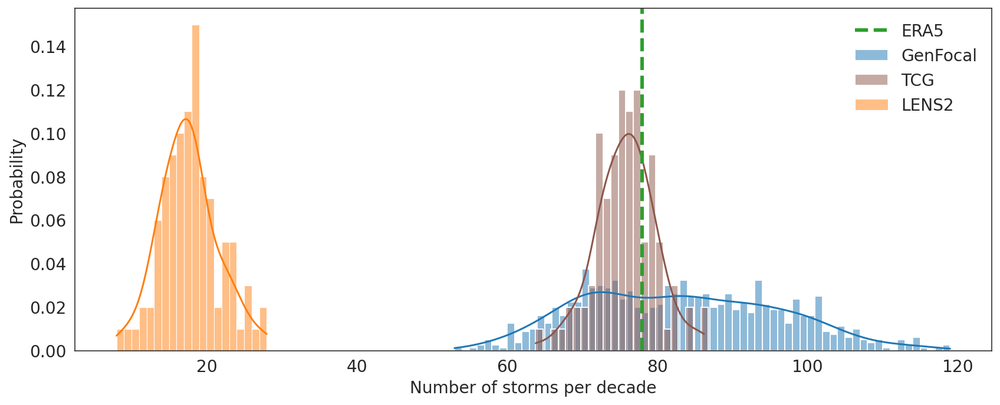}
\caption{\textbf{Distribution of decadal storm counts.} Histogram of decadal storm counts produced by the LENS2 ensemble, the count distribution predicted using the tropical cyclogenesis (TCG) index, and the count distribution in the \ourname ensembles.}
\label{fig:si_tcg_histogram}
\end{figure}
\begin{figure}[tbh!]
\centering
\includegraphics[width=0.99\textwidth]{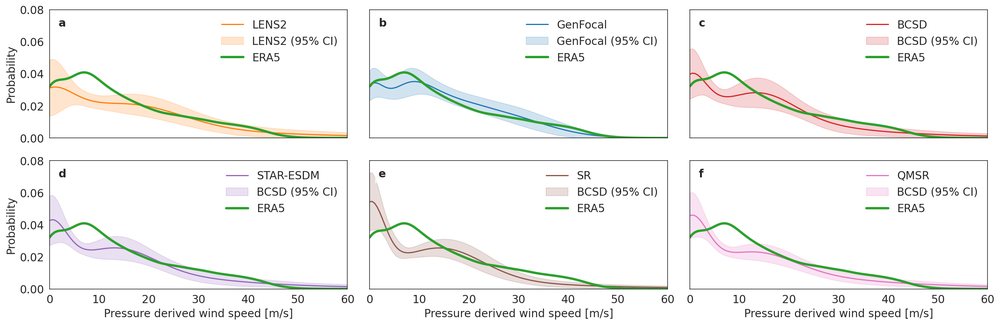}
\caption{\textbf{Distribution of TC wind speeds.} Distributions of the pressure-derived wind speed of tropical cyclones detected in the North Atlantic basin in the August-September-October period during 2010-2019, for LENS2 (\textbf{a}), \ourname (\textbf{b}), BCSD (\textbf{c}), STAR-ESDM (\textbf{d}), SR (\textbf{e}), and QMSR (\textbf{f}). In addition, we also add the distribution of the pressure-derived wind speed for the reference ERA5 dataset. The confidence intervals are computed across the ensemble dimension.
}\label{fig:si_wind_distribution}
\end{figure}

Fig.~\ref{fig:si_counts_distribution} shows the 
number of detected TCs in the North Atlantic in the August-September-October season of period 2010-2019. \ourname produces TC counts consistent with observations, in contrast to other methods, which underestimate the number of TCs. Fig.~\ref{fig:si_intensity_distribution} shows the 
distributions of detected TC intensities. \ourname generates distributions closely matching those in ERA5, whereas other methods tend to underestimate the number of Category 1 Hurricanes and Tropical Storms and Depressions while overestimating the number of Category 3 Hurricanes\footnote{This bias stems from the very small pressure drop threshold (induced by the calibration procedure in ~\ref{si:tc_calibration}) needed to calibrate other downscaling methods for optimal TC detection. The calibrated threshold pressure drop for methods other than \ourname is either $20$Pa or $40$Pa, whereas the pressure drop for \ourname is $120$Pa. A small threshold pressure drop inflates the calibrated pressure-derived wind speed and ultimately results in higher intensity storms.}.

Fig.~\ref{fig:si_tcg_histogram} demonstrates that LENS2 produces significantly fewer storms in a decade than anticipated by the Tropical Cyclogenesis (TCG) index, based on large-scale patterns, while \ourname produces decadal storm counts that are comparable to the TCG predictions and are able to generate plausible TCs whose fine details are realistic and detectable.

Fig.~\ref{fig:si_wind_distribution} shows the superior performance of \ourname at estimating the distribution of pressure-derived wind speeds, whereas other methods tend to systematically underestimate the probability of 5-10 (m/s), and overestimate the probability of 45-60 (m/s) winds, where the observed ones in reanalysis have almost no mass. This result is consistent with Fig.~\ref{fig:si_intensity_distribution}.

\subsection{Morphology of detected TCs tracks}

\begin{figure}[tbh!]
\centering
\includegraphics[width=0.99\textwidth]{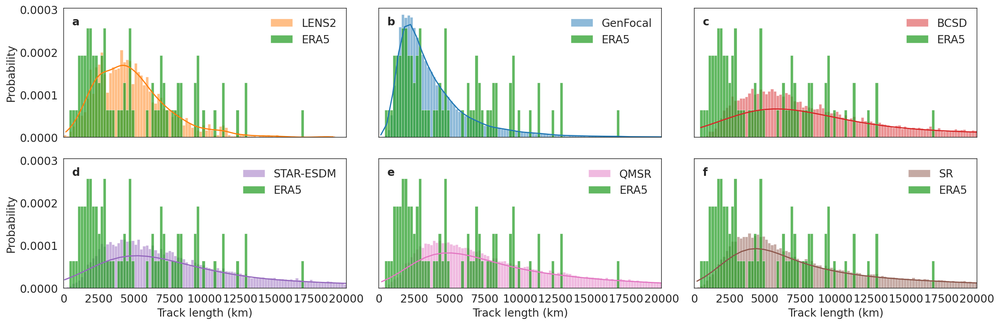}
\caption{\textbf{Distributions of TC track length.} Distributions of the track lengths of detected tropical cyclones in the North Atlantic in the August-September-October period during 2010-2019, for LENS2 (\textbf{a}), \ourname (\textbf{b}), BCSD (\textbf{c}), STAR-ESDM (\textbf{d}), SR (\textbf{e}), and QMSR (\textbf{f}). In addition, we also add the distribution of the track lengths detected in the reference ERA5 dataset.
}\label{fig:si_track_length_distribution}
\end{figure}

\begin{figure}[tbh!]
\centering
\includegraphics[width=0.99\textwidth]{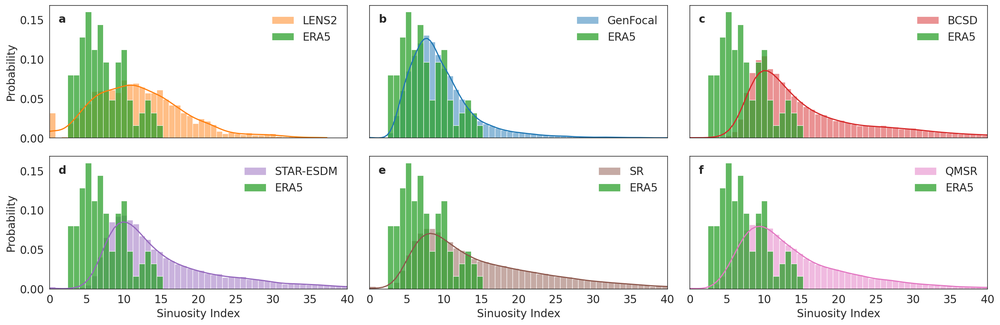}
\caption{\textbf{Distributions of sinuousity indices.} Distributions of the sinuosity indices of the detected tropical cyclones tracks in the North Atlantic in the August-September-October period during 2010-2019, for LENS 2 (\textbf{a}), \ourname (\textbf{b}), BCSD (\textbf{c}), STAR-ESDM (\textbf{d}), SR (\textbf{e}), and QMSR (\textbf{f}).
}\label{fig:si_sinuosity_idx_distribution}
\end{figure}

\ourname accurately captures the distribution of TC track lengths, closely matching the reanalysis data (Fig.~\ref{fig:si_track_length_distribution}). In contrast, other methods tend to overestimate track length. Fig.~\ref{fig:si_sinuosity_idx_distribution} shows that \ourname also excels at capturing  the sinuosity indices of the detected tracks. The sinuosity index ($SI$)  provides a proxy to the geometrical shapes of the tracks~\cite{Terry2015-sx}.  It is a transformation of the sinuosity, $S$ of a storm path, which is defined as 
\begin{equation}
    S =\frac{l_{\text{path}}}{l_\text{direct}},
\end{equation}
where $l_\text{path}$ is the total path length and $l_\text{direct}$ is the direct length between the start and end points of the track.  $SI$ is defined as
\begin{equation}
    SI = \sqrt[3]{\left(S -1\right)} \times 10
\end{equation}
A sinuosity index of 0 indicates a straight track, and it increases for more sinuous tracks. Fig.~\ref{fig:si_sinuosity_idx_distribution} shows that the tracks induced by \ourname have a distribution similar to ERA5, whereas other methods tend to produce overly sinuous (and even erratic) tracks, as observed in Fig.~\ref{fig:si_tracks}.

\section{\ourname models accurately multivariate spatiotemporal statistics}
\label{si:conus}

We assess how well the  multivariate probabilistic distributions over spatial and temporal dimensions are captured by the downscaling procedures, compared to those in ERA5 during the evaluation period (2010-2019). We focus on the summer (June-July-August) period over the Conterminous United States (CONUS) region.

We evaluate skill using the mean absolute bias (MAB) (\ref{si:bias}), mean Wasserstein distance (MWD) (\ref{si:wass}), and mean absolute error (MAE)  in the $99^{\text{th}}$ percentile (\ref{si:percentiles}). Please refer to those sections for the definitions. Our results demonstrate that \ourname effectively captures marginal distributions, joint distributions, distributions of derived variables (via nonlinear transformations), and their tails.

\subsection{Statistics of single variables}

\begin{table}[tbh!]
\footnotesize
\centering
\caption{Statistical modeling errors of directly downscaled variables in marginal distributions by different methods for the summers (June-July-August) in CONUS during 2010-2019. Best highlighted in bold.}
{
\setlength\tabcolsep{2.15pt}
\label{table:bias_and_wass}
\begin{tabular}{lccccc}
\hline
\multirow{2}{*}{Variable} & \ourname & BCSD  & STAR-ESDM & QMSR  & SR  \\ %
                          & \multicolumn{5}{c}{Mean Absolute Bias $\downarrow$}  \\\hline
Temperature (K)           & \textbf{0.41}	& 0.55	& 0.89	& 0.54	& 2.03	 \\
Wind speed (m/s)          & 0.19	& \textbf{0.12}	& 0.15	& 0.15	& 1.83	 \\
Specific humidity (g/kg)  & \textbf{0.31}	& 0.39	& 0.54	& 0.43	& 1.41	 \\
Sea-level pressure (Pa)   & 39.92	& 45.59	& \textbf{38.88}	& 44.98	& 160.98 \\ \hline
\multirow{2}{*}{} &    \multicolumn{5}{c}{Mean Wasserstein Distance $\downarrow$}              \\ %
Temperature (K)           & \textbf{0.47}	& 0.61	& 0.93	& 0.59	& 2.08	                      \\
Wind speed (m/s)          & 0.22	& 0.18	& 0.18	& \textbf{0.16}	& 1.84	                     \\
Specific humidity (g/kg)  & \textbf{0.36}	& 0.42	& 0.56	& 0.46	& 1.47	                       \\
Sea-level pressure (Pa)   & 52.09	& 47.46	& \textbf{41.47}	& 46.85	& 162.61	                   \\ \hline
\multirow{2}{*}{} & \multicolumn{5}{c}{Mean Absolute Error, $99^{\text{th}}$ $\downarrow$}          \\  %
Temperature (K)           & \textbf{0.61}	& 0.77	& 1.04	& 0.82	& 2.63                       \\
Wind speed (m/s)          & 0.48	& \textbf{0.39}	& 0.50	& 0.41	& 2.43                       \\
Specific humidity (g/kg)  & \textbf{0.45}	& 0.58	& 0.74	& 0.63	& 1.67                       \\
Sea-level pressure (Pa)   & 77.99	& 68.02	& 67.66 & \textbf{58.33}& 209.98                     \\ \hline
\end{tabular}
}
\end{table}

Table~\ref{table:bias_and_wass} compares the ability of \ourname and other methods to capture the marginal distributions of the downscaled variables for summers (JJA) during the 2010-2019 evaluation period. \ourname is highly competitive, often achieving the lowest errors.

While quantile mapping (QM) is statistically optimal for matching marginal distributions, it requires a subsequent super-resolution (SR) step to increase resolution. Consequently, methods like QMSR, BCSD, and STAR-ESDM show comparable performance. In contrast, SR alone performs poorly due to biases in the coarse simulation. Although \ourname's debiasing stage targets the joint distribution, it also frequently improves its performance on marginals.

\begin{figure}[tbh!]
\centering
\includegraphics[width=0.8\textwidth]{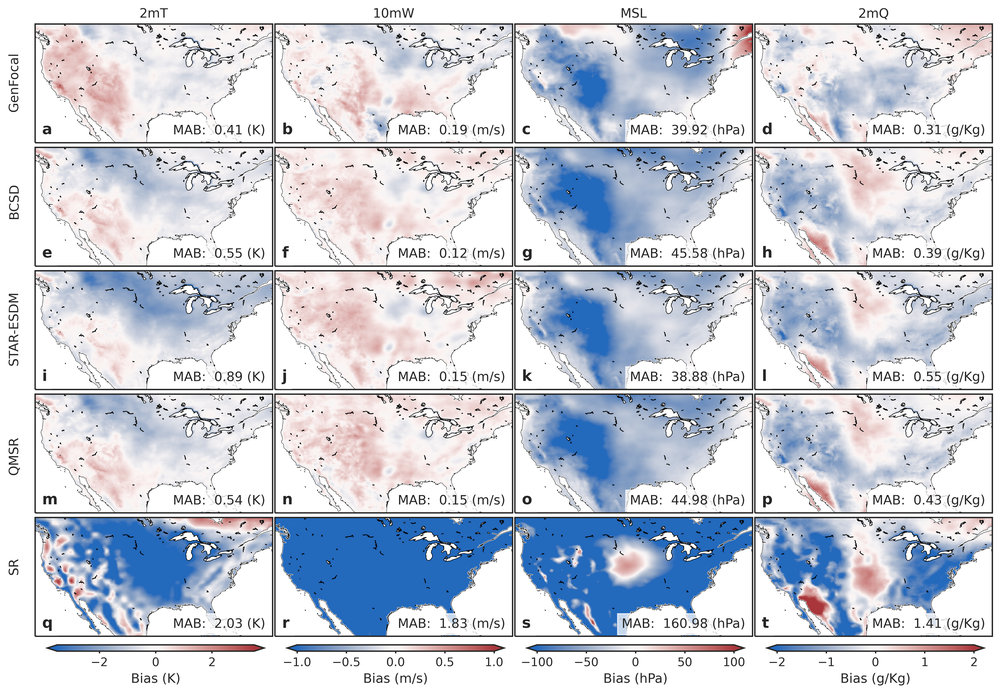}
\caption{\textbf{Pointwise bias of weather variables.} Pointwise bias over CONUS during the summers (June-August) of the evaluation period 2010-2019 for the 2 m temperature, 10 m wind speed, mean sea-level pressure and 2 m specific humidity for \ourname (\textbf{a-d}), BCSD (\textbf{e-h}), STAR-ESDM  (\textbf{i-l}), QMSR (\textbf{m-p}), and SR (\textbf{q-t}). 
}\label{fig:si_conus_bias}
\end{figure}

\begin{figure}[tbh!]
\centering
\includegraphics[width=0.8\textwidth]{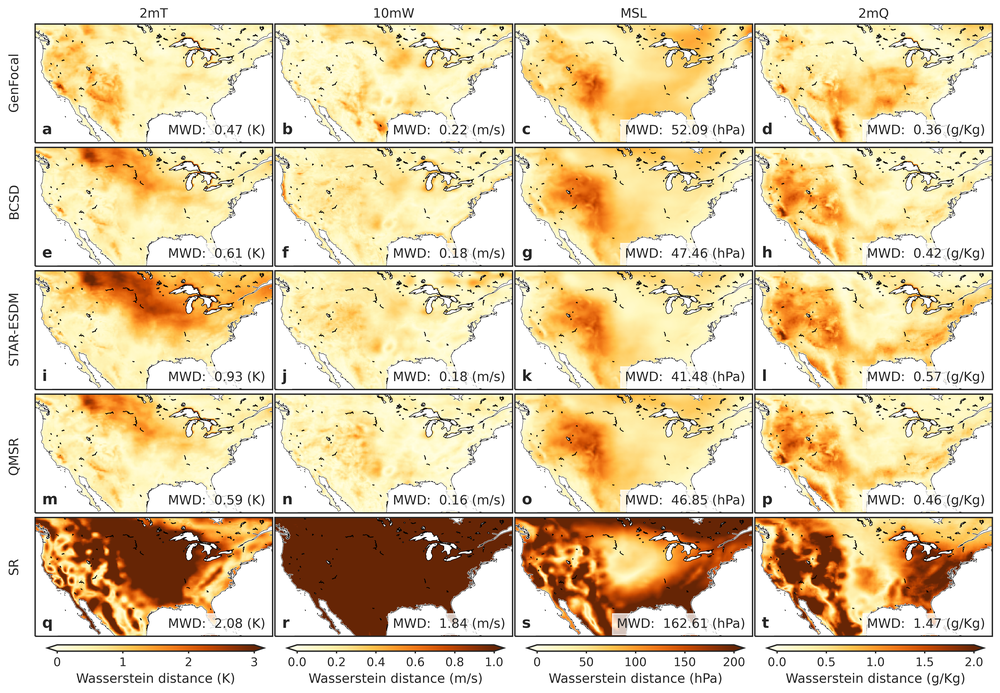}
\caption{\textbf{Pointwise Wasserstein distance.} Pointwise Wasserstein distance between marginals over CONUS during the summer (June-August) for the evaluation period 2010-2019 for the 2 m temperature, 10 m wind speed, mean sea-level pressure and 2 m specific humidity for \ourname (\textbf{a-d}), BCSD (\textbf{e-h}), STAR-ESDM (\textbf{i-l}), QMSR (\textbf{m-p}), and SR (\textbf{q-t}). 
}\label{fig:si_conus_wass}
\end{figure}

Figs.~\ref{fig:si_conus_bias}--\ref{fig:si_conus_wass} show the spatial distribution of bias and Wasserstein distances (defined in \ref{si:pointwise}) between the downscaled ensemble generated using different methods and ERA5 during the evaluation period. \ourname ensembles show low bias and Wasserstein distance across all variables (Fig.~\ref{fig:si_conus_bias}), outperforming all methods in near-surface temperature and humidity.
As previously noted, the absence of a debiasing step in SR leads to substantial biases and errors as shown in Fig.~\ref{fig:si_conus_bias}(q-t) and \ref{fig:si_conus_wass}(q-t). Generally, the fields are significantly underestimated, except for humidity, which is severely overestimated in the eastern California Gulf (Mexico) and the central United States. 

\begin{figure}[tbh!]
\centering
\includegraphics[width=0.8\textwidth]{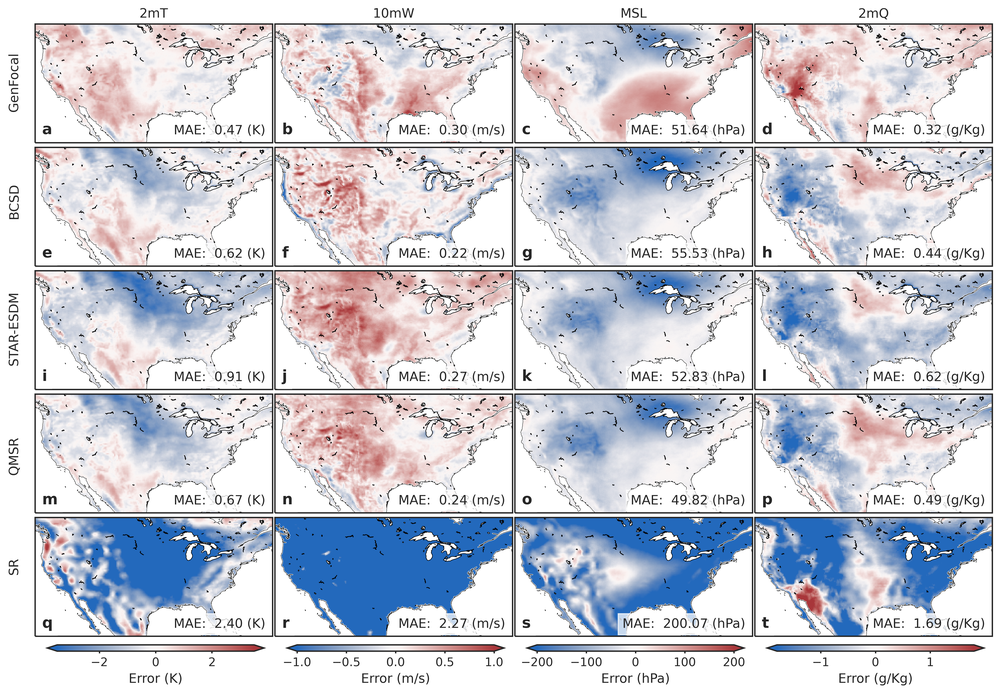}
\caption{\textbf{Error of the $95^{\text{th}}$ percentile.} Error of the $95^{\text{th}}$ percentile over CONUS during the summer (June-August) for the evaluation period 2010-2019 for the 2 m temperature, 10 m wind speed, mean sea-level pressure and 2 m specific humidity for \ourname (\textbf{a-d}), BCSD (\textbf{e-h}), STAR-ESDM  (\textbf{i-l}), QMSR (\textbf{m-p}), and SR (\textbf{q-t}). 
}\label{fig:si_conus_95}
\end{figure}

\begin{figure}[tbh!]
\centering
\includegraphics[width=0.8\textwidth]{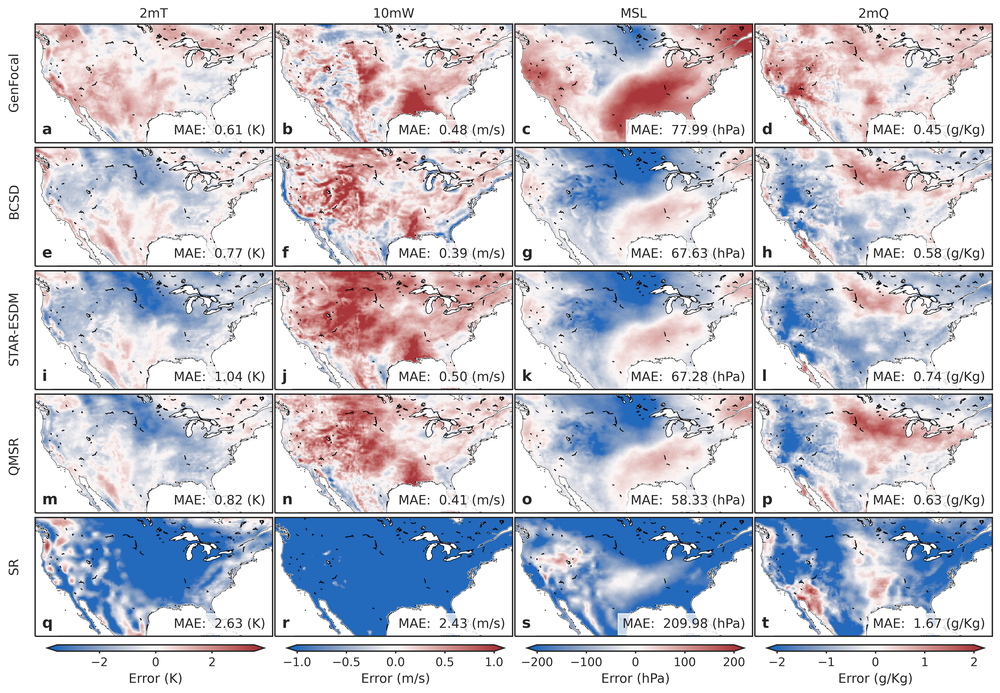}
\caption{\textbf{Error of the $99^{\text{th}}$ percentile.} Error of the $99^{\text{th}}$ percentile over CONUS during the summer (June-August) for the evaluation period 2010-2019 for the 2m temperature, 10m wind speed, mean sea-level pressure and 2 m specific humidity for \ourname (\textbf{a-d}), BCSD (\textbf{e-h}), STAR-ESDM  (\textbf{i-l}), QMSR (\textbf{m-p}), and SR (\textbf{q-t}). 
}\label{fig:si_conus_99}
\end{figure}

\ourname is also superior in recovering extreme statistics, as shown in
Fig.~\ref{fig:si_conus_95} and Fig.~\ref{fig:si_conus_99} for the pixel-wise errors at the $95^{\text{th}}$ and $99^{\text{th}}$ percentiles for directly modeled variables, respectively.

The results in Figs.~\ref{fig:si_conus_bias}-\ref{fig:si_conus_99} and  Table~\ref{table:bias_and_wass} demonstrate that the debiasing step, through either quantile mapping (QM) or \ourname, is crucial for obtaining statistically accurate high-resolution outputs. In contrast, super-resolution (SR) alone incurs large errors, especially in the distributional tails.

\subsection{Statistics of derived variables}

\begin{figure}[tbh!]
\centering
\includegraphics[width=0.8\textwidth]{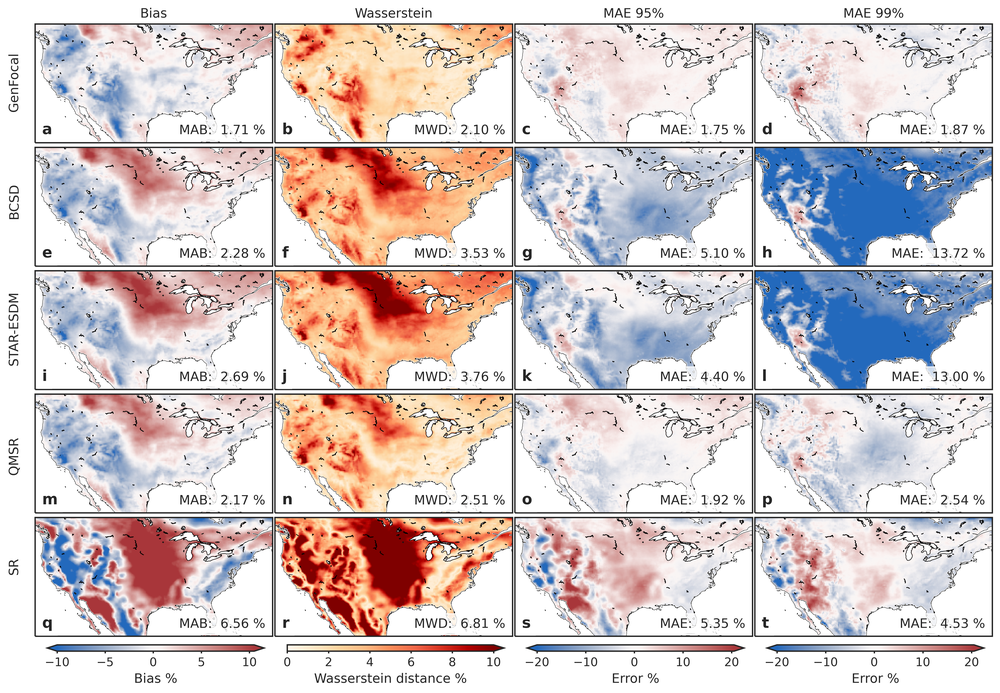}
\caption{\textbf{Spatial distribution of modeling errors.} Spatial distribution of modeling errors for the relative humidity over CONUS during the summer (June-August) of the evaluation period 2010-2019. Pointwise bias, Wasserstein distance, and errors of the $95^{\text{th}}$ percentile and $99^{\text{th}}$ percentile are reported for \ourname (\textbf{a-d}), BCSD (\textbf{e-h}), STAR-ESDM  (\textbf{i-l}), QMSR (\textbf{m-p}), and SR (\textbf{q-t}). 
}\label{fig:si_conus_rh}
\end{figure}

\begin{figure}[tbh!]
\centering
\includegraphics[width=0.8\textwidth]{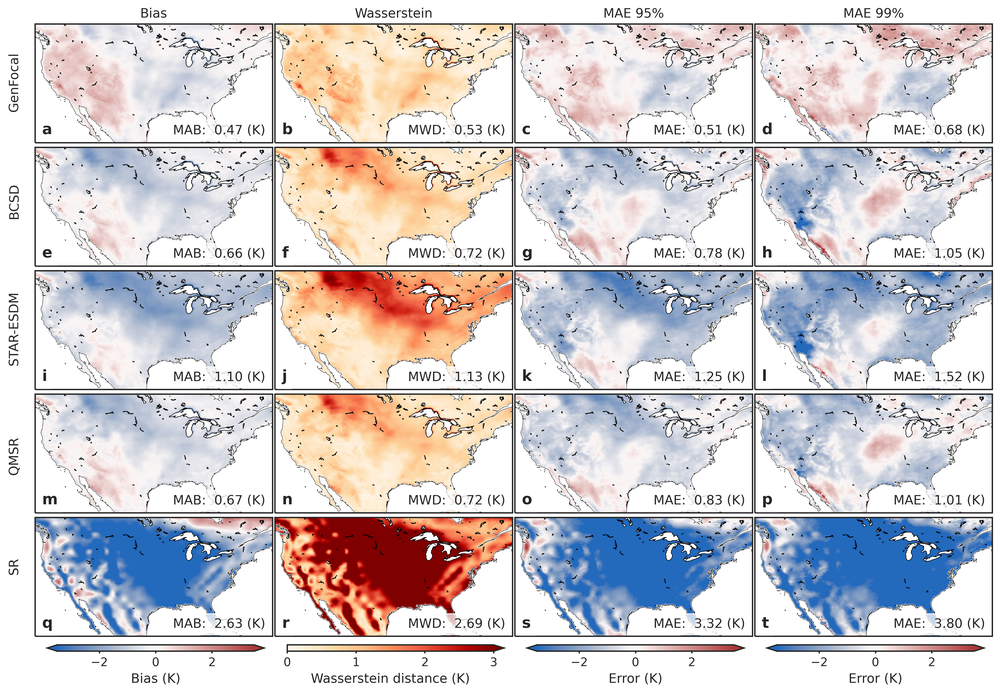}
\caption{\textbf{Spatial distribution of modeling errors.} Spatial distribution of modeling errors for the heat index, over CONUS during the summers (June-August) of the evaluation period 2010-2019. Pointwise bias, Wasserstein distance, and errors of the $95^{\text{th}}$ percentile and $99^{\text{th}}$ percentile are reported for \ourname (\textbf{a-d}), BCSD (\textbf{e-h}), STAR-ESDM  (\textbf{i-l}), QMSR (\textbf{m-p}), and SR (\textbf{q-t}). 
}\label{fig:si_conus_hi}
\end{figure}

\ourname models explicitly the joint distribution of output variables, capturing inter-variable correlations neglected by downscaling methods that model each variable independently. We showcase the benefits of this approach by computing the statistics of derived variables that depend nonlinearly on the directly modeled variables, and comparing them to the ground truth during the evaluation period. 
We consider the near-surface relative humidity and the heat index (see~\ref{si:derived} for the definition), nonlinear functions of temperature and humidity that have important effects on human health and comfort. The heat index is also used to define heat streaks in~\ref{si:heat_streaks}. The tracking errors of the statistics for the derived variables are summarized in Table~\ref{table:bias_and_wass_composite}, demonstrating that GenFocal substantially outperforms other methods.
The spatial distributions of tracking errors are illustrated in Figs.~\ref{fig:si_conus_rh} and \ref{fig:si_conus_hi}. GenFocal shows substantial reductions in relative humidity bias and Wasserstein distance with respect to other methods over the Midwestern United States (Fig.~\ref{fig:si_conus_rh}). Error reductions are even more substantial and broadly distributed at the tails of the distribution. Similarly, \ourname also reduces errors for the heat index, although less substantially.

\begin{table}[t!]
\footnotesize
\centering
\caption{Statistical modeling errors of \textbf{derived variables} by different models for the summers (June-August) in CONUS during 2010-2019.}
{
\setlength\tabcolsep{2.15pt}
\label{table:bias_and_wass_composite}
\begin{tabular}{lccccc}
\hline
\multirow{2}{*}{Variable} & \ourname & BCSD  & STAR-ESDM & QMSR  & SR  \\ 
                          & \multicolumn{5}{c}{Mean Absolute Bias $\downarrow$}  \\\hline
Relative humidity (\%)    & \textbf{1.71}	& 2.28	& 2.70	& 2.17	& 6.56	 \\
Heat index (K)            & \textbf{0.47}	& 0.66	& 1.11	& 0.67	& 2.63   \\ \hline
\multirow{2}{*}{} &    \multicolumn{5}{c}{Mean Wasserstein Distance $\downarrow$}              \\
Relative humidity (\%)    & \textbf{2.10}	& 3.54	& 3.77	& 2.51	& 6.81	                       \\
Heat index (K)            & \textbf{0.53}	& 0.72	& 1.14	& 0.72	& 2.69                      \\ \hline
\multirow{2}{*}{} & \multicolumn{5}{c}{Mean Absolute Error, $99^{\text{th}}$ $\downarrow$}          \\
Relative humidity (\%)    & \textbf{1.87}	& 13.68	& 12.96	& 2.54	& 4.53                       \\
Heat index (K)            & \textbf{0.68}	& 1.05	& 1.52	& 1.01	& 3.80                       \\ \hline
\end{tabular}
}
\end{table}

\subsection{Extreme statistics of joint distributions}
\begin{figure}[tbh!]
\centering
\includegraphics[width=0.8\textwidth]{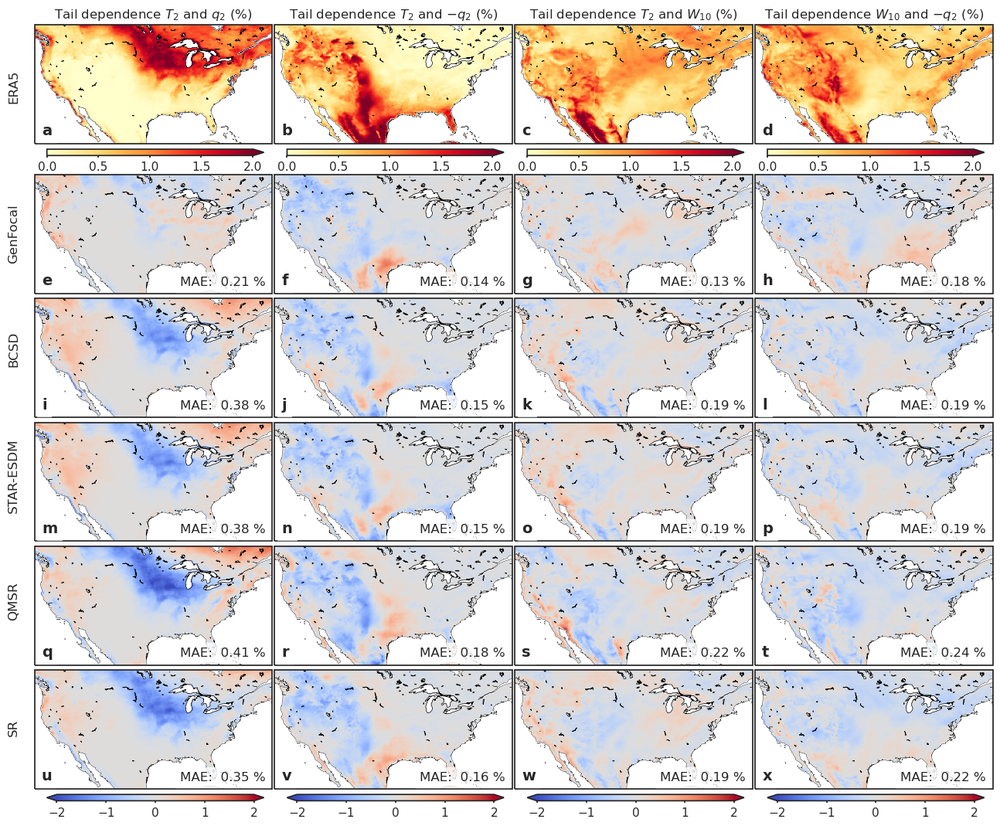}  %
\caption{\textbf{Tail dependence of meteorological extremes.} Tail dependence of pairs of meteorological extremes over period 2010-2019 from ERA5 and bias of downscaling methods. Tail dependence shown for high temperature and humidity (\textbf{a}), high temperature and low humidity (\textbf{b}), high temperature and high wind (\textbf{c}), and high wind and low humidity (\textbf{d}). Tail dependence biases are shown for \ourname (\textbf{e-h}), BCSD (\textbf{i-l}), STAR-ESDM (\textbf{m-p}), QMSR (\textbf{q-t}), and SR (\textbf{u-x}). 
}\label{fig:si_conus_tails}
\end{figure}

In Fig.~\ref{fig:si_conus_tails}, we investigate further \ourname's capacity in capturing the correlation of meteorological extremes in terms of the tail dependence (see~\ref{si:tail_dependence} for its definition). The tail dependence evaluates the probability that two variables will present extreme behavior simultaneously, which is of great importance for assessing compound risk. High temperature and humidity extremes can have important effects on human health, whereas dry hot extremes can increase agricultural losses.

\ourname captures well the frequency of humid and hot extremes (Fig.~\ref{fig:si_conus_tails}a,e). All other methods considered tend to underestimate the co-occurrence of extremely humid and hot conditions in the U.S. Midwest (Fig.~\ref{fig:si_conus_tails}i,m,q,u). All methods show higher skill at capturing dry and hot summer extremes, with \ourname and BCSD providing the most accurate assessment of compound risk (Fig.~\ref{fig:si_conus_tails}f,j). Results are also presented for the co-occurrence of high wind speeds and temperatures, and high wind speeds and low humidity. For both, \ourname presents the lowest tail dependence bias with respect to the ERA5 reanalysis.

\subsection{Spatial correlations} \label{si:sec:conus_spatial_correlation}

\begin{figure}[tbh!]
\centering
\includegraphics[width=.7\textwidth]{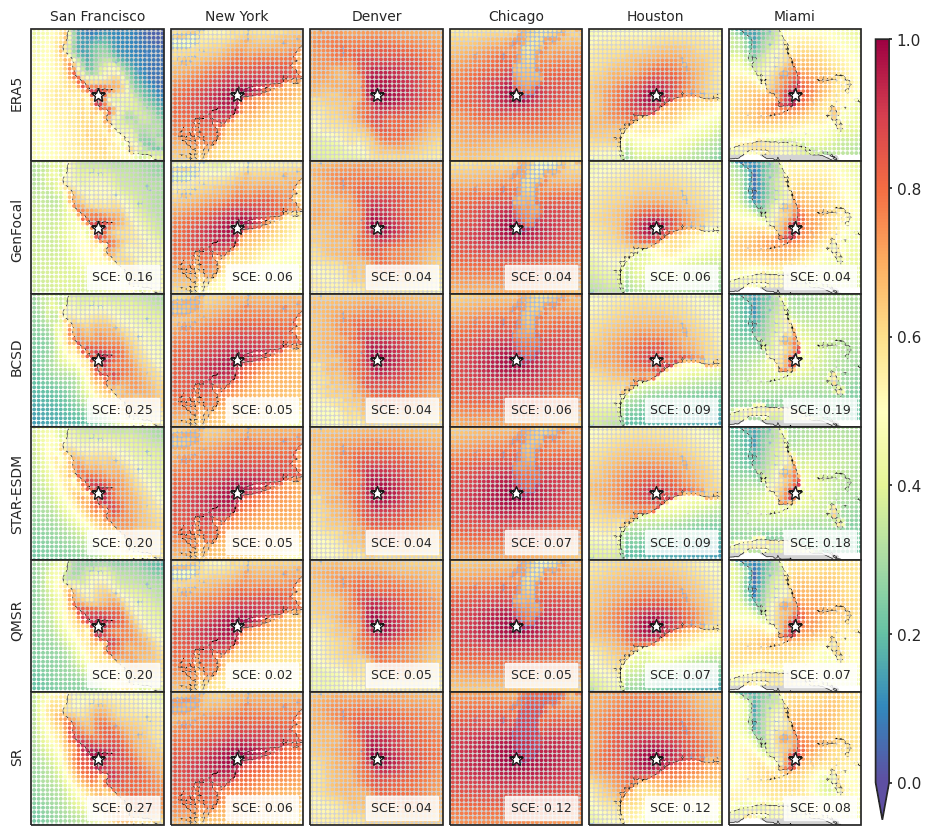}
\caption{\textbf{Spatial correlation for 2-m temperature.} Spatial correlation for 2~m temperature around selected populous US cities, evaluated for all snapshots at 18:00 UTC. The color scale represents the correlation coefficient relative to the city (stars) within a $\pm4^\circ$ longitude/latitude range.}
\label{fig:si_conus_spatial_correlations_2mT_18utc}
\end{figure}

\begin{figure}[tbh!]
\centering
\includegraphics[width=.7\textwidth]{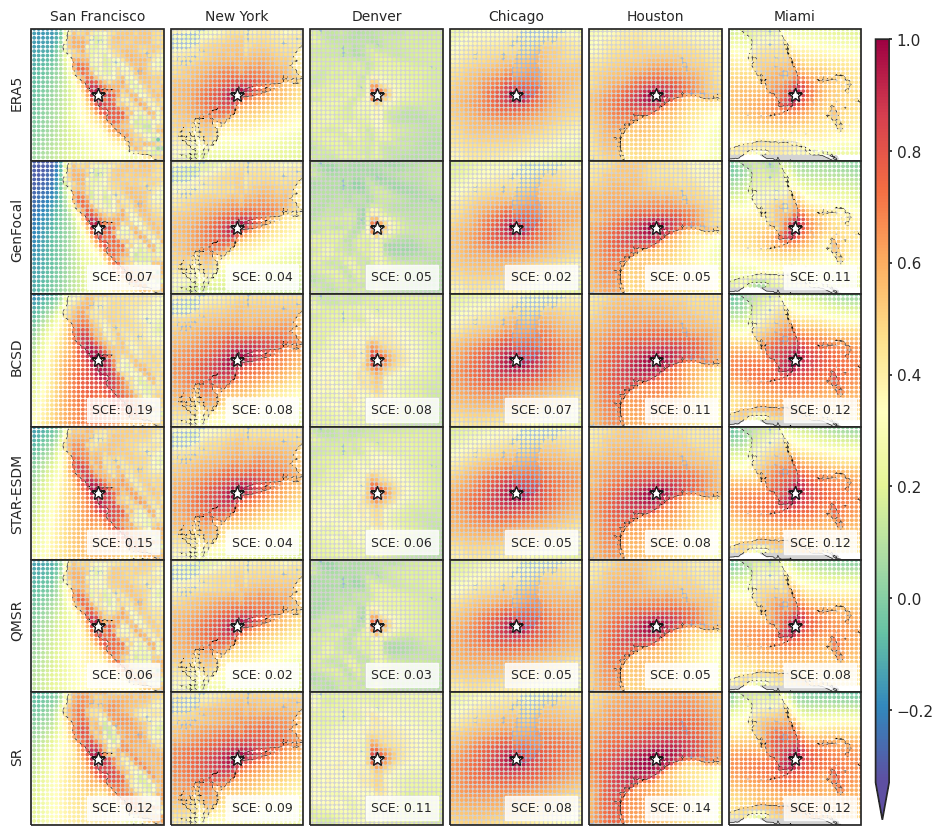}
\caption{\textbf{Spatial correlation for 10~m wind speed.} Spatial correlation for 10~m wind speed around selected populous US cities, evaluated for all snapshots at all times. The color scale represents the correlation coefficient relative to the city (stars) within a $\pm4^\circ$ longitude/latitude range.}
\label{fig:si_conus_spatial_correlations_10mW_18utc}
\end{figure}

\begin{figure}[tb]
\centering
\includegraphics[width=.7\textwidth]{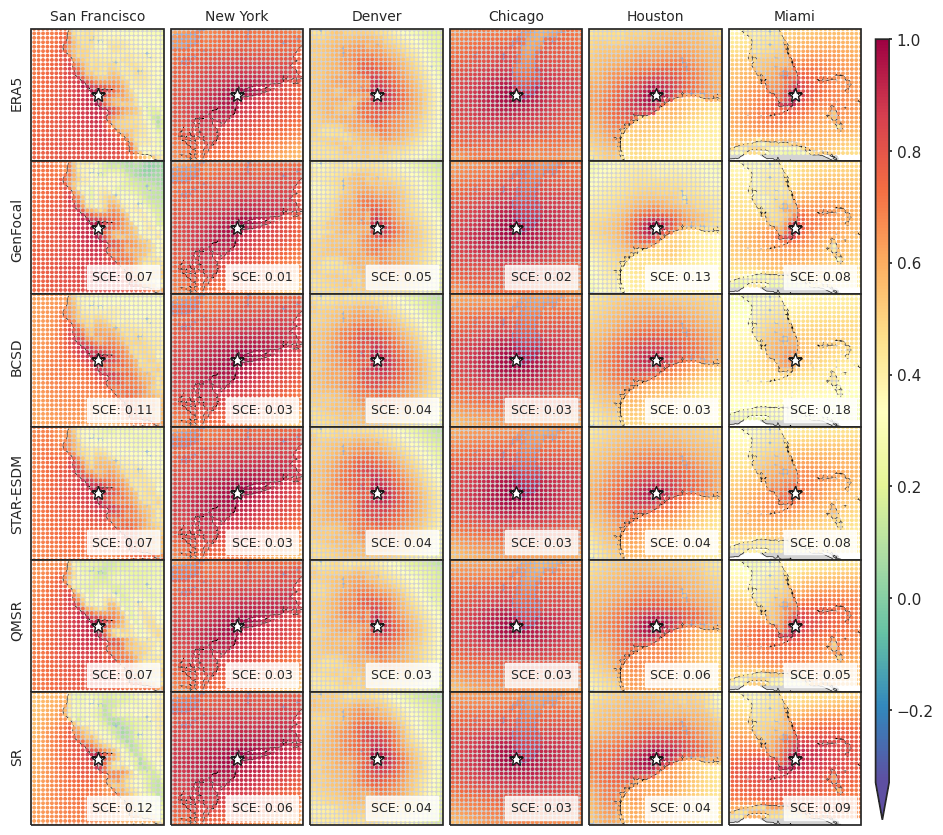}
\caption{\textbf{Spatial correlation for near-surface specific humidity.} Spatial correlation for near-surface specific humidity around selected populous US cities, evaluated for all snapshots at all times. The color scale represents the correlation coefficient relative to the city (stars) within a $\pm4^\circ$ longitude/latitude range.}
\label{fig:si_conus_spatial_correlations_q1000_18utc}
\end{figure}

\begin{figure}[tbh!]
\centering
\includegraphics[width=.7\textwidth]{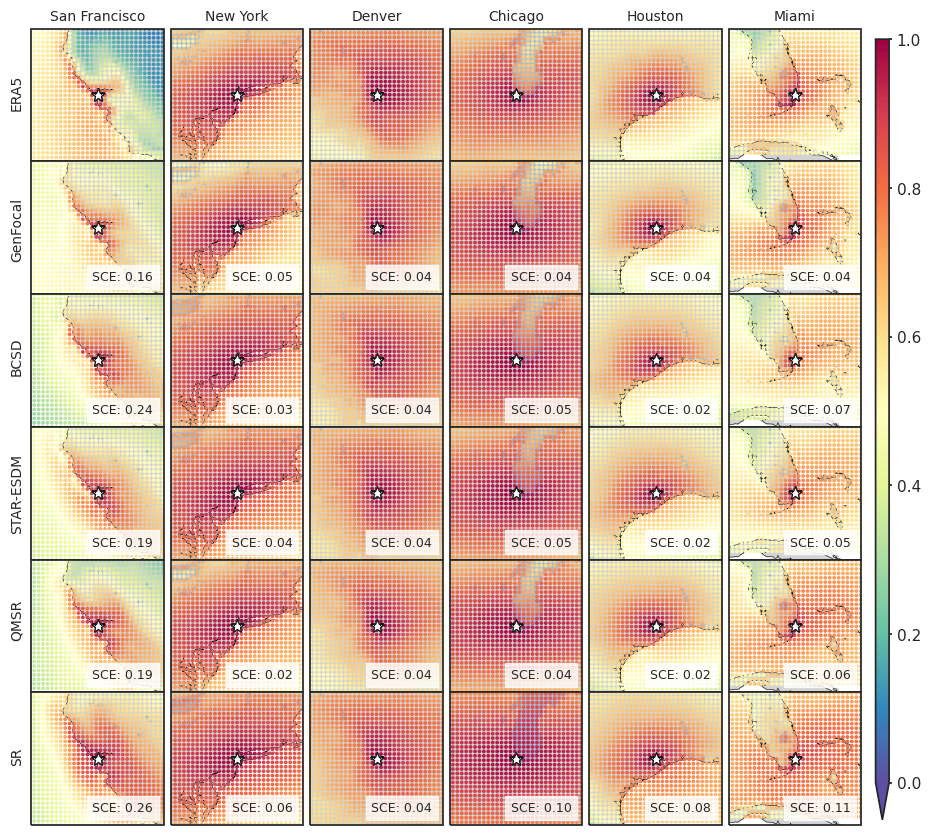}
\caption{\textbf{Spatial correlation for heat index.} Spatial correlation for heat index around selected populous US cities, evaluated for all snapshots at 18:00 UTC. The color scale represents the correlation coefficient relative to the city (stars) within a $\pm4^\circ$ longitude/latitude range.}
\label{fig:si_conus_spatial_correlations_hi_18utc}
\end{figure}

\ourname provides a more accurate representation of spatial correlations (defined in~\ref{si:correlations}), as shown in Figs.~\ref{fig:si_conus_spatial_correlations_2mT_18utc}-\ref{fig:si_conus_spatial_correlations_hi_18utc}. Furthermore, the QMSR variant achieves error levels similar to \ourname and lower than other methods, highlighting the diffusion-based super-resolution model's advantage over random historical analogs. 

\begin{figure}[tbh!]
\centering
\includegraphics[width=\textwidth]{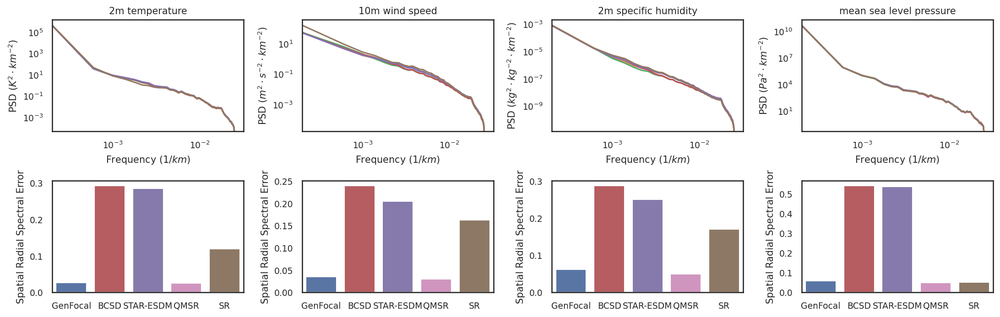}
\caption{\textbf{Spatial power spectra density.} Spatial radial power spectra density (following ~\ref{si:spatial_spectrum}), including the spectral error~\eqref{eq:spatial_psd_error}, for output variables generated with \ourname and other methods. }
\label{fig:si_conus_spatial_psd}
\end{figure}

Similar observations can be made from the spatial radial spectra in Fig.~\ref{fig:si_conus_spatial_psd}, where overall errors are lowest for \ourname and QMSR, signaling the effectiveness of diffusion-based super-resolution.

\subsection{Temporal correlations}

\begin{figure}[tbh!]
\centering
\includegraphics[width=.85\textwidth]{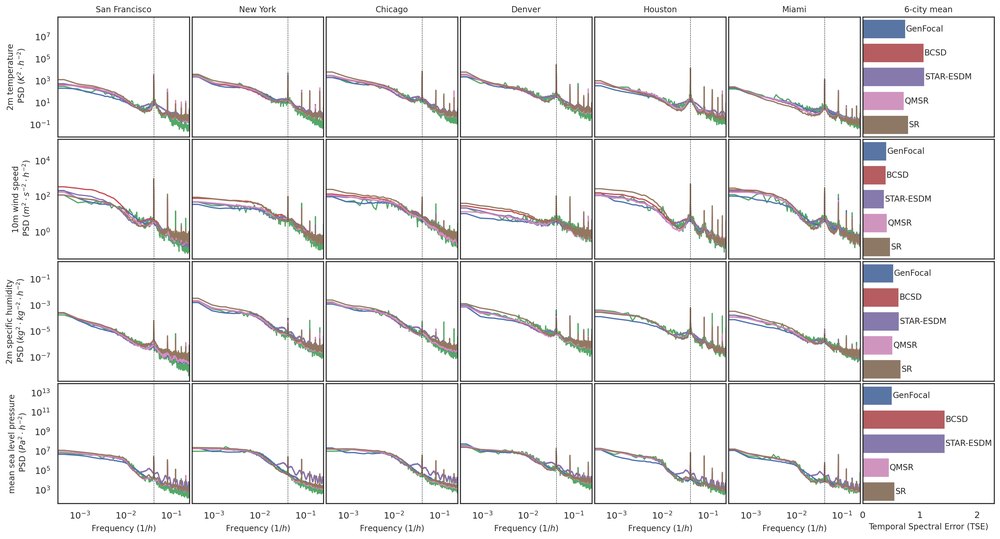}
\caption{\textbf{Temporal power spectra density.} Temporal power spectra density (following 
~\ref{si:temporal_spectrum}), including the spectral error~\eqref{eq:psd_error}, for a set of selected cities in CONUS and different variables for ensembles generated with \ourname and other methods. }
\label{fig:si_conus_temporal_psd}
\end{figure}

We also demonstrate the capacity of \ourname in capturing the temporal statistics of the directly modeled variables. Fig.~\ref{fig:si_conus_temporal_psd} shows the temporal power spectral density (following~\ref{si:temporal_spectrum}) of different variables for a set of different cities in CONUS during the evaluation period (summers in the 2010s). Overall, we observe that \ourname outperforms BCSD and STAR-ESDM in the 2m temperature and specific humidity, while remaining competitive for the 10m wind speed.

As both QMSR and SR use a similar time-coherence super-resolution approach as \ourname, they provide competitive results when compared to the disaggregation-based methods. We  observe from Fig.~\ref{fig:si_conus_temporal_psd} that QMSR also outperforms BCSD and STAR-ESDM, and in some cases is slightly better than \ourname, while SR outperforms BCSD and STAR-ESDM in all but the 10 m~wind speed case, trailing only \ourname in all the variables.

\subsection{Statistics of heat streaks}

\begin{figure}[tbh!]
\centering
\includegraphics[width=0.8\textwidth]{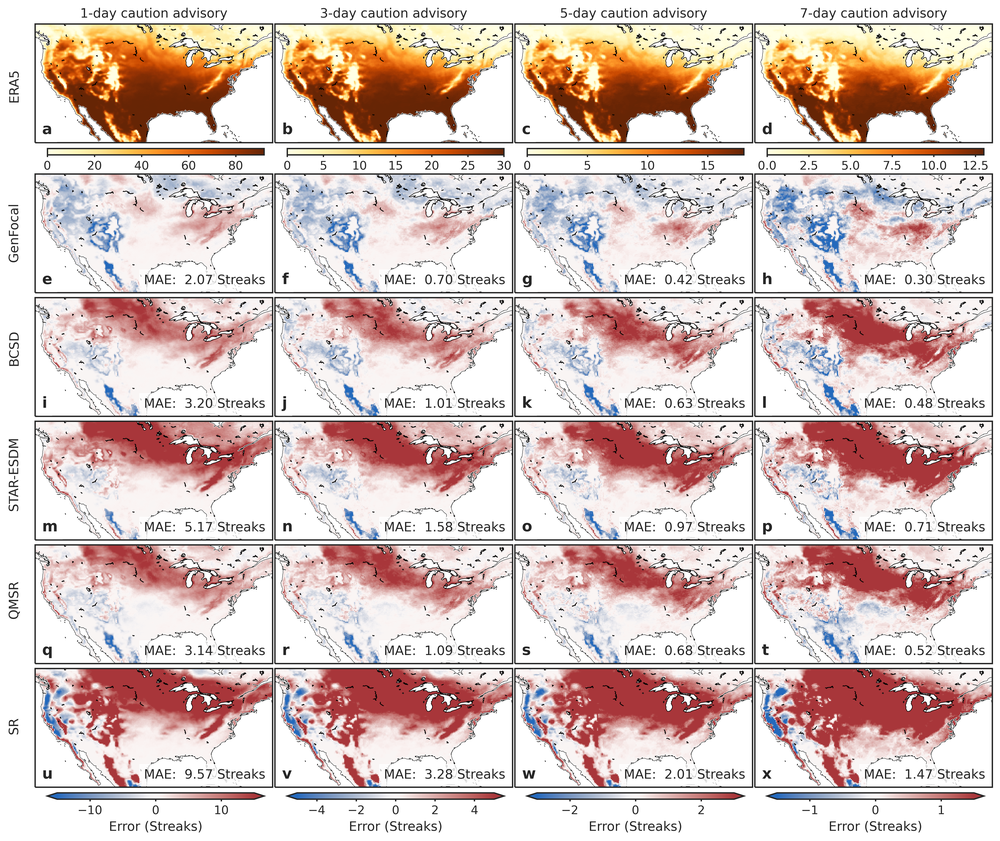}
\caption{\textbf{Bias in the number of heat streaks by different lengths.} Bias in the number of heat-streaks per year for \textbf{caution} advisory considering different lengths. We show the ground truth (ERA5)(\textbf{a-d}), and the pointwise errors of \ourname (\textbf{e-h}), BCSD (\textbf{i-l}), STAR-ESDM (\textbf{m-p}), QMSR (\textbf{q-t}) and SR  (\textbf{u-x}). 
}\label{fig:si_conus_heatwaves_caution}
\end{figure}

\begin{figure}[tbh!]
\centering
\includegraphics[width=0.8\textwidth]{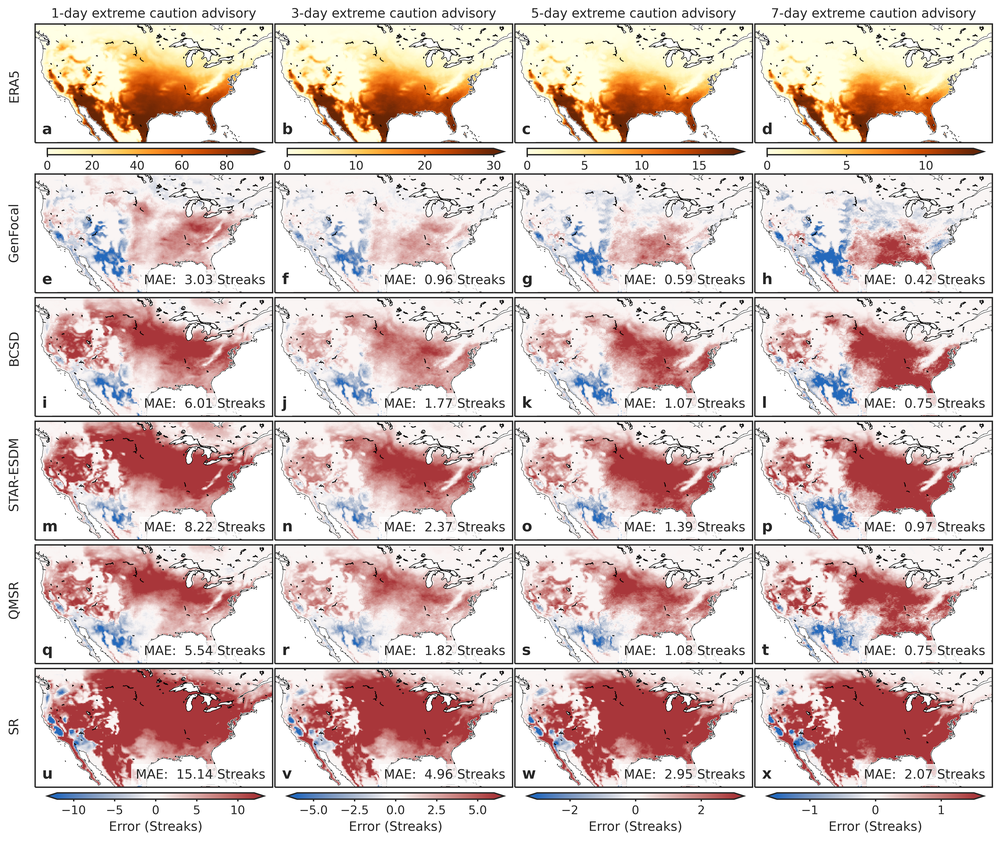}
\caption{\textbf{Bias in the number of heat streaks by different lengths.} Bias in the number of heat-streaks per year for \textbf{extreme caution} advisory considering different lengths. We show the ground truth (ERA5)(\textbf{a-d}), and the pointwise errors of \ourname (\textbf{e-h}), BCSD (\textbf{i-l}), STAR-ESDM (\textbf{m-p}), QMSR (\textbf{q-t}) and SR  (\textbf{u-x}).
}\label{fig:si_conus_heatwaves_extreme_caution}
\end{figure}

\begin{figure}[tbh!]
\centering
\includegraphics[width=0.8\textwidth]{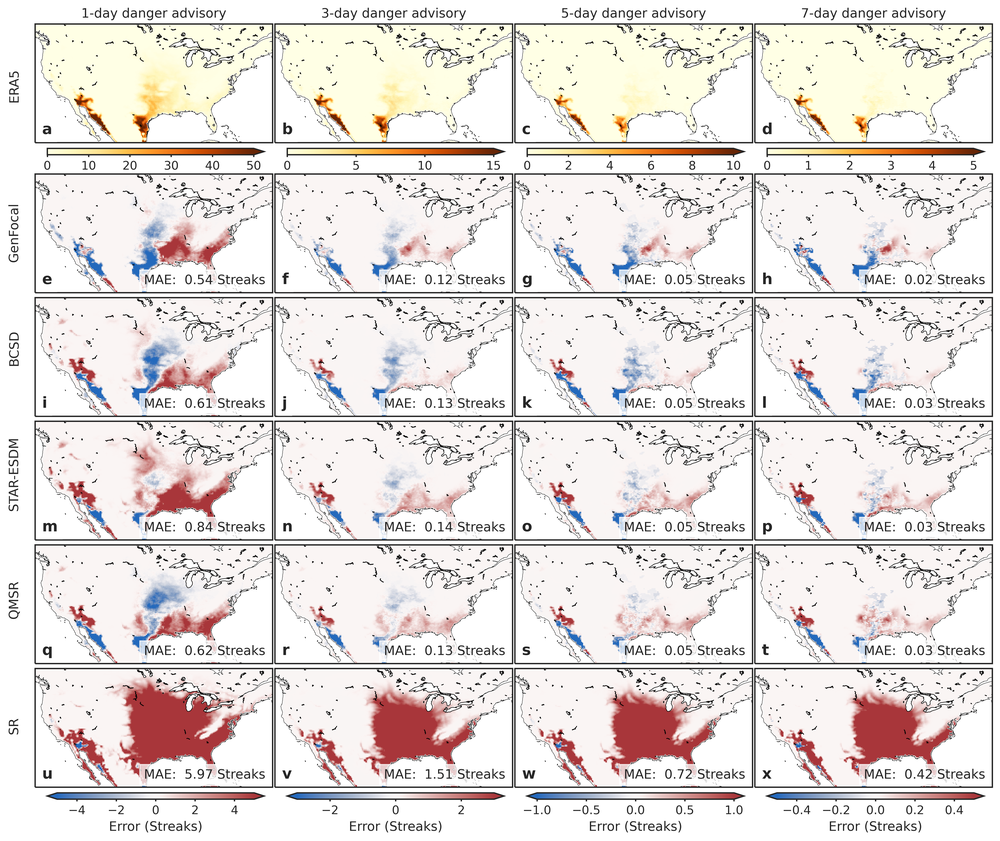}
\caption{\textbf{Bias in the number of heat streaks by different lengths.} Bias in the number of heat-streaks per year for \textbf{danger} advisory considering different heatwave lengths. We show the ground truth (ERA5)(\textbf{a-d}), and the pointwise errors of \ourname (\textbf{e-h}), BCSD (\textbf{i-l}), STAR-ESDM (\textbf{m-p}), QMSR (\textbf{q-t}) and SR  (\textbf{u-x}).
}\label{fig:si_conus_heatwaves_danger}
\end{figure}

Given \ourname\!'s superior performance in capturing temporal coherent statistics and distributions of derived variables, we further compare the heat streaks generated by the different models including the variants of \ourname introduced in~\ref{si:ablation_models}. We show the biases on the number of heat-streaks under different intensities and durations. Figs.~\ref{fig:si_conus_heatwaves_caution}--\ref{fig:si_conus_heatwaves_danger} show the local bias in the mean number of streaks per year for the increasing intensity (from ``caution'' to ``danger''). Each figure shows the bias for increasing duration (from 1 day to 7 days) for a fixed intensity. 

In general, \ourname outperforms other methods for a significant margin particularly as the intensity and duration increases. We observe that the geographical distribution of the bias is fairly similar among the methods that rely on QM for the debiasing, whereas \ourname and SR present different geographical bias patterns.

\section{Future climate risk assessment} \label{si:wus}

This section explores the application of \ourname to assess future changes in regional climate risk consistent with input coarse-scale climate projections. In particular, we analyze trends in the distribution of summer near-surface temperatures over the western U.S., and changes in tropical cyclone activities in the North Atlantic basin.

\subsection{Changes in summer temperatures over the Western U.S.}

The distribution of near-surface temperature is strongly affected by increasing atmospheric greenhouse gas concentrations. This results in significant changes in the risk of temperature extremes over time. We analyze the ability of \ourname to project these changes at a regional scale over major cities in the western U.S., focusing on periods 2017-2023 and 2077-2083.

Since observational references do not exist for future time periods, we compare \ourname projections to projections from the Western United States Dynamically Downscaled Dataset (WUS-D3)~\cite{Rahimi2024}. In particular, we evaluate the distribution changes from projections of CESM2 dynamically downscaled to 45$~\mathrm{km}$ and 9$~\mathrm{km}$ resolution using the Weather Research and Forecasting (WRF) model~\cite{skamarock2021}. We denote these projections as WRF 45$~\mathrm{km}$ and WRF 9$~\mathrm{km}$, respectively. Dynamical downscaling is performed after debiasing the CESM2 projections with respect to the ERA5 reanalysis over the historical period. Therefore, the debiasing and downscaling setup is similar to that of \ourname. Note that WUS-D3 is generated using physics-based dynamical downscaling, an expensive operation that was restricted to the western US. 

We align the grids of all projections by interpolating the \ourname and WRF 45$~\mathrm{km}$ data to the WRF 9$~\mathrm{km}$ grid. The WRF 9$~\mathrm{km}$ is averaged to a similar effective scale as \ourname by Gaussian filtering. Results are also provided for the statistical downscaling baselines BCSD and STAR-ESDM, using the same interpolation as \ourname.

\begin{figure}[tbh!]
\centering
\includegraphics[width=0.95\textwidth]{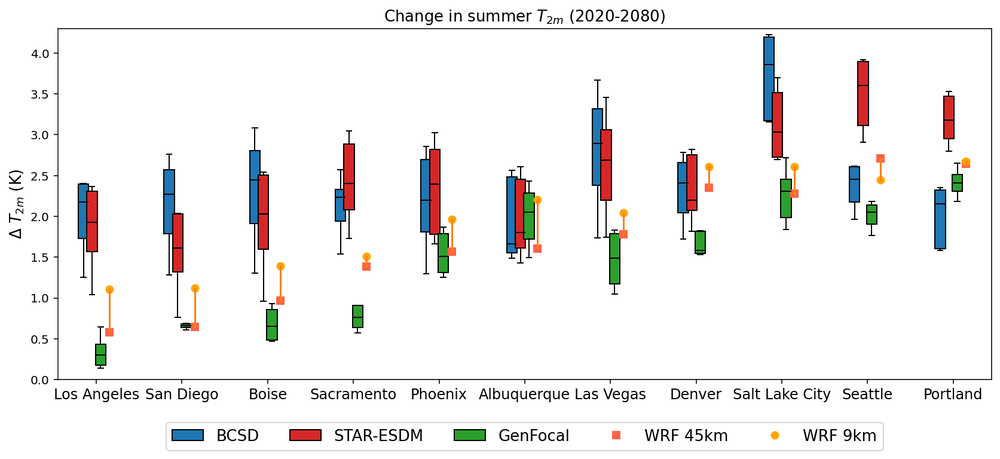}
\caption{\textbf{Projected change in daily mean near-surface temperature.} Projected change in daily mean near-surface temperature in 11 cities across the Western United States, from 2017 to 2083. Results are computed as the average over $1^\circ\times 1^\circ$ regions, and 7 summers (June-August) centered around 2020 and 2080. Boxes for BCSD, STAR-ESDM, and \ourname show the interquartile range of an ensemble of 8 projections, and whiskers represent the $12.5\%$ and $87.5\%$.
}\label{fig:si_clim_change_wus_t2m}
\end{figure}

Fig.~\ref{fig:si_clim_change_wus_t2m} illustrates changes in daily mean near-surface temperature in 11 cities across the western U.S. \ourname projects large differences in temperature changes across locales, consistent with the dynamically downscaled projections. Coastal California cities like Los Angeles or San Diego experience a much slower warming rate than inland cities Portland or Salt Lake City. BCSD and STAR-ESDM fail to capture these regional differences, projecting a much more uniform warming.

\begin{figure}[tbh!]
\centering
\includegraphics[width=0.95\textwidth]{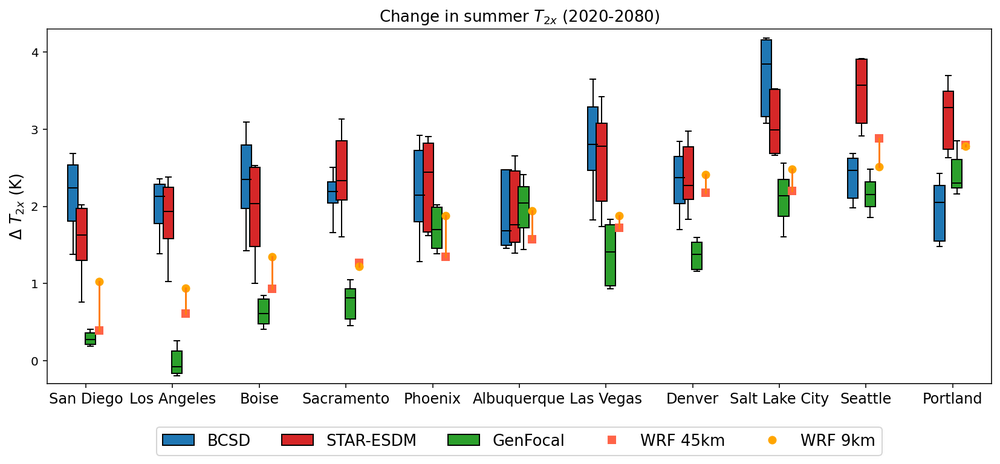}
\caption{\textbf{Projected change in daily maximum near-surface temperature.} Projected change in daily maximum near-surface temperature in 11 cities across the Western United States, from 2017 to 2083. Results are computed as the average over $1^\circ \times 1^\circ$ regions, and 7 summers (June-August) centered around 2020 and 2080. Boxes for BCSD, STAR-ESDM, and \ourname show the inter-quartile range of an ensemble of 8 projections, and whiskers represent the $12.5\%$ and $87.5\%$.
}\label{fig:si_clim_change_wus_t2x}
\end{figure}

\begin{figure}[tbh!]
\centering
\includegraphics[width=0.95\textwidth]{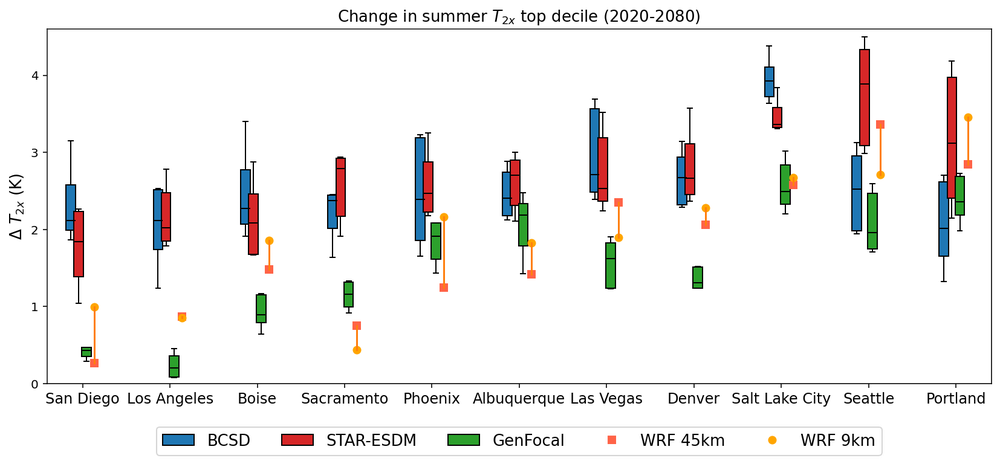}
\caption{\textbf{Projected change in the top decile of daily maximum near-surface temperature.} Projected change in the top decile of the daily maximum near-surface temperature in 11 cities across the Western United States, from 2017 to 2083. Results are computed as the average over $1^\circ \times 1^\circ$ regions, and 7 summers (June-August) centered around 2020 and 2080. Boxes for BCSD, STAR-ESDM, and \ourname show the interquartile range of an ensemble of 8 projections, and whiskers represent the $12.5\%$ and $87.5\%$.
}\label{fig:si_clim_change_wus_t2x_top_decile}
\end{figure}

\ourname not only captures changes in daily mean temperature, but also changes in summer temperature extremes. Fig.~\ref{fig:si_clim_change_wus_t2x} shows the change in daily maximum temperatures, and Fig.~\ref{fig:si_clim_change_wus_t2x_top_decile} illustrates changes in the top decile of daily maximum temperatures. The regional differences in these changes projected by \ourname and WRF are also largely consistent. BCSD and STAR-ESDM, again, show a much smaller variations among regions.

\subsection{Changes in North Atlantic tropical cyclone activity}

We assess the sensitivity of TC activity projected by \ourname to changing environmental conditions in the North Atlantic by downscaling climate projections from the early (2010-2019) through the mid (2050-2059) $21^{\text{st}}$ century under the SSP3-7.0 shared socioeconomic pathway~\cite{ONeill_2016}. The mid-century projection (2050-2059) roughly corresponds to the first decade surpassing a $2^{\circ}C$ global surface temperature change since preindustrial levels, a common warming level in climate change assessments of TC activity~\cite{Knutson2020}.

\begin{figure}[tbh!]
\centering
\includegraphics[width=0.55\textwidth]{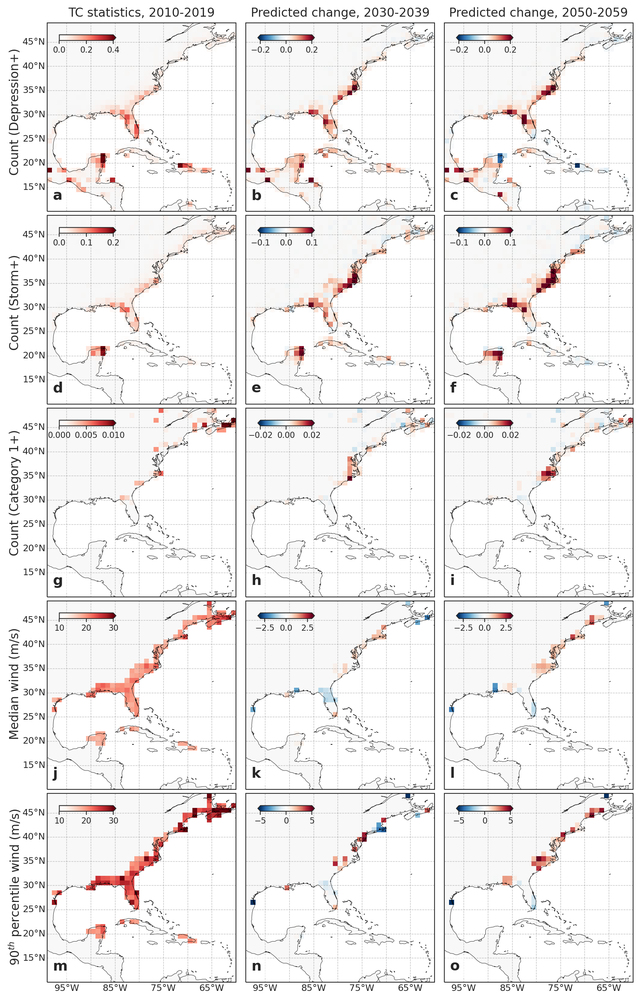}
\caption{\textbf{Change in TC landfall frequency and intensity.} Change in TC landfall frequency and intensity over the first half of the 21$^\text{th}$ century. \textbf{a}. Number of TC landfalls during the August-October season of years 2010-2019. \textbf{b, c}. Projected change in the number of TC landfalls from 2010-2019
to 2030-2039, and 2050-2059. \textbf{d-f}. Number and projected change in the number of tropical storm and hurricane landfalls. \textbf{g-i}. Number and projected change in the number of hurricane landfalls. 
\textbf{j-l}. Median maximum pressure-derived wind speed of TC landfalls and its projected change. \textbf{m-o}. $90^{\text{th}}$ percentile of maximum pressure-derived wind speed of TC landfalls and its projected change. All results are computed as the average over 800 downscaled projections. Wind speed changes (\textbf{k, l, n, o}) are only shown if statistically significant ($p < 0.05$ in a two-sided Mann-Whitney U test) and set to zero otherwise.}
\label{si:fig:tc-change}
\end{figure}

Fig.~\ref{si:fig:tc-change} evaluates North Atlantic TC activity changes from 800 downscaled projections generated with \ourname from the original 100 climate projections in the LENS2 ensemble (\ourname samples downscale 8 trajectories per ensemble member.). Changes are evaluated for decades 2030-2039 and 2050-2059 with respect to 2010-2019. \ourname projects increased TC risk over the U.S. East Coast, and particularly from the Carolinas to New Jersey, due to an increase in landfall frequency (Fig.~\ref{si:fig:tc-change}~c,f,i) and intensity (Fig.~\ref{si:fig:tc-change}~l,o).

Elevated coastal risk is already observed in the 2030-2039 projections, but becomes more pronounced by mid-century. Increases in landfall frequency are projected both for low-intensity tropical depressions, for tropical storms, and for hurricanes; although for the latter the increased risk is limited to the Carolinas and Virginia. Increases in the TC-driven winds are also found to be significant for TCs of median and high intensity between Florida and Massachusetts. Increased coastal TC risk over a similar span of the U.S. East Coast has also been projected by~\cite{balaguru_2023} using the Risk Analysis Framework for Tropical Cyclones (RAFT) model. Discrepancies between \ourname\! and RAFT, which forecasts reduced risk in the Northeast, may stem from RAFT's assumption of no change in the location of TC genesis, whereas other studies have projected a northward shift of TC genesis in the North Atlantic basin~\cite{Garner_2021}.

\clearpage 

\section{Statistical downscaling baselines} 
\label{si:baselines_and_ablations}

Most existing ML-based downscaling methods are inapplicable for climate risk assessment, as they require time-aligned data which is often unavailable (Table \ref{table:ml_downscaling_studies}; \ref{si:related}). We therefore compare \ourname to two prominent statistical techniques that do not have this requirement: the widely-used Bias Correction and Spatial Disaggregation (BCSD)~\cite{wood2002long,wood2004hydrologic,Thrasher2022} and the state-of-the-art STAR-ESDM~\cite{Hayhoe2024}, recommended for the US Fifth National Climate Assessment~\cite{osti_2202926}. As both rely on matching univariate marginals via quantile mapping, they fail to effectively model joint distributions.

To isolate the source of performance gains, we test two variants of \ourname (\ref{si:ablation_models}). First, removing the debiasing step entirely—thus treating the model as supervised under a perfect prognosis assumption (\ref{si:sr})—produces large biases and artifacts, such as generating TCs in the Sahara (Figs.~\ref{fig:si_tracks}, \ref{fig:si_conus_bias}, \ref{fig:si_conus_wass}). Second, replacing our debiasing step with quantile mapping outperforms other baselines but still falls short of \ourname. While this variant performs well on metrics insensitive to spatiotemporal coherence, the performance gap widens considerably on metrics that are, such as TC and heatwave statistics.

\ourname's superior performance in assessing risks from events with complex spatiotemporal and inter-variable dependencies—detailed in \ref{si:nao}, \ref{si:conus}, and \ref{si:ablations}—demonstrates the importance of fully probabilistic, high-dimensional modeling. We did not explore other alternatives, such as deterministic super-resolution or GAN-based models, because prior work shows they underperform for large super-resolution factors~\cite{wan2023debias}. Deterministic models tend to produce overly smooth samples by collapsing to the conditional mean~\cite{molinaro2024generative}, while GANs suffer from training instability and rapid quality degradation. These limitations are particularly relevant given our high super-resolution factor of $6\times 6$ spatially and $12$ temporally, totalling $432=6\times6 \times12$.

\subsection{Bias Correction and Spatial Disaggregation (BCSD)} \label{si:bcsd}

Bias Correction and Spatial Disaggregation (BCSD) is a widely used statistical downscaling method~\cite{Thrasher2022,Rode2021,Ortiz-Bobea}, originally designed for applications in hydrology~\cite{Wood2002}. The method consists of three main stages: bias correction, spatial disaggregation, and temporal disaggregation.

\noindent \textbf{Bias correction based on Gaussian quantile mapping.}\ The goal of this step is to map the quantile of $y$ to that of the coarse-grained observation data $y'$:
\begin{equation}
    \tilde{y}_\text{anom} = \frac{y - \texttt{clim\_mean}[y]}{\texttt{clim\_std}[y]} \cdot \texttt{clim\_std}[y'],
\end{equation}
where the climatological mean and standard deviation are calculated over the BCSD training period (1961-1999). The quantiles are computed relative to member-specific climatology. Unlike \ourname, which normalizes using the aggregated climatology of the limited set of training members (4 total), this method may favor BCSD due to better climatology estimates because of its incorporation of more training data. The competitive performance of \ourname, despite this difference, highlights its robustness.

\noindent \textbf{Spatial disaggregation.}\ In the second stage, cubic interpolation is applied to the quantile-mapped anomaly, followed by the addition of the climatological mean of the high-resolution observations:
\begin{equation}
    x_{\text{daily\_mean}} = \texttt{Interp}[\tilde{y}_\text{anom}] + \texttt{clim\_mean}[x].
\end{equation}
This step yields outputs with the desired spatial resolution, but retains the temporal resolution of the input data, which is daily.

\noindent \textbf{Temporal disaggregation.}\ The final stage involves randomly selecting a historical sample sequence from the high-resolution dataset covering the period represented by the spatially disaggregated data, in this case a day, and corresponding to the same time of the year, in this case the same day-of-year. The spatially disaggregated data is then substituted by this sequence after adjusting it to match the daily mean of the spatially disaggregated sample:
\begin{equation}
    x_\text{BCSD} = x_\text{hist\_sample} - \texttt{daily\_mean}[x_\text{hist\_sample}] + x_{\text{daily\_mean}}.
\end{equation}
This step ensures that the outputs achieve the target temporal resolution.

\subsection{Seasonal Trends and Analysis of Residuals Empirical Statistical Downscaling model (STAR-ESDM)} \label{si:star_esdm}

STAR-ESDM is a statistical downscaling method that decomposes climate fields into several components, each characterized by different timescales of variability~\cite{Hayhoe2024}. The method relies on access to high-resolution data over a reference period, which is used to correct biases in the input data. The coarse input climate field $y$ is modeled as
\begin{equation}
\label{eq:staresm_input}
    y = \tau_{y} + \texttt{clim\_mean}[y-\tau_{y}] + \Delta\texttt{clim\_mean}[y-\tau_{y}] + y_\text{anom},
\end{equation}
where $\tau_y$ is a third-order parametric fit of the long-term trend of the coarse climate field, $\texttt{clim\_mean}[y-\tau_{y}]$ is its detrended climatological mean over the reference period, $\Delta\texttt{clim\_mean}[y-\tau_{y}]$ represents the climatological mean change of the detrended field from the reference to the testing period, and $y_\text{anom}$ is the resulting residual anomaly.

STAR-ESDM downscales coarse input fields by mapping each of the components of decomposition \eqref{eq:staresm_input} to the distribution of the high-resolution reference dataset. First, the long-term trend is debiased such that its mean $m_y$ over the reference period coincides with that of the high-resolution dataset $m_x$,
\begin{equation}
    \tilde{\tau}_x = \texttt{Interp}[\tau_y - m_y] + m_x.
\end{equation}

Second, the climatological mean of the coarse field is mapped to the climatological mean of the high-resolution data, assuming that the change in climatology of the coarse data from the reference to the test period is a good approximation of the same change at high-resolution:
\begin{equation}
    \Delta\texttt{clim\_mean}[x-\tau_{x}] \approx \texttt{Interp}[\Delta\texttt{clim\_mean}[y-\tau_{y}]].
\end{equation}

Finally, the coarse anomaly $y_\text{anom}$ is mapped to the distribution of the high-resolution climate data, using again the climate change in the coarse data as a proxy for the climate change in the high-resolution data,
\begin{equation}\label{si:eq_x_anom}
    \tilde{x}_\text{anom} = \texttt{Interp}[y_\text{quant}] \cdot \texttt{clim\_std}[x]\cdot\frac{\texttt{clim\_std}[y-\tau_{y}] + \Delta\texttt{clim\_std}[y-\tau_{y}]}{\texttt{clim\_std}[y-\tau_{y}]},
\end{equation}
where $\Delta\texttt{clim\_std}[y-\tau_{y}]$ is the difference in climatological standard deviation of the coarse climate data between the test period and the reference period, and the quantile of the coarse anomaly is computed with respect to the modified climatology,
\begin{equation}
    y_\text{quant} = \frac{y_\text{anom}}{\texttt{clim\_std}[y-\tau_{y}] + \Delta\texttt{clim\_std}[y-\tau_{y}]}.
\end{equation}
In equation \eqref{si:eq_x_anom}, $\texttt{clim\_std}[x]$ is the climatological standard deviation of the high-resolution dataset over the reference period. The STAR-ESDM downscaled climate field is then constructed as
\begin{equation}
    x_\text{STAR} = \tilde{\tau}_x + \texttt{clim\_mean}[x-\tau_{x}] + \Delta\texttt{clim\_mean}[x-\tau_{x}] + \tilde{x}_\text{anom}.
\end{equation}

\clearpage

\section{Data}
\label{si:data}
\subsection{Input datasets}
We use the Community Earth System Model Large Ensemble (LENS2) dataset~\cite{LENS2} for our low-resolution climate dataset. LENS2 was produced using the Community Earth System Model Version 2 (CESM2), a climate model that has interactive atmospheric, land, ocean, and sea-ice models~\cite{Danabasoglu2020-uj}. LENS2 is configured to estimate historical climate and the future climate scenario SSP3-7.0, following the CMIP6 protocol~\cite{Eyring2016-yu}. LENS2 skillfully represents the response of historical climate to external forcings~\cite{Fasullo2024-bk}. LENS2 output is available from 1850-2100, with a horizontal grid spacing of 1$^\circ$, and 100 simulation realizations. In this work, we use a coarse-grained version of the LENS2 ensemble at 1.5$^\circ$ horizontal resolution.

The ERA5 reanalysis dataset~\cite{hersbach2020era5} is our high-resolution weather dataset.  ERA5 uses a modern forecast model and data assimilation system with all available weather observations to produce an estimate of the atmospheric state. This estimate includes conditions ranging from the surface to the stratosphere. ERA5 data is available from 1940 to near present at a horizontal grid spacing of 0.25$^\circ$. ERA5 estimated extremes of temperature and precipitation agree well with observations in areas where topography changes slowly~\cite{Soci2024-gw}.  

\subsection{Modeled variables}
\label{si:modeled_vars}
\begin{table}[t!]
\centering
\footnotesize
\caption{Meteorological fields modeled by \ourname with their corresponding variable names in the ERA5 and LENS2 datasets. All 10 fields serve as both input and output for the debiasing model, while the super-resolution model uses the top 4 fields in CONUS and top 6 fields in the North Atlantic. Units reflect those used for model training and are converted as needed in the main text.}
\label{table:modeled_vars_debiasing}
\begin{tabular}{lccc}
\hline
Meteorological field                    & Unit  & ERA5 variable                                                 & LENS2 variable \\ \hline
Near-surface temperature          & K     & 2m\_temperature                                           & TREFHT    \\
Near-surface wind speed magnitude & m/s   & $\left( \text{u\_component\_of\_wind}^2 + \right . $ & WSPDSRFAV \\
&& $ \left . \text{v\_component\_of\_wind}^2 \right)^{\frac{1}{2}}$ \\
Near-surface specific humidity    & kg/kg & specific\_humidity                   & QREFHT    \\
& & (level=1000 hPa)     \\
Sea level pressure           & Pa    & mean\_sea\_level\_pressure                                & PSL       \\ \hline
Geopotential at 200 hPa     & m    & geopotential                               & Z200       \\
& & (level=200 hPa)   \\
Geopotential at 500 hPa     & m    &                               & Z500       \\
& & geopotential (level=500 hPa)  
\\ \hline
U component of wind at 200 hPa   & m/s    & $\text{u\_component\_of\_wind}$                            & U200       \\ 
& & (level=200 hPa)     & \\
U component of wind at 850 hPa   & m/s    & $\text{u\_component\_of\_wind}$                               & U850 
\\
& & (level=850 hPa)   &\\ \hline
V component of wind at 200 hPa   & m/s    & $\text{v\_component\_of\_wind}$                              & V200       \\ 
& & (level=200 hPa)   \\
V component of wind at 850 hPa   & m/s    & $\text{v\_component\_of\_wind}$                               & V850    \\
& & (level=850 hPa)  \\ \hline
\end{tabular}
\end{table}

We consider a set of four surface variables to downscale, which were chosen in order to evaluate the statistics of the spatiotemporal events of interest, namely heat-streaks and TCs. 

The two-step nature of \ourname renders it highly versatile, as the debiasing step and the super-resolution steps are decoupled. This allows for some interesting properties, e.g., the debiasing step can be performed globally, while the super-resolution can be performed within different regions, and the meteorological fields downscaled can also be different, provided that the fields in the super-resolution step are a subset of the debiased ones.

We showcase these two properties by downscaling climate data over the North Atlantic and over CONUS (see \ref{si:nao} and \ref{si:conus} respectively), and by using different variables for the debiasing and the super-resolution steps. In what follows we show the variables used for each step with their names and units.  

\subsubsection{Debiasing}
As shown in \ref{si:sec:debiasing-variables}, modeling extra variables in the debiasing steps results in improved results, particularly for TC tracking (see \ref{si:sec:debiasing-variables}). As such, we explicitly model 10 variables in the debiasing step. We include 4 surface variables: near-surface temperature, wind speed magnitude, specific humidity, and sea level pressure; and 6 variables within the mid or upper troposphere: geopotential at $200$ and $500$ hPa, and both components of the wind speed at $200$ and $500$ hPa. The official names for these variables, as documented in the datasets, are listed in Table~\ref{table:modeled_vars_debiasing}.

Although we do not super-resolve the above-surface variables, they provide extra signal for the debiasing step, as they are correlated with some of the near-surface variables.

\subsubsection{Super-resolution}
We target four surface variables in our downscaling pipeline: near-surface temperature, wind speed magnitude, specific humidity, and sea level pressure, which constitute a subset of the debiased variables (top 4 rows in Table~\ref{table:modeled_vars_debiasing}). In CONUS, these variables coincide with the modeled variables (both input and output). In North Atlantic, we additionally include two geopotential fields, at 200 and 500~hPa respectively, in the super-resolution model.

\subsection{Regridding}
\label{si:regridding}
The ERA5 dataset is natively 0.25$^\circ$ and LENS2 is 1$^\circ$. Here we use linear interpolation to regrid the data to 1.5$^\circ$ using the underlying spherical geometry of the data, instead of performing interpolation in the lat-lon coordinates. We additionally compute daily averages of the ERA5 data to match the temporal resolution of LENS2 in the debiasing process.

\clearpage
\section{Evaluation metrics} \label{si:evaluation_metrics}

This section details the various metrics employed to assess statistical accuracy. In particular, we focus on measuring the marginals (i.e. pointwise distribution errors), such as bias, Wasserstein distance and extreme quantiles. Additionally, we incorporate metrics that account for correlations across space, time and fields. 
    
For completeness, the trajectory in the downscaled ensemble is represented as a five-tensor: 
\begin{equation}
    x_{i, j, t, f, m},
\end{equation}
where the $i,j$ indices account for the space (latitude and longitude), $t$ for the time, $f$ for the different fields (or variables), and $m$ for the members in the ensemble. The reference data from ERA5 reanalysis shares the same structure but lacks the member index, and is denoted as $x_{i,j,t,f}^{\text{ref}}$.

While most metrics involve temporal aggregation over the evaluation period, the time index can sometimes be further decomposed into three components $t = (t_h, t_d, t_y)$, representing hour, day-of-the-year, and year indices. This decomposition is commonly used in climatological computations, where each sub-index is contracted differently. In this section, however, it is only applied to the computation of the tail dependence, requiring special attention to avoid evaluations dominated by the diurnal cycle.

\subsection{Pointwise distribution errors} \label{si:pointwise}

The following metrics measure the distribution difference between the predicted samples concatenated into a 5-tensor $x$, and the reference samples concatenated into a 4-tensor $x^\text{ref}$, where $x \in \mathbb{R}^{N_{\text{lat}} \times N_{\text{lon}} \times N_t \times N_f \times N_{\text{ens}}}$ and $ x^\text{ref} \in \mathbb{R}^{N_{\text{lat}} \times N_{\text{lon}} \times N_t \times N_f}$.
Here $N_f=4$ (or $6$ when considering the derived variables in ~\ref{si:derived}), $N_m$ is $100$ for LENS2 (see \ref{si:debiasing_data_set}), and $800$ for BCSD, STAR-ESDM, QMSR, SR, and \ourname (each LENS2 member yields $8$ new downscaled samples).

\subsubsection{Mean absolute bias (MAB)} \label{si:bias}
We define the bias as the difference between the ensemble mean of the point-wise distributions

\begin{equation}
    \text{Bias}_{i,j,f} = \frac{1}{N_t}   \left [  \frac{1}{N_m}\sum_{m, t} x_{i, j, t, f, m}  - \sum_{t} x_{i,j, t, f} \right ]
\end{equation}
where $t$ covers the period under consideration, e.g., summer (June-July-August) during the evaluated years. The bias for different variables is
plotted in Figs.~\ref{fig:si_conus_bias}, \ref{fig:si_conus_rh}, and \ref{fig:si_conus_hi} over CONUS.

The mean absolute bias is defined as the spatial average f the absolute bias, 
\begin{equation} \label{si:eq:mab_definition}
    \text{MAB}_f = \frac{1}{N_{\text{lon}} N_{\text{lat}}} \sum_{i, j} \left| \,\text{Bias}_{i,j,f} \, \right|.
\end{equation}
This quantity is reported in Table \ref{table:bias_and_wass} for the directly modeled variables, and in 
Table \ref{table:bias_and_wass_composite} for the derived variables. The MAB is also reported in the annotations in Figs.~\ref{fig:si_conus_bias}, \ref{fig:si_conus_rh}, and \ref{fig:si_conus_hi}.

\subsubsection{Mean Wasserstein distance (MWD)} \label{si:wass}

The Wasserstein-1 metric for each location represents the $L^1$ norm between the predicted and reference distributions.

Algorithmically, this metric involves constructing empirical cumulative distribution functions CDF and $\text{CDF}^\text{ref}$ for the predicted and reference samples respectively. For the first we aggregate both in time and ensemble, ($t$ and $m$ indices), and for the second we only aggregate in time. We can write this data dependency as 
\begin{equation}
    x_{i,j,:,f,:} \rightarrow \text{CDF}_{i,j,f}(\cdot) \qquad x^{\text{ref}}_{i,j,:,f} \rightarrow \text{CDF}^\text{ref}_{i,j, f}(\cdot),
\end{equation}
where the $m$-index is aggregated for the $800$ ensemble members, and the $t$ is aggregated during the evaluation period. 

Then the pointwise Wasserstein distance is computed 
\begin{equation}
    \text{WD}_{i,j,f} = \sum_{q=1} \left|\text{CDF}_{i, j, f}(x_{q}) - \text{CDF}_{i, j, f}^\text{ref}(x_{q})\right| \omega_q,
\end{equation}
where $x_q$ are the quadrature points over which the integrand is evaluated, and are chosen to cover the union of the support for both predicted and reference distributions; and $\omega_q$ are the quadrature weights, which in this case are defined by $\omega_q: = x_{q+1} - x_{q}$. This quantity is shown for different variables in Fig.~\ref{fig:si_conus_wass}.

The (spatially averaged) Mean Wasserstein distance (MWD) as reported in Tables \ref{table:bias_and_wass} and \ref{table:bias_and_wass_composite} is then computed as:
\begin{equation}
    \text{MWD}_f = \frac{1}{N_{\text{lon}}\cdot N_{\text{lat}} }\sum_{i,j}  \text{WD}_{i,j,f}.
\end{equation}

\subsubsection{Percentile mean absolute error (MAE)} \label{si:percentiles}

This metric measures the mean absolute difference between the $p^\text{th}$ percentiles of the predicted and reference samples.
For each $i,j$ coordinate and each $f$ field, we aggregate over the member and time indices to create histograms from which the percentiles are computed. For the reference data, we only aggregate over the time index. We use \texttt{numpy.percentile} function (abbreviated to \texttt{Pctl}) with different data following
\begin{equation}
    x_{i,j,:,f, :} \rightarrow \texttt{Pctl}_{i, j, f}(\cdot) \qquad x^{\text{ref}}_{i,j,:,f} \rightarrow \texttt{Pctl}^\text{ref}_{i,j, f}(\cdot).
\end{equation}
We define the pointwise percentile error of the $p^\text{th}$ percentile as
\begin{equation}
    \text{AE}_{i,j,f}(p) = \left|\texttt{Pctl}_{i,j,f}(p) - \texttt{Pctl}_{i,j,f}^\text{ref}(p)\right|.
\end{equation}
This is the quantity shown in Figs.~\ref{fig:si_conus_95}, \ref{fig:si_conus_99}, \ref{fig:si_conus_rh}, and \ref{fig:si_conus_hi}. We also consider a spatially averaged quantity for each field given by
\begin{equation}
    \text{MAE}_{f}(p) = \frac{1}{N_{\text{lon}} N_{\text{lat}}} \sum_{i,j} \text{MAE}_{i,j,f}(p).
\end{equation}
This is the quantity reported in Table \ref{table:bias_and_wass}.

\subsection{Correlations} \label{si:correlations}

\subsubsection{Spatial correlation} \label{si:spatial_correlations}

For a given target location given by indices $i,j$ and a nearby location $k, l$, we first compute their sample means following
\begin{equation}
    \bar{x}_{i,j,f} = \frac{1}{N_{\text{ens}} N_t} \sum_{t, m} x_{i,j,t,f,m},  \qquad \text{and}  \qquad \bar{x}_{k,l,f} = \frac{1}{N_{\text{ens}} N_t} \sum_{t, m} x_{k,l,t,f,m},
\end{equation}
which allows us to compute the correlation between locations $(i,j)$ and $(k, l)$ as
\begin{equation}
    \rho_{ij, kl, f} = \frac{\sum_{t,m} \left(x_{i,j,t,f,m} - \bar{x}_{i,j,f}\right)\left(x_{k,l,t,f,m} - \bar{x}_{k,l,f}\right)}{\sqrt{\sum_{t,m}\left(x_{i,j,t,f,m} - \bar{x}_{i,j,f}\right)^2}\sqrt{\sum_{t,m}\left(x_{k,l,t,f,m} - \bar{x}_{k,l,f}\right)^2}}.
\end{equation}

The reference correlation is computed similarly but without aggregation in the member index, i.e., 
\begin{equation}
    \bar{x}^{\text{ref}}_{i,j,f} = \frac{1}{N_t} \sum_{t} x^{\text{ref}}_{i,j,t,f}, 
\end{equation}
\begin{equation}
    \rho^{\text{ref}}_{ij, kl, f} = \frac{\sum_{t} \left(x^{\text{ref}}_{i,j,t,f} - \bar{x}^{\text{ref}}_{i,j,f}\right)\left(x^{\text{ref}}_{k,l,t,f} - \bar{x}^{\text{ref}}_{k,l,f}\right)}{\sqrt{\sum_{t}\left(x^{\text{ref}}_{i,j,t,f} - \bar{x}^{\text{ref}}_{i,j,f}\right)^2}\sqrt{\sum_{t}\left(x^{\text{ref}}_{k,l,t,f} - \bar{x}^{\text{ref}}_{k,l,f}\right)^2}}.
\end{equation}

Computing the correlation coefficient across all nearby locations within a selected range yields the correlation matrix $P_{ij,f} = \{\rho_{ij, kl, f }\}$. This matrix is shown in the plots in ~\ref{si:sec:conus_spatial_correlation} and in Figs.~\ref{fig:conus}d-e.
Then we compute the pointwise spatial correlation error (SCE) as
\begin{equation}
 \text{SCE}_{ij,kl, f} = \left| \rho_{ij, kl, f } - \rho^{\text{ref}}_{ij, kl, f}\right|,
\end{equation}
which is shown in Figs.~\ref{fig:conus}({i}, {j}, {n}, {o}, {s}, and {t}).

Finally, the SCE is then quantified using the $\ell^1$ norm as a flattened vector between the predicted and reference correlation matrices:
\begin{equation}
    \text{SCE}_{ij, f} = \|P - P_\text{ref}\|_{\ell^1} = \frac{1}{N_k N_l} \sum_{k,l}  \left| \rho_{ij, kl, f } - \rho^{\text{ref}}_{ij, kl, f}\right|.
\end{equation}
This is the metric shown in all the plots of ~\ref{si:sec:conus_spatial_correlation}.

\subsubsection{Spatial spectrum} \label{si:spatial_spectrum}

Spatial structure can be analyzed through the power spectral density (PSD). The outputs are first transformed to frequency domain via the 2-dimensional Discrete Fourier Transform (DFT):
\begin{equation}
    x_{:,:,t,f,m}\rightarrow X_{t,f,m}(\cdot, \cdot),
\end{equation}
where $X$ denotes the Fourier coefficients. The energy of a frequency component $(\xi_k, \xi_l)$ is given by
\begin{equation}
    \Phi_{t,f,m}(\xi_k, \xi_l) = \frac{1}{A}\left|X_{t,f,m}(\xi_k, \xi_l)\right|^2,
\end{equation}
where $A$ represents the area of the region (approximated as a rectangle) over which the spectrum is computed. The 2-dimensional spectrum is converted into a 1-dimension radial spectrum by binning along radial frequency $\xi_r = \sqrt{\xi_k^2 + \xi_l^2}$ and summing the frequency components within each bin
\begin{equation}
    \tilde{\Phi}_{t,f,m}(\xi_r) = \sum_{\sqrt{\xi^2_k + \xi^2_l}\in[\xi_r-\Delta\xi_r, \xi_r+\Delta\xi_r]}\Phi_{t,f,m}(\xi_k, \xi_l).
\end{equation}

The spatial radial spectral error (SRSE) between the predicted and reference spectra is computed by first averaging along time and ensemble dimension, taking the absolute difference in their logarithms and averaging across frequencies
\begin{equation} \label{eq:spatial_psd_error}
    \text{SRSE}_{f} = \frac{1}{N_{\xi_r}}\sum_{\xi_r}\left|\frac{1}{N_tN_\text{ens}}\sum\log{\tilde{\Phi}_{t,f,m}} - \frac{1}{N_t}\sum\log{\tilde{\Phi}^{\text{ref}}_{t,f}}\right|,
\end{equation}
where $N_{\xi_r}$ denotes the number of radial frequency bins. The average spectra and errors are shown in Fig.~\ref{fig:si_conus_spatial_psd}.

\subsubsection{Temporal spectrum} \label{si:temporal_spectrum}
Temporal correlations in the output can be similarly analyzed through the PSD. The outputs are first transformed to the frequency space via the 1-dimensional DFT in time:
\begin{equation}
    x_{i,j,:,f,m}\rightarrow X_{i,j,f,m}(\cdot),
\end{equation}
with corresponding energy
\begin{equation}
    \Phi_{i,j,f,m}(\xi_s) = \frac{1}{T}|X_{i,j,f,m}(\xi_s)|^2,
\end{equation}
where $T$ represents the length of the time series, $\xi_s$ is the $s$th frequency component. 
The temporal spectral error (TSE) between the predicted and reference spectra is quantified by the mean log ratio difference:
\begin{equation} \label{eq:psd_error}
    \text{TSE}_{i,j,f} = \frac{1}{N_{\xi_{s}}}\sum\limits_{\xi_s} \left|\frac{1}{N_\text{ens}}\sum\limits_{m}\log{\Phi_{i,j,f,m}(\xi_s)} - \log{\Phi^{\text{ref}}_{i,j,f}(\xi_s)}\right|,
\end{equation}
where $N_{\xi_{s}}$ denotes the number of frequency components considered in the temporal DFT. We aggregate the error over spatial dimensions
\begin{equation} \label{eq:psd_error_agg}
    \text{TSE}_{f} = \frac{1}{N_{\text{lon}} N_{\text{lat}}} \sum_{i,j} \text{TSE}_{i,j,f},
\end{equation}
which are shown in the last column of Fig.~\ref{fig:si_conus_temporal_psd}.

\subsection{Tail dependence} \label{si:tail_dependence}
We evaluate the correlation of extremes of climate fields $f$ and $g$ through the tail dependence, estimated non-parametrically following Schmidt and Stadtmüller \cite{tail_dep}. We start by computing the percentiles for both variables
\begin{equation} \label{eq:tail_dep_p}
    x_{i,j,:,f,:} \rightarrow \texttt{Pctl}_{i,j,f}(\cdot) \qquad x_{i,j,:,g,:} \rightarrow \texttt{Pctl}_{i,j,g}(\cdot),
\end{equation}
and the co-occurrence of both variables exceeding a certain percentile
\begin{equation} \label{eq:tail_dep_lambda}
    \Lambda_{i,j,fg}(p) = \frac{100}{N_\text{ens}N_t \cdot p}\sum_{t,m} \mathbf{1}_{[(x_{i,j,t,f,m} > \texttt{Pctl}_{i,j,f}(p)) \land (x_{i,j,t,g,m} > \texttt{Pctl}_{i,j,g}(p))]},
\end{equation}
where $\mathbf{1}_S$ is the indicator function that evaluates to 1 or 0 depending on whether the logical expression $S$ is true or not.
Drawing upon the homogeneity property of tail copulae \cite{tail_dep}, we compute the tail dependence by averaging over a list (length $N_p$) of threshold percentiles evenly spaced in the range $[90, 95]$:
\begin{equation}
    \tilde{\Lambda}_{i,j,fg} = \frac{1}{N_p}\sum_{p\in[90, 95]}\Lambda_{i,j,fg}(p).
\end{equation}

The tail dependence for the reference data is computed in a similar fashion: the only difference is the exclusion of ensemble index $m$ in~\eqref{eq:tail_dep_p} and~\eqref{eq:tail_dep_lambda}. The tail dependence error (TDE) is taken as the absolute difference with the corresponding reference tail dependence
\begin{equation}
    \text{TDE}_{i,j,fg} = \left|\tilde{\Lambda}_{i,j,fg} - \tilde{\Lambda}_{i,j,fg}^{\text{ref}}\right|,
\end{equation}
and optionally aggregated over spatial dimensions
\begin{equation}
    \text{TDE}_{fg} = \frac{1}{N_\text{lon}N_\text{lat}}\sum_{i,j} \text{TDE}_{i,j,fg}.
\end{equation}

This metric is reported in Figs.~\ref{fig:conus} and \ref{fig:si_conus_tails}.
Note that the tail dependence for both the upper and lower extremes can be readily assessed by negating the involved variables accordingly. For instance, we may apply transformation $g \rightarrow -g$ to evaluate the dependence of high percentiles of $f$ and low percentiles of $g$.

\section{Evaluation protocol} \label{si:evaluation_protocol}

In this section we describe how the derived variables are computed from the \ourname outputs defined in Sec. \ref{si:modeled_vars}, and how spatiotemporal events of interest are defined and detected, particularly heat streaks in Sec.~\ref{si:heat_streaks} and TCs in Sec.~\ref{si:definition_tc}. For the latter phenomena we also describe how the detection and calibration are performed.

\subsection{Derived variables} \label{si:derived}
Here we describe how the derived variables are calculated. In addition to the explicitly modeled variables, we utilize surface elevation, a static quantity, to convert sea level pressure to pressure at surface height.

\textbf{Relative humidity.}
To calculate the near-surface relative humidity, we first compute the pressure at surface height $z_s$ using the barometric formula
\begin{equation} \label{eq:barotropic}
    P = P_0\cdot\left(1 + \frac{\Gamma \cdot z_s}{T_\text{ref}}\right)^{-\frac{g\cdot M}{R\cdot\Gamma}},
\end{equation}
where $P_0$ denotes the sea level pressure (Pa), $T_\text{ref}=288.15$ is the reference surface temperature (K), $\Gamma$ is the standard tropospheric lapse rate for temperature (-0.0065~K/m), $M$ is the molar mass of air (0.02896~kg/mol), $g$ and $R$ are the gravitational acceleration (9.8~m/$\text{s}^2$) and universal gas constant (8.31447~J/mol/K) respectively.

Next we compute the saturation vapor pressure using the Clausius–Clapeyron relation \cite{Duarte14a}
\begin{equation} \label{eq:saturation_pressure}
    e_s(T) = P_\text{trip}\left(\dfrac{T}{T_\text{trip}}\right)^\alpha \exp{\left[\beta_v\left(\dfrac{1}{T_\text{trip}}-\dfrac{1}{T}\right)\right]}, 
\end{equation}
where $P_\text{trip}=611$ Pa, $T_\text{trip}=273.15$ K, $T$ is the temperature at 2 meters, $\beta_v=6773.38$ K, and $\alpha=-4.98$. Finally, we compute the actual vapor pressure as
\begin{equation}
    e=\frac{q\cdot P}{\epsilon + (1-\epsilon)\cdot q},
\end{equation}
where $q$ denotes the near-surface specific humidity in kg/kg, and $\epsilon=0.622$. Finally, the relative humidity is expressed as the ratio of actual vapor pressure to the saturation vapor pressure. Written as a percentage:
\begin{equation}
    RH = \frac{e}{e_s} \cdot 100.
\end{equation}

\textbf{Heat index.} The heat index quantifies the perceived temperature by modeling the human body's thermoregulatory response to combined air temperature and humidity. It was initially introduced by Steadman~\cite{Steadman_1979} for a range of moderate temperatures and humidities, and recently extended to all combinations of the aforementioned variables by Lu and Romps~\cite{Lu_Romps_2022}. In this work, we use the full extension of Lu and Romps, which yields the heat index as the solution of a system of algebraic equations defined by the temperature and relative humidity. A numerical solution to this system, which requires using an iterative solver, is provided in Appendix A of their original work~\cite{Lu_Romps_2022}.

As a cautionary tale, we note that we initially followed NOAA's heat index definition, based on a polynomial extrapolation of Steadman's model~\cite{heat_index}. Using this polynomial fit to estimate summer extremes in the heat index across the continental U.S. yielded unrealistically high values of the heat index over the Rockies, the Sierra Nevada, the Cascades, and the Great Lakes. This stresses the importance of using the full definition of the heat index to explore extremes, even in the current climate~\cite{Romps2022}.

\subsection{Heat streaks} \label{si:heat_streaks}

NOAA provides 4 advisory levels based on the heat index: caution, extreme caution, danger and extreme danger; triggered by heat index values exceeding $80 ^\circ\text{F}$, $90 ^\circ\text{F}$, $103 ^\circ\text{F}$ and $125 ^\circ\text{F}$ (300K, 305K, 312.6K, 325K), respectively.

Here, we define heat streaks as non-overlapping $s$-day periods where the daily maximum heat index meets or exceeds a specified advisory level $HI_{\text{advisory}}$ on each day.
We calculate the number of $s$-day heat streaks from a time series of daily maximum heat indices $\{HI_{\text{max},1},...,HI_{\text{max},n}\}$ as follows:
\begin{enumerate}
    \item Identify all days where $HI_{\text{max},i} \ge HI_{\text{advisory}}$. Let the indices of these days be $\{i\}_{\text{advisory}}$.
    \item Count the number of non-overlapping sequences of $s$ consecutive indices within $\{i\}_{\text{advisory}}$. This count represents the number of $s$-day heat streaks, denoted as $H^h_{\text{advisory}}$.
\end{enumerate}
For a given period (e.g., 2010-2019), we compute the annual average $s$-day heat streak count for each heat advisory level (i.e. \{caution, extreme caution, danger, extreme danger\}) across all ensemble members.
The error is then the mean absolute difference between the predicted and reference annual average heat streak counts:
\begin{equation}
    \text{heat streak error} = \overline{\left|H^s_\text{advisory} - H^s_\text{advisory, ref}\right|},
\end{equation}
where the mean $\bar{(\cdot)}$ is calculated over the ensemble members.

\subsection{Tropical cyclone detection} \label{si:definition_tc}
\subsubsection{Criteria} \label{si:tropical_cyclones}

Tropical Cyclones (TCs) are detected using the open-source TempestExtremes~\cite{ullrich2021tempestextremes} software package with the following criteria:
\begin{itemize}
    \item Downscaled time slices are analyzed at 6 hour intervals (i.e. a temporal downsampling factor of 3 with respect to the \ourname output time resolution). LENS2 time slices are analyzed at daily intervals, matching the input time resolution.
    \item Local minima in sea level pressure (SLP) are identified, requiring an SLP increase of at least 200 Pa within a 5.0 great circle distance (GCD). Smaller minima within a 6.0 GCD are merged.
    \item Wind speed must exceed 10 m/s for at least 2 days of snapshots (8 for downscaled and 2 for LENS2). The surface elevation of the minima must remain below 100 meters for the same duration.
    \item The minima must persist for at least 54 hours, with a maximum gap of 24 hours in the middle.
    \item The maximum allowable distance between points along the same path is 8.0 GCD.
\end{itemize}
We note that detecting tropical cyclones typically requires further filtering based on upper-level geopotential gap or temperature thresholds to identify the presence of warm-core structures. Such qualifications are excluded from the definition above, as our emphasis is on downscaling near-surface variables. Nonetheless, the criteria remain consistent for both predicted and reference samples, and provide a representative assessment of the associated risks.

Instances of cyclones detected above criteria are outputted as sequences of (longitude, latitude) coordinates representing the locations of the SLP minima, along with the associated SLP values.

\subsubsection{TCG index}
We can estimate how many storms we could expect in the LENS2 ensembles using the Tropical Cyclogenesis (TCG) index~\cite{Tippett2011-gu}. This index predicts the number of storms in a region as a function of the monthly means of several different variables (wind shear, low-level vorticity, relative humidity, and sea-surface temperatures).

\subsubsection{Calibration} \label{si:tc_calibration}

Due to inherent limitations of the LENS2 input, the magnitude of SLP depressions is systematically underestimated in downscaled projections. This results in a reduced frequency of occurrence of tropical storms and hurricanes when applying TC detection algorithms directly on the downscaled data. To address this limitation, we follow the prevalent approach of calibrating the downscaled output to match the observed frequency of TCs over a reference period~\cite{Emanuel_2021, Jing2024}. This is achieved via a conditional affine transformation of the magnitude of SLP depressions:
\begin{equation} \label{eq:tc_slp_calibration}
    P_0^* = 
    \begin{cases}
        K P_0 + (1 - K) P_{0,\text{amb}} & \text{if } P_0 \leq P_{0,\text{amb}} \\
        P_0 & \text{if } P_0 > P_{0,\text{amb}}
    \end{cases}
\end{equation}
where $P_0^*$ denotes the calibrated SLP minimum of the tropical cyclone, $K > 1$ a calibration constant and $P_{0,\text{amb}} = 1010~\text{hPa}$ represents the ambient SLP.

This calibration effectively sharpens local pressure gradients by proportionally decreasing the SLP values below the ambient threshold. It enables the detection algorithm to identify weaker signals that would otherwise be missed. We perform a sensitivity analysis across a range of $K$ values and select the value that results in the best overall match of TC statistics (count, track length and lifetime) during the training period. The same selected scaling constant is then applied for evaluation and future projections. 

\begin{table}[t!]
    \centering
    \caption{Calibration scaling constant ($K$) for Tropical Cyclone (TC) detection. Values were chosen for best fit to TC count, track length, and lifetime in the training period (from $1/K \in \{0.1, 0.2, ..., 0.9 \}$). }
    \label{tab:calib_scaling}
    {
    \renewcommand{\arraystretch}{1.05} %
    \begin{tabular}{l|c}
        \toprule
        Method          & Inverse scaling constant ($1/K$) \\
        \midrule
        GenFocal        & 0.6 \\ 
        BCSD            & 0.2 \\
        Star-ESDM       & 0.2 \\
        QMSR            & 0.2 \\
        SR              & 0.2 \\ \hline
        LENS2           & 0.1 \\ 
        \bottomrule
    \end{tabular}
    }
\end{table}

This calibration procedure is applied to all baselines and ablation models to establish a consistent basis for comparison. The selected $K$ are listed in Table~\ref{tab:calib_scaling}. 
Notably, GenFocal exhibits the smallest required calibration change, as indicated by a $K$ value closest to 1.

\subsubsection{Characteristics}

To ensure consistent interpretation throughout this work, the definitions of the TC characteristics referred to are provided below.

\textbf{Count.} The total number of TCs identified within a specified region and time period.

\textbf{Cyclogenesis density.} A geospatial quantification of the frequency of TC formation in a given region, represented by a histogram of the first point in a TC track binned to a specified spatial resolution over a particular period. To represent the overall density, we average the frequency over the ensemble and present results in a spatial map or zonal/meridional averages.  

\textbf{Length.} The cumulative distance (in km) traversed by a single TC instance from its genesis to dissipation.

\textbf{Lifetime.} The total duration (in hour) for which a TC instance maintains its identity, from its genesis to dissipation.

\textbf{Pressure-derived wind.} Wind speed calculated directly from the detected minimum SLP, following~\cite{atkinson1977tropical}.

\textbf{Saffir-Simpson category.} A classification scale  (tropical depression, tropical storm, and category 1 to 5 hurricanes) for TC intensity based on the pressure derived wind speed~\cite{saffir1973hurricane, simpson1974hurricane}. 

\textbf{Sinuosity index.}  A measure of the curvature of a tropical cyclone track~\cite{Terry2015-sx}.

\textbf{Track density.} 
A geospatial quantification of the frequency of TC passage through a given region, represented by a histogram of detected TC centers binned to a specified spatial resolution over a particular period. To represent the overall density, we average the frequency over the ensemble and present results either in a spatial map of raw average count (e.g. Fig.~\ref{tc-change}a) or a contour plot (e.g. Fig.~\ref{fig:tc}a-c).

\textbf{Landfalls.} Landfall locations are identified by comparing TC tracks with a quarter-degree land–sea grid. A track point is considered over land if the nearest grid cell contains more than 50\% land area. For each track, the landfall location is defined as the first point that meets this criterion.  The 50\% threshold land area threshold filters out the small islands of the West Indies, which is why our landfall plots do not show any landfalls over the Antilles.  

\clearpage

\section{\ourname: methodology and implementation details}
\label{si:method}

\ourname is an end-to-end statistical learning approach for downscaling. In particular, it focuses on downscaling from climate simulations to reanalysis, which is a proxy to the ground-truth weather states in the past. Once learned, the downscaling operation can be applied to future climate projections so that climate impact risks can be assessed at a high-resolution in both spatial and temporal dimensions. \ourname grounds risk assessment of future climate projection on past observations. 

The design behind \ourname addresses three important modeling challenges in downscaling from global climate simulation to observed regional weather states (using reanalysis as a proxy in this work). First, climate simulation is coarse and thus biased with respect to fine-scaled weather. Second, the two sets of data  lack temporal alignment at the granularity needed for risk assessment, such as days or hours; their correspondence is, at best, decadal. Third, downscaled states need to maintain temporal coherence over extended periods (weeks or seasons), which is crucial for robustly estimating compound extreme weather events such as tropical cyclones or heat streaks.

\subsection{Main idea}

The main idea of \ourname is schematically illustrated in Fig.~\ref{fig:framework_diagram}. To overcome the challenges of bias and lack of granular alignment, \ourname introduces an intermediate latent variable $y' \in \mathcal{Y}'$, a sample of the low-resolution but unbiased weather-consistent state:
\begin{equation}\label{eq:genfocal}
    p(x | y) = \int_{\mathcal{Y}'} p(x| y') p(y' | y) \, d y' = p(x | C' x = y')\delta(y' = Ty),
\end{equation}
where $C'$ is a deterministic {\it known} coarse-graining map while $T$ is a deterministic {\it unknown} debiasing map, forming a Dirac distribution at the bias-corrected but low-dimensional $y'$. 

The debiasing operator $T$ is instantiated as a rectified flow~\cite{liu2023flow} to \emph{match the distributions} of the \emph{global} low-resolution climate and weather spaces (see Fig. \ref{fig:framework_diagram}b). The super-resolution step $p(x|y')$ employs a conditional diffusion model~\cite{song2020score} to add fine-grained details in space and increase the temporal resolution from daily means to 2-hourly (see Fig. \ref{fig:framework_diagram}c). To model and enhance temporal coherence, \ourname ``stacks'' multiple snapshots ($y$s) as inputs. The super-resolution step then employs a domain decomposition technique to ensure temporal consistency across long sequences of $x$ (see \ref{si:multi-diffusion} and Fig.~\ref{fig:dfn_sampling_schematic}).

Consequentially, the design philosophy of \ourname establishes a probabilistic description of the problem as a foundational principle. This description is necessary due to the lack of spatiotemporal correspondence between climate simulations and reanalysis data, except on the coarse levels of $\mathcal{O}(100~\text{km})$ and decades. Concretely, any downscaling approach, physical or statistical, needs to address two issues: debiasing the input data and increasing its coarse resolution. The latter is reminiscent of image and video super-resolution, which can be tackled with statistical learning approaches. The former is very challenging as traditional statistical approaches, such as postprocessing, are inadequate due to the lack of direct correspondence required for supervised learning. 

\ourname addresses both of these challenges. From a methodological standpoint, it is noteworthy that while the two statistical learning models employed by \ourname are named differently, they share the unified underlying theme of framing generative AI as probabilistic distribution matching and density estimation for high-dimensional random variables.

\subsection{Setup}

We formulate the statistical downscaling problem by modeling two stochastic processes, $X_t \in \mathcal{X} := \mathbb{R}^d$ and $Y_t \in \mathcal{Y} = \mathbb{R}^{d'}$ with $d > d'$, representing a high-resolution weather process and low-resolution simulated climate process~\cite{majda2012challenges} respectively. These processes are governed by
\begin{align} 
    \label{eq:si_weather_model}
    dX_t & = F(X_t, t) dt, \\
    \label{eq:si_climate_model}
    dY_t & = \text{GCM}(Y_t, t)\, dt  + \sigma(Y_t, t)dW_t,
\end{align}
where $F$ embodies the generally unknown high-fidelity dynamics of $X_t$, and the dynamics of $Y_t$ are often parameterized by a stochastically forced GCM~\cite{ONeill_2016}, in which the form of $\sigma$ is a modelling choice.
Each stochastic process\footnote{For simplicity in exposition, we follow~\cite{ONeill_2016} where the important time-varying effects of the seasonal and diurnal cycles have been ignored, along with jump process contributions.} is associated with a time-dependent measure, $\mu_x(X,t)$ and $\mu_y(Y, t)$, such that $X_t \sim \mu_x(t)$ and $Y_t \sim \mu_y(t)$,
each governed by their corresponding Fokker-Planck equations. 
We assume an \textit{unknown} time-invariant statistical model $C\colon \mathcal{X} \rightarrow \mathcal{Y}$ that relates $X_t$ and $Y_t$ via a possibly nonlinear downsampling map. For brevity, we omit the time-dependency of the random variables $X$ and $Y$ in subsequent discussion. 

In general, \eqref{eq:si_climate_model} is calibrated via measurement functionals to \eqref{eq:si_weather_model} using a single observed trajectory: the historical weather.
The goal of statistical downscaling is to approximate the inverse of $C$ with a downscaling map $D$, trained on data for $t < T$, for a finite horizon $T$, such that $D_{\sharp} \mu_y(t) \approx \mu_x(t)$ for $t > T$. Here, $D_{\sharp} \mu_y(t)$ denotes the push-forward measure of $\mu_y(t)$ through $D$, and $D$ is assumed to be time-independent. 

Note that $D$ is necessarily a stochastic mapping. Thus, we formulate the task of identifying $D$ as sampling from a conditional distribution~\cite{molinaro2024generative}. We define the operator $D \times id$, where $id$ is the identity map, such that $(D \times id)_{\sharp} \mu_y(t) = D_{\sharp} \mu_y(t) \times \mu_y(t) \approx \mu_{x,y}(t)$, where $\mu_{x,y}(t)$ is the underlying \emph{unknown} joint distribution.
Assuming this joint distribution admits a conditional decomposition, we have:
\begin{equation} \label{eq:decomposition}
\mu_{x,y}(X, Y, t) \approx  D_{\sharp} \mu_y (X,t) \times \mu_y (Y,t) = p(X \, | \, Y) \mu_y(Y, t),
\end{equation}
where $p$ is time-independent. 

Thus far, we have cast statistical downscaling as learning to sample from  $p(x\,|\,y)$, which allows us to compute statistics of interest of $D_{\sharp} \mu_y(t) \approx \mu_x(t)$ via Monte-Carlo methods. 
We rewrite $p(x  \,| \, y)$ as the conditional probability distribution $p(x  \,| \, C(x) = y)$. 
Finally, as $p$ is assumed time-independent we model the elements $X \in \mathcal{X}$ and $Y\in \mathcal{Y}$ as random variables with marginal distributions, $\mu_x$ and $\mu_y$ where $\mu_x = \int \mu_x(X, t) dt$ and $\mu_y = \int \mu_y(Y, t) dt$. Thus, our objective is to learn to sample $p(x  \,| \, C(x) = y)$ given only access to samples of the marginals $X$ and $Y$.

There are two issues: we do not know $C$ and even if $C$ is given (approximately), it is not obvious how we can sample efficiently from $p(x  \,| \, C(x) = y)$.

\subsection{Overview}

Without any additional assumption, it is difficult to learn $C$ from training data. \ourname stipulates a \emph{structural decomposition inductive prior}:
\begin{equation}
\label{eq:factorization}
    C = T^{-1} \hspace{-2pt}\circ C', \quad \text{such that} \quad (T^{-1} \hspace{-2pt}\circ C')_{\sharp} \mu_x = \mu_y, 
\end{equation}
where $C$ consists of two components:
\begin{itemize}
\item \emph{Downsampling}\footnote{Here we suppose that the downsampling map acts both in space and in time, by using interpolation in space, and by averaging in time using a window of one day.} The range of $C' \colon \mathcal{X} \rightarrow \mathcal{Y}'$ defines an \emph{intermediate} space $\mathcal{Y}' = \mathbb{R}^{d'}$ of low-resolution samples with measure $\mu_{Y'}:=C'_{\sharp} \mu_x $ (see Fig. \ref{fig:framework_diagram}c). The key assumption is that this step only reduces resolution but does not introduce bias.
\item \emph{Biasing} The invertible biasing map $T^{-1}: \mathcal{Y}' \rightarrow \mathcal{Y}$ defines a correspondence between the two low-dimensional spaces. Conversely, $T$, the inverse of this biasing map, defines the map to debias: $ T_{\sharp} \mu_y =\mu_{y'} =  C'_{\sharp} \mu_x  $ (see Fig. \ref{fig:framework_diagram}b).
\end{itemize}

Thus, downscaling, the inverse of $C$,  becomes a sequential two-step process:
\begin{itemize}
    \item \textit{Bias correction}: Apply a transport map to match the probabilistic distributions such that 
    \begin{equation} \label{eq:constraint} T_{\sharp} \mu_y = C'_{\sharp} \mu_x. \end{equation}
    \item \textit{Statistical Super-resolution}: For the joint variables ${X} \times {Y}'$, approximate $p(x \, | \, C'x = y')$.
\end{itemize}

Introducing the intermediate space $\mathcal{Y}'$ is, in equivalence, to define the conditional distribution $p(x |y)$ via a latent variable, which in return leads to \eqref{eq:genfocal}. The Dirac distribution is chosen to reflect the deterministic and invertible mapping. An extension to probabilistic mapping is possible and left for future work.

\ourname employs two state-of-the-art generative AI techniques to build the bias correction and super-resolution maps: the bias correction step is instantiated by a conditional flow matching method~\cite{liu2023flow}, whereas the super-resolution step is instantiated by a conditional denoising diffusion model~\cite{song2020improved} coupled with a domain decomposition strategy~\cite{bar2024lumiere} to upsample both in space and time and create time-coherent sequences.

\subsection{Bias correction}
\label{si:debias}

For debiasing, since the samples from $\mathcal{Y}$ and $\mathcal{Y'}$ are not aligned, we seek a map between the distributions. This is a weaker notion than sample-to-sample correspondence, which physics-based downscaling methods might be able to offer. In exchange, statistical distribution match, as shown in this work, can also be effective in debiasing yet remaining computationally advantageous. 

The notion of distribution matching has a long history in applied mathematics going back to Gaspar Monge in the late 1700s and Leonid Kantorovich in the 50s, who formalized this idea, and kicked off the field of optimal transport~\cite{villani2009optimal}. In our context, the optimal transport framework would seek to solve the problem
\begin{equation} \label{eq:ot_formulation}
    \min_{T} \int c(Ty, y) d\mu_y (y) \ \text{with}\ T_{\sharp} \mu_y = \mu_{y'}:= C'_{\sharp} \mu_x,
\end{equation}
for a cost function $c$ measuring the cost moving ``probabilistic mass'.
Note that following this approach, the debiasing map $T$ satisfies the constraint in \eqref{eq:constraint} by construction.

Due to limitations of existing methods for solving \eqref{eq:ot_formulation} (which are briefly summarized in \ref{si:related_ot}), we adopt a rectified flow approach~\cite{liu2023flow}, a methodology under the umbrella of generative models. Rectified flow results in a {\it invertible} map instantiated by the solution map of an ODE, which solves an entropy-regularized optimal transport problem~\cite{liu2022rectified}, and it has empirically shown to be well suited for relatively large dimension (as compared to control based approaches such as neural ODE \cite{neuralODE}), and it has a relatively low sample complexity.

\subsubsection{Rectified flow}
Rectified flow constructs the debiasing map $T$ as the solution map of an ODE given by
\begin{equation} \label{si:eq:reflow_ode}
    \frac{d y}{d \tau} = v_{\phi}(y, \tau) \qquad \text{for } \tau \in [0, 1],
\end{equation}
whose vector field $v_{\phi}(x, \tau)$ is parameterized by a neural network (see \ref{si:debiased_architecture} for further details). By identifying the input of the map as the initial condition, we have that $T(y) := y(\tau = 1)$.  We train $v_{\phi}$ by solving 
\begin{equation} \label{si:eq:reflow_ode_estimation}
    \ell(\phi) = \min_{\phi} \mathbb{E}_{\tau \sim \mathcal{U}[0,1]} \mathbb{E}_{(y_0, y_1) \sim \pi \in \Pi(\mu_y, \mu_{y'})} \| (y_1 - y_0) - v_{\phi}(y_{\tau}, \tau) \|^2, 
\end{equation}
where $y_{\tau} = \tau y_1 + (1-\tau) y_0$.  $\Pi(\mu_y, \mu_{y'})$ is the set of couplings with marginals given by the distributions from $\mathcal{Y}$ and $\mathcal{Y'}$ respectively.  Once $v_(\phi)$ is learned, we debias any given $y$ by solving \eqref{si:eq:reflow_ode} using the $4^{\text{th}}$-order Runge-Kutta solver.

\subsubsection{Modeling details}

Our implementation of rectified flow introduces several custom modeling choices that are highly effective when dealing with climate data: (a) modeling in the anomaly space; (b) modeling the seasonality of climate by climatological distribution coupling; (c) modeling temporal coherence.

Let $y \in \mathcal{Y}$ denote a biased low-resolution sequence of consecutive snapshots (namely, the climate state at times $t, t+\Delta t, ..., t+n_s \Delta t$), and  $y' \in \mathcal{Y}'$ denote an unbiased low-resolution sequence, where $ \mathcal{Y}'$ is the image of $\mathcal{X}$ through the linear downsampling map $C'$ (see Fig.~\ref{fig:framework_diagram}a).
In our setup, the space of biased low-resolution dataset $\mathcal{Y}$ is given by a collection of $100$ trajectories from the LENS2 ensemble dataset. Each trajectory, which we denote by $\mathcal{Y}^i$ (such that $\mathcal{Y} = \bigcup_{i} \, \mathcal{Y}^i$), has slightly different spatiotemporal statistics that we leverage to further extract statistical performance from our debiasing step. We characterize the statistics of each trajectory using their climatological mean and standard deviation in the training set, namely $\bar{y}^i$ and $\sigma^i_y$, which are estimated using the samples within the training range. The space of the unbiased low-resolution sequences $\mathcal{Y}'$, is given by the daily means of the ERA5 historical data regridded to $1.5$° resolution. We denote the climatological mean and standard deviation of the set as $\bar{y}'$ and $\sigma_{y'}$ respectively.

To render the training more efficient, we normalize the input and output data using their {\it climatology} following: 
$\hat{y} = (y - \bar{y}^i) / \sigma_y^i$ for $y \in \mathcal{Y}^i$, and $\hat{y}' = (y' - \bar{y}') / \sigma_{y'}$. Then we seek to find the smallest deviation between the two {\it anomalies}.

We specialize the map $T$ as follows.  We incorporate the climatological mean and standard deviation into the vector field $v_{\theta}(y, \tau; \bar{y}^i, \sigma_y^i)$ and identify the solution of the revised ODE
\begin{equation} \label{si:eq:app_reflow_ode}
    \frac{d \hat{y}}{d \tau} = v_{\theta}(\hat{y}, \tau; \bar{y}^i, \sigma_y^i), \qquad \text{for } \tau \in (0, 1),
\end{equation}
at the terminal time as the unbiased anomaly, i.e., $\hat{y}' = \hat{y}(1)$. This is then de-normalized,  resulting in $Ty = y' = \hat{y}' \odot \sigma_{y'} + \bar{y}'$, where $\odot$ is the Hadamard product.

The training loss is revised accordingly
\begin{equation}
    \label{eq:app_reflow_loss}
    \min_{\theta} \, \mathbb{E}_{i\in \mathbb{I}} \mathbb{E}_{\tau \sim \mathcal{U}[0,1]} \mathbb{E}_{(y_0, y_1) \sim \pi \in \Pi(\mu_{y}^i, \mu_{y'})} \| \hat{y}_0 - \hat{y}_1 - v(\hat{y}_{\tau}, \tau; \hat{y}^i, \sigma_y^i) \|^2, 
\end{equation}
where $\hat{y}_{\tau} = \tau \hat{y}_1 + (1-\tau) \hat{y}_0$, $\Pi(\mu_{y^i}, \mu_{y'})$ is the set of couplings with climatologically aligned marginals, and $\mathbb{I}$ are the indexes of the training trajectories instantiated by the different LENS2 ensemble members. 

The choice of the coupling \rev{implicitly defines the spatiotemporal structure of the bias to be rectified. We assume a seasonally-varying bias by} coupling data pairs that correspond to similar time stamps (possibly up to a couple of years) for both LENS2 and ERA5 samples. \rev{Although, for simplicity in this case we use a coupling that uses the same time-stamps for both LENS2 and ERA5 samples to ensure the same climatology and to capture the correct slowly-varying drift in the distributions induced by the climate change signal.} \rev{For an ablation study see section \ref{si:ablations_coupling}}

Time-coherence is implicitly included in this step. At each iteration, data is extracted from a long contiguous sequence of snapshots. For example, with a batch size of 16 and a debiasing sequence length of 8, we extract $128 = 16 \times 8$ consecutive snapshots from a single LENS2 member and ERA5. That long sequence is then divided into 16 short sequences and fed to each core. We observed that choosing short sequences from the training dataset in a fully independent manner was prone to overfitting; this effect was attenuated by feeding a batch of contiguous sequences as described above. This approach also helps optimize training by reducing data loading latency, as it minimizes the number of reads from disk. 

For the length of each debiasing sequence, empirically we found that 2–8 contiguous days provides good performance on the validation set. For an ablation study of the chunk size, please see Section~\ref{si:ablations}.
Once the model is trained, we solve \eqref{si:eq:app_reflow_ode} using an adaptive Runge-Kutta solver, which allows us to align the simulated climate manifold to the weather manifold.

\subsubsection{Neural architecture}
\label{si:debiased_architecture}
For the architecture we use a 3D U-ViT~\cite{bao2022all}, with 6 levels.
The input to the network are three 4-tensors, $\hat{y}$, $\bar{y}^i$, and $\sigma_y^i$; each of dimensions $8 \times 240 \times 121 \times 10$ plus a scalar corresponding to the evolution time $\tau$. Here 
the 8 corresponds to the 8 contiguous days, and the 10 channels correspond to the surface and level fields being modeled as shown in Table \ref{table:modeled_vars_debiasing}. The output is one 4-tensor corresponding to the instant velocity of $\hat{y}_{\tau}$. In this case, $\bar{y}^i$, and $\sigma_y^i$ are used as conditioning vectors. These variables are interpolated to the new grid, and pre-processed using a convolutional neural network, then they are concatenated to $\hat{y}$ along the channel dimension.

\paragraph{Resize and aggregation layers for encoding}
As the spatial dimensions of input tensors, $240\times 121$, are not easily amenable to downsampling, i.e, they are not multiples of small prime numbers, we use a resize layer at the beginning. The resize layers performs a cubic interpolation to obtain a 3-tensor of dimensions $8 \time 256 \times 128 \times 10$, followed by a two-dimensional convolutional network with lat-lon boundary conditions: periodic in the longitudinal dimension (using the \texttt{jax.numpy.pad} function with \texttt{wrap} mode) and constant padding in the latitudinal dimension, which repeats the value at the end of the array (using the \texttt{jax.numpy.pad} function with \texttt{edge} mode).

For the $\hat{y}$ inputs, the convolutional network works as a dealiasing step. It has a kernel size of $(7, 7)$, which we write as: 
\begin{equation} \label{eq:resizing}
    h_{\hat{y}} = \texttt{Conv2D}(8, 7, 1)\circ \mathcal{I}(\hat{y}),
\end{equation}
where $\texttt{Conv2D}(N, k, s)$ denotes a convolutional layer with $N$ filters, kernel size $(k, k)$ and stride $(s, s)$.

The conditioning inputs, i.e., the statistics  $\bar{y}^i$, and $\sigma^i$, go through a slightly different process: they are are also interpolated, but they go trough a shallow convolutional network composed of one two-dimensional convolutional layers followed by a normalization layer with a swish activation function, and another two-dimensional convolutional layer. Here, both convolutional layers have a kernel size $(3,3)$. The first has an embedding dimension of $10$ as it acts  as an anti-aliasing layer while the second has an embedding dimension of $128$ as it seeks to project the information into the embedding space. In summary, we have
\begin{align}
    h_{\bar{y}^i} = \texttt{Conv2D}(128, 3, 1) \circ\texttt{Swish}\circ\texttt{LN}\circ\texttt{Conv2D}(4, 3, 1)\circ \mathcal{I}(\bar{y}^i),\\
    h_{\sigma^i} = \texttt{Conv2D}(128, 3, 1)\circ\texttt{Swish}\circ\texttt{LN}\circ\texttt{Conv2D}(4, 3, 1)\circ \mathcal{I}(\sigma^i).
\end{align}
Then all the fields are concatenated along the channel dimension, i.e., 
\begin{equation}
    h = \texttt{Concat}[h_{\hat{y}};\, h_{\bar{y}^i};\,  h_{\sigma^i}],
\end{equation}
of dimensions $8 \times 256 \times 128 \times 266$. The last dimension comes from the concatenation of $h_{\hat{y}}$ which has channel dimension $10$, together with $h_{\bar{y}^i}$ and $h_{\sigma^i}$, which have a channel dimension of 128 each.

\paragraph{Spatial downsampling stack}
After the inputs are rescaled, projected and concatenated, we feed the composite fields to an U-ViT. For the downsampling stack we use $4$ levels, at each level we downsample by a factor two in each dimension, while increasing the number of channels by a factor of two, so we only have a mild compression as we descend through the stack.

The first layer takes the output of the merge and resizing, and we perform a projection
\begin{equation}\label{eq:dstack_proj}
    h_0 = \texttt{Conv2D}(128, 3, 1)(h),
\end{equation}
where $h$ is the latent input from the encoding step. Then $h_0$ is successively downsampled using a convolution with stride $(2,2)$, and an embedding dimension of $\text{hidden}_i$, where $i$ is the level of the U-Net.
\begin{equation}\label{eq:dstack_lvl}
    h^{\text{down}}_{i,0} = \texttt{Conv2D}(\text{hidden}_i, 1, 2)(h_{i-1,n_{res}-1}),
\end{equation}
where $n_{res}$ is the number of resnet at each level, and $\text{hidden}_i$ is the dimension of the hidden states for each level as given in Table \ref{table:flow_hyp}. The output of the downsampled embedding is then then processed by a sequence of $n_{res}=6$ resnet blocks following:
\begin{equation}\label{eq:resblock_downsample}
    h^{\text{down}}_{i,j+1} = h^{\text{down}}_{i,j} + \texttt{Conv2D}(c^i, 3, 1) \circ \texttt{Do}(0.5)\circ\texttt{Swish}\circ\texttt{FiLM}(e)\circ\texttt{GN}\circ \texttt{Conv2D}(c^i, 3, 1)\circ\texttt{Swish}\circ\texttt{GN}(h^{\text{down}}_{i,j}),
\end{equation}
where $c^i = \text{hidden}_i$, the number of channels at each level, \texttt{Do}$_{p}$ is dropout layer with probability $p$, here $j$ runs from $0$ to $n_{res}-1$. In addition, time embedding $e$, is processes with a Fourier embedding layer with a dimension of $256$, which is then used in conjunction with a \texttt{FiLM} layer following
\begin{equation}
    \begin{gathered}
        \texttt{FiLM}(x; \sigma_\tau) = (1.0 + \texttt{Dense}\circ\texttt{FourierEmbed}(\sigma_\tau)) \cdot x + \texttt{Dense}\circ\texttt{FourierEmbed}(\sigma_\tau), \\
        \texttt{FourierEmbed}(\sigma_\tau) = \texttt{Dense} \circ \texttt{SiLU} \circ \texttt{Dense} \circ \texttt{Concat} ([\cos(\alpha_k\sigma_\tau), \sin(\alpha_k\sigma_\tau)]_{k=1}^K)
    \end{gathered}
\end{equation}
where $\alpha_k$ are non-trainable frequencies evenly spaced on a logarithmic scale between $0$ and $10000$, and $K= 128$. Finally, \texttt{GN} stands for a group normalization layer with $4$ groups. 

\paragraph{Attention Processing}
For the attention layers we use a ViViT-like model with 2D position encoding, axial transformer in each direction, $128$ heads, the token sizes depends at which level the attention processing is performed. Also, the temporal and spatial attentions are decoupled so they can be used (or not) independently. 

\paragraph{Spatial upsampling stack}
The upsampling stack takes the downsampled latent variables and sequentially increases their resolution while merging them with skip connections until the original resolution is reached. This process, within the U-ViT model, is completely different from the super-resolution stage of the framework as shown in Fig.~\ref{fig:framework_diagram}, which is treated in detail in~\ref{si:upsampling} 
The upsampling stack contains the same number of levels and residual blocks as the downsampling one. At each level, it adds the corresponding skip connection in the 
upsampling stack:
\begin{equation}%
    h_{i, 0}^{\text{up}} = h_{i,0}^{\text{up}} + h_{i, 0}^{\text{down}},
\end{equation}
followed by the same blocks defined in~\eqref{eq:resblock_downsample}, followed by an upsampling block
\begin{equation}
    h^{\text{up}}_{i-1, n_{res}-1} = \texttt{Conv2D}(\text{hidden}_{i-1}, 3, 1)
    \circ \texttt{channel2space} 
    \circ \texttt{Conv2D}(\text{hidden}_i \cdot 2^2, 3, 1) 
    \circ h_{i, n_{res}-1}^{\text{up}},
\end{equation}
where the \texttt{channel2space} operator expands the $\text{hidden}_i \cdot 2^2$ channels into $2\times 2\times \text{hidden}_i$  blocks locally, effectively increasing the spatial resolution by $2$ in each direction.

\paragraph{Decoding and resizing}
We apply a final block to the output of the upsampling stack. 
\begin{equation}
    x_\text{out} = \texttt{Conv2D}(10, 3, 1) 
    \circ \texttt{SiLU} 
    \circ \texttt{LayerNorm} 
    \circ h^{\text{up}}_0.
\end{equation}
followed by a resizing layer as the one defined in~\eqref{eq:resizing}, with number of channels equal to the number of input fields. This operation 
brings back the output to the size of the input.

\subsubsection{Hyperparameters}
\label{si:flow_hyp}

Table~\ref{table:flow_hyp} shows the set of hyperparameters used for the flow architecture, as well as those applied during the training and sampling phases of the rectified flow model. We also include the optimization algorithm used for minimizing \eqref{eq:app_reflow_loss}, along with the learning rate scheduler and weighting.

\begin{table}[t]
\centering
\footnotesize
\caption{Hyperparameters for the debiasing model.}
\label{table:flow_hyp}
{
\renewcommand{\arraystretch}{1.0} %
\begin{tabular}{p{13em}p{20em}}

\hline
\multicolumn{2}{l}{Debias architecture} \\ \hline
Output shape & $8 \times 240\times121\times 10$ \\
Spatial downsampling ratios   & [2, 2, 2, 2, 2, 2] \\
Residual blocks & [6, 6, 6, 6, 6, 6]    \\
Hidden channels                 & [768, 768, 768, 1024, 1280, 1536]  \\ 
Axial attention layers in space  & [
\texttt{False}, \texttt{False}, \texttt{False}, \texttt{False}, \texttt{True}, \texttt{True}] \\ 
Axial attention layers in time  & [
\texttt{False}, \texttt{False}, \texttt{False}, \texttt{True}, \texttt{True}, \texttt{True}]  
\\ 
Trainable parameters & 2,656,553,626 \\ \hline
\multicolumn{2}{l}{Training} \\ \hline
Device                           & TPU v5p, $4 \times 4$   \\
Duration                         & 500,000 steps \\
Batch size                       & $16$ (with data parallelism) \\
Learning rate                    & cosine annealed (peak=$1\times10^{-4}$, end=$1\times10^{-7}$), 1,000 linear warm-up steps                  \\
Gradient clipping                & max norm = 0.6                                                                                           \\
Time sampling                    & $\mathcal{U}(10^{-3}, 1 - 10^{-3})$                                      \\
Condition dropout ($p_u$)  & 0.5  \\ \hline
\multicolumn{2}{l}{Inference }                                                                            \\ \hline
Device                           & $8\times$Nvidia H100s  \\
Integrator                       & Runge-Kutta 4th order \\
Solver number of steps                & $100$ \\ \hline
\end{tabular}
}
\end{table}

\subsubsection{Training, evaluation and test data}
\label{si:debiasing_data_set}

We trained the debiasing stage of \ourname using 4 LENS2 members \texttt{cmip6\_1001\_001}, \texttt{cmip6\_1251\_001}, \texttt{cmip6\_1301\_010}, and \texttt{smbb\_1301\_020}, using data from $1980$ to $1999$. We point out that the first three members share the same forcing using the original CMIP6 BMB protocol~\cite{LENS2}, but different initializations to sample internal variability, whereas the last one uses a smoothened version of the same forcing (see details in ~\cite{LENS2}). Debiasing is performed with respect to the coarse-grained ERA5 data for the same period.

For model selection we used the following 14 LENS2 members:
 \texttt{cmip6\_1001\_001}, \texttt{cmip6\_1041\_003},
 \texttt{cmip6\_1081\_005}, \texttt{cmip6\_1121\_007},
 \texttt{cmip6\_1231\_001}, \texttt{cmip6\_1231\_003},
 \texttt{cmip6\_1231\_005}, \texttt{cmip6\_1231\_007},
 \texttt{smb\_1011\_001}, \texttt{smbb\_1301\_011},
 \texttt{cmip6\_1281\_001}, \texttt{cmip6\_1301\_003},
 \texttt{smbb\_1251\_013}, and \texttt{smbb\_1301\_020}, using data from $2000$ to $2009$.

For testing we use the full 100-member LENS2 ensemble from $2010$ to $2019$. The full ensemble used for testing contains members with different forcings and perturbations.

\subsubsection{Computational cost}
\label{si:reflow_cost}
Training the rectified flow model took approximately three days using four TPU v5p nodes (16 cores total), with one sample per core. Each host loaded a sequence of 32 contiguous daily snapshots per iteration (four sequences of 8 consecutive snapshots), which were then distributed among the cores.
For inference, each sample of 8 snapshots takes around 45 seconds to be debiased in an H100. The full debiasing step took around 9 hours to process each 140-year ensemble member on a host with 8 H100 GPUs. As the process is embarrassingly parallel, debiasing the full 100-member LENS2 ensemble for 140 years took about 9 hours using 100 nodes, each equipped with 8 H100 GPUs. Estimating each H100 costs about $\$5$ USD per hour at the current market rate, this debiasing step costs about $\$36,000$ USD and can be further reduced through engineering optimization.

\subsection{Super-resolution}
\label{si:upsampling}

In contrast to  bias correction, super-resolution is a probabilistic supervised learning problem. The coarse-graining map $C'$ is an operation by downsampling the ERA5 data from 2-hourly and $0.25^{^{\circ}}$  to daily $1.5^{^{\circ}}$, thus forming a pair of aligned data sample $(y_i'= C'x_i, x_i)$. To learn the super-resolution operation, i.e., the inverse of the downsampling, we use a conditional diffusion model~\cite{song2020improved,Song_2020}, popularized by latest advances in image and video generation.  

\subsubsection{Conditional diffusion model}
\label{si:diffusion}

In this section, we provide a brief high-level description of the generic diffusion-based generative modeling framework. While different variants exist, we mostly follow that of~\cite{karras2022elucidating} and refer interested readers to its Appendix for a detailed explanation of the methodology.

Diffusion models are a type of generative model that work by gradually adding Gaussian noise to real data until they become indistinguishable from pure noise (forward process). The unique power of these models is their ability to reverse this process, starting from noise and progressively refining it to create new samples that resemble the original data (backward process, or sampling).

Mathematically, we describe the forward diffusion process as a stochastic differential equation (SDE)
\begin{equation}\label{eq:forward_sde}
    dz_\tau = \sqrt{\dot{\sigma}_\tau\sigma_\tau}\; d\omega_\tau, \quad z_0 \sim p_{\text{data}},\; \tau\sim[0, 1]
\end{equation}
where $\sigma_\tau$ is a prescribed noise schedule and a strictly increasing function of the diffusion time $\tau$ (note: to be distinguished from real physical time $t$), $\dot{\sigma}_\tau$ denotes its derivative with respect to $\tau$, and $\omega_\tau$ is the standard Wiener process. The linearity of the forward SDE implies that the distribution of $z_\tau$ is Gaussian given $z_0$:
\begin{equation}\label{eq:perturb_kernel}
    q(z_\tau|z_0) = \mathcal{N}(z_\tau; z_0, \sigma^2_\tau I),
\end{equation}
with mean $z_0$ and diagonal covariance $\sigma^2_\tau I$. For $\tau = 1$, i.e. the maximum diffusion time, we impose $\sigma_{\tau=1} \gg \sigma_{\text{data}}$ such that $q(z_1|z_0)$ can be faithfully approximated by the isotropic Gaussian $\mathcal{N}(z_1; 0, \sigma^2_1I) := q_1$. 

The main underpinning of diffusion models is that there exists a \emph{backward SDE}, which, when integrated from $\tau=1$ to $0$, induces the same marginal distributions $p(z_\tau)$ as those from the forward SDE~\eqref{eq:forward_sde} \cite{anderson1982reverse, Song_2020}:
\begin{equation}\label{eq:backward_sde}
    dz_\tau = - 2\dot{\sigma}_\tau\sigma_\tau\nabla_{z_\tau} \log{p\left(z_\tau, \sigma_\tau\right)}\;d\tau + \sqrt{2\dot{\sigma}_\tau\sigma_\tau}\;d\omega_\tau.
\end{equation}
In other words, with full knowledge of~\eqref{eq:backward_sde} one can easily draw samples $z_1\sim q_1$ to use as the initial condition and run a SDE solver to obtain the corresponding $x_0$, which resembles a real sample from $p_\text{data}$. However, in~\eqref{eq:backward_sde}, the term $\nabla_{z_\tau} \log{p\left(z_\tau, \sigma_\tau\right)}$, also known as the \emph{score function}, is not directly known. Thus, the primary machine learning task associated with diffusion models is centered around expressing and approximating the score function with a neural network, whose parameters are learned from data. Specifically, the form of parameterization is inspired by Tweedie's formula~\cite{efron2011tweedie}:
\begin{equation}\label{eq:tweedies}
    \nabla_{z_\tau} \log{p\left(z_\tau, \sigma_\tau\right)} = \frac{\mathbb{E}[z_0|z_\tau]-z_\tau}{\sigma_\tau^2} \approx \frac{D_{\theta}(z_\tau, \sigma_\tau) - z_\tau}{\sigma_\tau^2},
\end{equation}
where $D_{\theta}$ is a \emph{denoising} neural network that predicts the clean data sample $z_0$ given a noisy sample $z_\tau = z_0 + \varepsilon\sigma_\tau$ ($\varepsilon$ is drawn from a standard Gaussian $\mathcal{N}(0, I)$). Training $D_{\theta}$ involves sampling both data samples $z_0$ and diffusion times $\tau$, and optimizing parameters $\theta$ with respect to the mean denoising loss defined by
\begin{equation}\label{eq:dfn_train_loss}
    \mathcal{L}(\theta) = \mathbb{E}_{z_0\sim p_{\text{data}}} \mathbb{E}_{\tau\sim[0, T]} \;\big[\lambda_\tau\|D_{\theta}(z_0 + \epsilon\sigma_\tau, \sigma_\tau) - z_0||^2\big],
\end{equation}
where $\lambda_\tau$ denotes the loss weight for noise level $\tau$. We use the weighting scheme proposed in~\cite{karras2022elucidating} as well as the pre-conditioning strategies therein to improve training stability.

At sampling time, the score function in SDE~\eqref{eq:backward_sde} is substituted with the learned denoising network $D_\theta$ using expression \eqref{eq:tweedies}. 

In the case that an input is required, i.e. sampling from conditional distribution $p(z_\tau|y)$, the input $y$ is incorporated by the denoiser $D_\theta$ as an additional input. Classifier-free guidance (CFG)~\cite{ho2022} may be employed to trade off between maintaining coherence with the conditional input and increasing coverage of the target distribution. Concretely, CFG is implemented through modifying the denoising function $\tilde{D}_\theta$ at sampling time:
\begin{equation}\label{eq:guidance}
    \tilde{D}_\theta = (1+g){D}_\theta(z_\tau, \sigma_\tau, y) - g{D}_\theta(z_\tau, \sigma_\tau, \varnothing),
\end{equation}
where $g$ is a scalar that controls the guidance strength (increasing $g$ means paying more attention to $y$) and $\varnothing$ denotes the null conditioning input (i.e., a zero-filled tensor with the same shape as $y$), such that ${D}_\theta(x_\tau, \sigma_\tau, \varnothing)$ represents unconditional denoising. The unconditional and conditional denoisers are trained jointly using the same neural network model, by randomly dropping the conditioning input from training samples with probability $p_u$.

\subsubsection{Modeling details}
\label{si:clim_norm}

We specialize the general framework of conditional distribution models to modeling weather and climate data. \ourname has several specific components that take into consideration the unique properties of the data to facilitate learning.

We take advantage the prior knowledge that a spatially-interpolated linear mapping $\mathcal{I}(y')$ is a strong approximation to $x$ by modeling the residual $r :=  x - \mathcal{I}(y')$ by using the conditional diffusion model to sample from $p(r | y')$ and add the residual back to $\mathcal{I}(y')$ as the final output of the super-resolution. Furthermore,  a substantial portion of the variability in $r_h$ is due to its strong seasonal and diurnal periodicity. To avoid learning these predictable patterns and direct the model's focus toward capturing non-trivial interactions, we learn to sample $\tilde{r}$, the residual normalized by its climatological mean and standard deviation computed over the training dataset:
\begin{equation} \label{eq:sr_output_normalize}
    \tilde{r} = \frac{r - \texttt{clim\_mean}[r]}{\texttt{clim\_std}[r]}.
\end{equation}

The input $y'$ is also strongly seasonal. However, we do not remove its seasonal components and instead normalize with respect to its date-agnostic mean and standard deviation:
\begin{equation} \label{eq:sr_input_normalize}
    \tilde{y}' = \frac{y' - \texttt{mean}[y]}{\texttt{std}[y]},
\end{equation}
which ensures that the model is still able to to leverage the seasonality in the input signals when deriving its output.

In summary, samples are obtained as
\begin{equation} \label{eq:sr_ansatz}
    x(y'; \omega) = \mathcal{I}(y') + \texttt{clim\_mean}[r] + \texttt{clim\_std}[r] \cdot S(\tilde{y}'; \omega)
\end{equation}
where $S(\tilde{y}'; \omega)$ denotes the sampling function (i.e. solving the reverse time SDE end-to-end) for $\tilde{r}$ given the normalized coarse-resolution input $\tilde{y}'$, and a noise realization $\omega$.

\subsubsection{Sampling long temporal sequence}
\label{si:multi-diffusion}

After the denoiser is trained, we may initiate a backward diffusion process by solving~\eqref{eq:backward_sde} from $\tau=1$ to $\tau=0$, using initial condition $z_1\sim q_1$. We employ a first-order exponential solver, whose step formula (going from noise level $\sigma_i$ to $\sigma_{i-1}$) reads
\begin{equation}
    z_{i-1} = \frac{\sigma^2_{i-1}}{\sigma^2_i}z_i + \left(1-\frac{\sigma^2_{i-1}}{\sigma^2_i}\right)D_\theta\left(z_\tau, \sigma_\tau, \tilde{y}'\right) + \frac{\sigma_{i-1}}{\sigma_{i}}\sqrt{\sigma^2_{i} - \sigma^2_{i-1}}\varepsilon,
\end{equation}
where $\varepsilon\sim\mathcal{N}(0, I)$. The generated sample would be the residual for a 7-day window (i.e. model duration) corresponding to the daily mean in $\tilde{y}'$.

To generate an arbitrarily long sample trajectory with temporal coherence, we stagger multiple 7-day windows, denoted by $\{S_0, \dots, S_{M-1}\}$, such that there is a one-day overlap between neighboring windows $S_j$ and $S_{j\pm1}$. A separate backward diffusion process is initiated on each period $S_j$, leading to denoise outputs $\{d_j\}$. As such, each overlapped window has two distinct denoise outputs at every step, denoted $d_j^{\text{right}}$ and $d_{j+1}^{\text{left}}$, which in general do \emph{not} take on the same values despite the corresponding inputs $z_j^{\text{right}}$ and $z_{j+1}^{\text{left}}$ being the same. 

\begin{figure}[t]
    \centering
    \includegraphics[width=\textwidth]{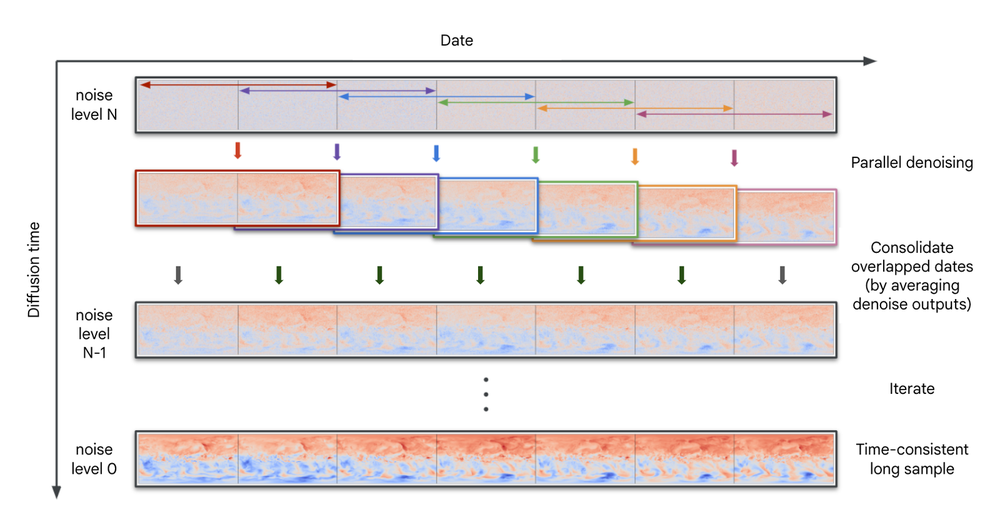}
    \caption{\textbf{Schematic of long trajectory sampling using parallel section denoisers.}}
    \label{fig:dfn_sampling_schematic}
\end{figure}

To consolidate, we take the average between them, and replace the corresponding outputs on both sides to ensure that $d_j$ is consistent between the left and right denoising windows. This in turn creates a ``shock'' that renders the overlapped region \emph{less coherent} with respect to the other parts in their respective native denoising windows. However, the incoherence are expected to be insignificant under the presence of noise and more importantly, should decrease in magnitude as the backward process proceeds and the noise level reduces. At the end of denoising, one would expect a fully coherent sample across all denoising windows. A schematic for this technique is shown in Fig.~\ref{fig:dfn_sampling_schematic}.

Mathematically, the step formula in the overlapped region can be described as
\begin{equation}
    \begin{aligned}
        z_{i-1,j}^{\text{right}} = \frac{\sigma^2_{i-1}}{\sigma^2_i}z_{i,j}^{\text{right}} &+ \frac{1}{2}\left(1-\frac{\sigma^2_{i-1}}{\sigma^2_i}\right)
        \left(d_{i,j}^{\text{right}} + d_{i,j+1}^{\text{left}}\right)  \\
        &+ \frac{\sigma_{i-1}}{\sigma_{i}}\sqrt{\sigma^2_{i} - \sigma^2_{i-1}}\varepsilon_{j}^{\text{right}}.
    \end{aligned}
\end{equation}
It is important to note that the random vector in the same overlapped region should be identical, i.e. ($\varepsilon_{j}^{\text{right}} = \varepsilon_{j+1}^{\text{left}}$).

The complete sampling procedure is described in Algorithm~\ref{alg:sampling}. In practice, we place 
each denoising window on a different TPU core so that all windows can be denoised in parallel. Consolidation of overlapping windows then takes place through collective permutation operations (\texttt{lax.ppermute} functionality in JAX), which efficiently exchanges information among cores.

\begin{algorithm}[t]
\footnotesize
\caption{Sampling long trajectories using overlapped denoisers. Each denoiser takes $84\times 4D_\text{lon}\times 4D_\text{lat}\times4$ input noise shape and generates outputs of the same shape. With 1-day overlap windows and $M=16$ denoisers, the total trajectory shape amounts to $1164\times 4D_\text{lon}\times 4D_\text{lat}\times4$ (97 days).}
\label{alg:sampling}
\begin{algorithmic}[1]
\Procedure{LongTrajectorySampler}{$D_\theta(z, \sigma, y)$, $\sigma_{i \in \{N, \dots, 0\}}, S_{j \in \{0, \dots, M-1\}}$} 
    \State sample $z_N \sim \mathcal{N}(0, \sigma_N^2 I)$ \Comment{Sample shape is the that of the overall trajectory.}
    \State $\{z_{N,0}, \dots, z_{N,M-1}\} \gets \text{extract}(z_N, \{S_0,\dots,S_{M-1}\})$ \Comment{Each $z_{N,j}$ is in denoiser shape.}
    \For{$i \in \{N, \dots, 1\}$} \Comment{Iterate over diffusion steps.}
        \For{$j \in \{0, \dots, M-1\}$}
            \State $d_{i,j} \gets D_\theta(z_{i,j}, \sigma_i, y_j)$ 
            \Comment{Denoise each section independently.}
        \EndFor
        \For{$j \in \{0, \dots, M-1\}$}
            \State $d_{i,j}^{\text{left}} \gets (d_{i,j}^{\text{left}} + d_{i,j-1}^{\text{right}}) / 2$ \Comment{Consolidate with left neighbor (for $j \neq 0$).}
            \State $d_{i,j}^{\text{right}} \gets (d_{i,j}^{\text{right}} + d_{i,j+1}^{\text{left}}) / 2$ \Comment{Consolidate with right neighbor (for $j \neq M-1$).}
        \EndFor
        \For{$j \in \{0, \dots, M-1\}$} \Comment{Update overlapping regions in the denoise targets.}
            \State $d_{i,j} \gets \text{setLeft}(d_{i,j}, d_{i,j}^{\text{left}})$
            \State $d_{i,j} \gets \text{setRight}(d_{i,j}, d_{i,j}^{\text{Right}})$
        \EndFor
        \State sample $\varepsilon_j \sim \mathcal{N}(0, I)$ \Comment{Draw new noise for the current SDE step.}
        \State $\{\varepsilon_{i,0}, \dots, \varepsilon_{i,M-1}\} \gets \text{extract}(\varepsilon_j, \{S_0,\dots,S_{M-1}\})$ \Comment{The same overlap region gets the same noise.}
        \For{$j \in \{0, \dots, M-1\}$} \Comment{Apply consolidated exponential denoise update.}
            \State $z_{i-1,j} \gets (\sigma_{i-1}^2/\sigma^2_{i-1})z_{i,j}+(1-\sigma_{i-1}^2/\sigma^2_{i-1})d_{i,j} + (\sigma_{i-1}/\sigma_i)\sqrt{\sigma_{i}^2 - \sigma_{i-1}^2}\varepsilon_{i,j}$ 
        \EndFor
    \EndFor
    \State $z_0 \gets \text{combine}(\{z_{0,0}, \dots, z_{0,M-1}\}, \{S_0,\dots,S_{M-1}\})$ \Comment{Combines denoiser sections into a complete trajectory.}
    \State \Return $z_0$
\EndProcedure
\end{algorithmic}
\end{algorithm}

\subsubsection{Neural architecture}
\label{si:architecture}

The diffusion model denoiser $D_\theta$ is implemented using a U-ViT, which consists of a downsampling and a upsampling stack, each composed of convolutional and axial attention layers. The denoiser takes as inputs noised samples $z_\tau$, the conditioning inputs $\tilde{y}'$, and the noise level $\sigma_\tau$. The output is the climatology-normalized residual sample
\begin{equation}
    \tilde{r}_\text{h} = D_\theta(z_\tau, \sigma_\tau, \tilde{y}').
\end{equation}
The output samples $\tilde{r}_\text{h}$ span $D_\text{lon}$ degrees in longitude, $D_\text{lat}$ degrees in latitude and 7 days in time, leading to tensor shape $84\times4D_\text{lon}\times4D_\text{lat}\times4$ (quarter degree spatial and bi-hourly temporal resolutions), whose dimensions representing time, longitude, latitude and variable dimensions respectively. $z_\tau$ is a noisy version of $\tilde{r}_\text{h}$ and thus share the same size. $\tilde{y}'$ also has the same number of dimensions, but is in lower resolution with shape $7\times D_\text{lon}\times D_\text{lat}\times4$, while $\sigma_\tau$ is a scalar.

\textbf{Encoding.}
The input $\tilde{y}'$ is merged with the noisy sample $z_\tau$. We first apply an encoding block
\begin{equation}
    \begin{aligned}
        h_{\tilde{y}'} = \texttt{Conv2D}(192, 3, 1) \circ\texttt{SiLU}\circ\texttt{LN} \circ \texttt{Conv2D}(4, 7, 1) \circ \texttt{Interp} \circ \tilde{y}',
    \end{aligned}
\end{equation}
which first transfers $\tilde{y}'$ to the same shape as $z_\tau$ through interpolation (cubic in space and nearest neighbor in time), followed by a layer normalization (LN), sigmoid linear unit (SiLU) activation and a spatial convolutional layer (parameters inside the brackets indicate output feature dimension, kernel size and stride respectively) that encode the input into latent features. The latent features are concatenated with $z_\tau$ in the channel dimension and projected by a convolutional layer to form the input to the subsequent downsampling stack:
\begin{equation}
    h = \texttt{Conv2D}(128, 3, 1) \circ \texttt{Concat}([z_\tau, h_{\tilde{y}'}]).
\end{equation}

\textbf{Downsampling stack.}
The downsampling stack consists of a sequence of levels, each at a coarser resolution than the previous. Each level, indexed by $i$, further comprises a strided convolutional layer that applies spatial downsampling
\begin{equation}
    h^{\text{down}}_{i,0} = \texttt{Conv2D}(c_i, 3, q_i) \circ h_{i-1}^{\text{down}},
\end{equation}
followed by 4 residual blocks defined by
\begin{equation}
    \label{eq:res_block}
    \begin{aligned}
        h_{i,j}^{\text{down}}  = h_{i,j-1}^{\text{down}} & + \texttt{Conv2D}(c_i, 3, 1)
        \circ \texttt{SiLU}
        \circ \texttt{FiLM}(\sigma_\tau) \\
        & \circ \texttt{LN} 
        \circ \texttt{Conv2D}(c_i, 3, 1) 
        \circ \texttt{SiLU}
        \circ \texttt{LN}
        \circ h_{i,j-1}^{\text{down}}
    \end{aligned}
\end{equation}
where $j$ denotes the index of the residual block. \texttt{FiLM} is a linear modulation layer
\begin{equation}
    \begin{gathered}
        \texttt{FiLM}(x; \sigma_\tau) = (1.0 + \texttt{Linear}\circ\texttt{FourierEmbed}(\sigma_\tau)) \cdot x + \texttt{Linear}\circ\texttt{FourierEmbed}(\sigma_\tau), \\
        \texttt{FourierEmbed}(\sigma_\tau) = \texttt{Linear} \circ \texttt{SiLU} \circ \texttt{Linear} \circ [\cos(\alpha_k\sigma_\tau), \sin(\alpha_k\sigma_\tau)],
    \end{gathered}
\end{equation}
where $\alpha_k$ are non-trainable embedding frequencies evenly spaced on a logarithmic scale. 

At higher downsampling levels (corresponding to lower resolutions), we additionally apply a sequence of axial multi-head attention (MHA) layers along each dimension (both spatial and time) at the end of each block, defined by
\begin{equation}\label{eq:axial_attn}
    h_i^{\text{down}} = h_i^{\text{down}} +
    \texttt{Linear}(c_i)
    \circ \texttt{MHA}(k)
    \circ \texttt{LayerNorm}
    \circ \texttt{PosEmbed}(k)
    \circ h_{i}^{\text{down}},
\end{equation}
where $k$ denotes the axis over which attention is applied. The fact that attention is sequentially applied one dimension at a time ensures that the architecture scales favorably as the input dimensions increase.

The outputs from each block are collected and fed into the upsampling stack as skip connections, similar to the approach used in classical U-Net architectures.

\textbf{Upsampling stack.}
The upsampling stack can be considered the mirror opposite of the downsampling stack - it contains the same number of levels and residual blocks. At each level, it first adds the corresponding skip connection in the upsampling stack:
\begin{equation}%
    h_{i}^{\text{up}} = h_{i}^{\text{up}} + h_{i}^{\text{down}},
\end{equation}
followed by the same residual and attention blocks defined in~\eqref{eq:res_block} and~\eqref{eq:axial_attn}. At the end of the level, we apply an upsampling block defined by
\begin{equation}
    h^{\text{up}}_i = \texttt{Conv2D}(c_i, 3, 1)
    \circ \texttt{channel2space} 
    \circ \texttt{Conv2D}(c_i q_i^2, 3, 1) 
    \circ h_i^{\text{up}},
\end{equation}
where the \texttt{channel2space} operator expands the $c_iq_i^2$ channels into $q_i\times q_i\times c_i$ blocks locally, effectively increasing the spatial resolution by $q_i$.

\textbf{Decoding.}
We apply a final block to the output of the upsampling stack:
\begin{equation}
    x_\text{out} = \texttt{Conv2D}(4, 3, 1) 
    \circ \texttt{SiLU} 
    \circ \texttt{LayerNorm} 
    \circ h^{\text{up}}_0.
\end{equation}

\textbf{Preconditioning.} As suggested in~\cite{karras2022elucidating}, we \emph{precondition} $D_\theta$ by writing it in an alternative form
\begin{equation}
    D_\theta(z_\tau, \sigma_\tau, \tilde{y}') = c_\text{skip}(\sigma_\tau)z_\tau + c_\text{out}(\sigma_\tau)F\left(c_\text{in}(\sigma_\tau)z_\tau, c_\text{noise}(\sigma_\tau), \tilde{y}'\right),
\end{equation}
where $F$ is the U-ViT architecture described above and 
\begin{equation}
    c_\text{skip} = \frac{1}{1+\sigma^2_\tau}; \quad c_\text{out} = \frac{\sigma_\tau}{\sqrt{1+\sigma^2_\tau}}; \quad c_\text{in} = \frac{1}{\sqrt{1+\sigma^2_\tau}}; \quad c_\text{noise} = 0.25\log{\sigma_\tau},
\end{equation}
such that the input and output of $F$ is approximately normalized.

\subsubsection{Hyperparameters}
\label{si:diffusion_hyp}

Table~\ref{table:diffusion_hyp} shows the set of hyperparameters used for the denoiser architecture, as well as those applied during the training and sampling phases of the diffusion model. We also include the optimization algorithm, learning rate scheduler and weighting for minimizing \eqref{eq:dfn_train_loss}.

\begin{table}[t]
\centering
\footnotesize
\caption{Hyperparameters for the super-resolution model.}
\label{table:diffusion_hyp}
{
\renewcommand{\arraystretch}{1.} %
\begin{tabular}{p{13em}p{20em}}
\hline
\multicolumn{2}{l}{Denoiser architecture} \\ \hline
Output shape   & $84\times240\times120\times4$ (CONUS); $84\times360\times180\times6$ (Atlantic)                                                                        \\
Time span & 7 days \\
Spatial downsampling ratios   & [3, 2, 2, 2] (CONUS); [3, 3, 2, 2] (Atlantic)                                                                                            \\
Residual blocks  & [4, 4, 4, 4]                                                                                                       \\
Hidden channels                 & [128, 256, 384, 512]                                                                                     \\ 
Use attention layers   & [\texttt{False}, \texttt{False}, \texttt{True}, \texttt{True}]                                           \\ 
Trainable parameters & around 150 million (both CONUS and Atlantic) \\ \hline
\multicolumn{2}{l}{Training} \\ \hline
Device                           & TPUv5p, $2\times4\times4$                                                                                \\
Duration                         & 300,000 steps                                                                                               \\
Batch size                       & 128 (with data parallelism)                                                                              \\
Learning rate                    & cosine annealed (peak=$1\times10^{-4}$, end=$1\times10^{-7}$), 1,000 linear warm-up steps                  \\
Gradient clipping                & max norm = 0.6                                                                                           \\
Noise sampling                   & $\sigma_\tau\sim$ \texttt{LogUniform}(min=$1\times10^{-4}$, max=80)                                      \\
Noise weighting ($\lambda_\tau$) & $1 + 1/\sigma_{\tau}^2$                                                                   \\ 
Condition dropout ($p_u$)  & 0.15  \\ \hline
\multicolumn{2}{l}{Inference}                                                                            \\ \hline
Device                           & TPUv5e, $4\times4$                                                                                         \\
Noise schedule                   & $\sigma_\tau=\frac{\tan{(3\tau-1.5)}-\tan{(-1.5)}}{\tan{(1.5)} - \tan{(-1.5)}}\cdot80$, $\tau\sim[0, 1]$ \\
SDE solver type                  & 1st order exponential \\
Solver steps                     & $(\sigma_{\text{max}}^{1/7} + \frac{i}{255}(\sigma_{\text{min}}^{1/7} - \sigma_{\text{max}}^{1/7}))^{7}$ \\
CFG strength ($g$)               & 1.0                                                                                                      \\ 
Overlap                          & 1 day (12 time slices) \\ 
\# of days coherently denoised & 97 days \\ \hline
\end{tabular}
}
\end{table}

\subsubsection{Training, evaluation, and test data}

The super-resolution stage is trained \emph{independently of the debiasing stage}, using perfectly time-aligned ERA5 data samples at the input (1.5-degree, daily) and output (0.25-degree, bi-hourly) resolutions.

Training is conducted on continuous 7-day windows randomly selected in the training range, with each day beginning at 00:00 UTC. Spatially, the model super-resolves a rectangular patch of fixed size. Consistent with the debiasing step, data from 1960–1999 is used for training, 2000–2009 for evaluation, and 2010–2019 for testing.

\subsubsection{Computational cost}
\label{si:diffusion_cost}

The diffusion model is trained on TPUv5p hosts, utilizing a total of 32 cores, which takes approximately 3 days. For sampling, 16 TPUv5e cores are employed in parallel. Each core denoises a single 7-day window, collectively generating a 97-day temporally consistent long sample\footnote{Our parallel strategy yields 97 days, calculated as: number of cores \{16\} × (model days \{7\} - overlap \{1\}) + overlap \{1\}. This means increasing the number of cores effectively extends the total sample length. Alternatively, sequential sampling of 7-day windows can be performed, where sample length is independent of the number of cores and scales with inference time.}. Excluding JAX compile time, a one-time overhead that makes subsequent realizations significantly more efficient, each sample requires about 3 minutes to complete. The temporal length of the generated samples scales linearly with the number of TPU cores used, while clock time remains relatively constant.  At a cost of estimated $\$1.2$ USD per hour of the current market rate, the super-resolution step incurs a cost of $\$0.08$ USD per sample day in a $[60^\circ, 30^\circ]$ region. For 100 ensemble members over 10 years (3 months per year, 8 samples per ensemble member), the estimated total inference cost is approximately $\$61,440$. This cost can be further reduced with accelerated sampling algorithms and other engineering optimization.

\subsection{\ourname Variants} 
\label{si:ablation_models}

The two-stage design of \ourname enables a ``plug and play'' approach for integrating different bias correction and super-resolution components.  We describe two such components below, which we use as ablation studies to examine the effectiveness of our bias connection component, introduced in ~\ref{si:debias}.

\subsubsection{Direct Super-Resolution (SR)}
\label{si:sr}

We can examine how well a super-resolution operation, optimized on the reanalysis ERA5 can overcome the bias in the low-resolution climate data. We term this method of downscaling as SR, with the generative super-resolution described in ~\ref{si:upsampling} being directly applied on LENS2.

\subsubsection{Quantile Mapping Super-Resolution (QMSR)} \label{si:qmsr}

We have also experimented with the quantile mapping component of BCSD (described in ~\ref{si:bcsd}), with a bit adaptation, as a debiasing procedure, followed by \ourname's super-resolution operation.  We term this approach as QMSR. The adaptation we need is to add back the mean of the downsampled data:
\begin{equation}
    y_\text{qm} = \frac{y - \texttt{clim\_mean}[y]}{\texttt{clim\_std}[y]} \cdot \texttt{clim\_std}[C'x] + \texttt{clim\_mean}[C'x].
\end{equation}
The resulting output $y_\text{qm}$ retains the low spatial resolution and can serve as the input for our diffusion-based upsampling model. This is the ``QM'' baseline referred to in Table~\ref{table:bias_and_wass}.

For both variants, during the generative super-resolution steps, the inputs and outputs are respectively normalized and denormalized in the same way as described by~\eqref{eq:sr_input_normalize} and~\eqref{eq:sr_output_normalize}, where the normalization statistics are derived from ground truth low-resolution ERA5. (For SR, experiments with input normalization using statistics of the LENS2 dataset led to worse evaluation results across almost all metrics. )

\clearpage
\section{Ablation studies: model selection and design choices} \label{si:ablations} 

We study the sensitivity of \ourname\!' to a few design choices and implementation details. \rev{Earlier studies provide further insight into variations of diffusion-based super-resolution, see \S5.1.3 in~\cite{wan2024statistical}, and \cite{Lopez-Gomez25a}}. The main ablation studies and findings in this section cover the following:
\begin{itemize}
\item Reference periods used for training \rev{the debiasing stage} (\ref{si:ablations_years}). \rev{Using training data from recent reference periods, which is better constrained by satellite observations,} results in better models. More data improves the representation of extremes.
\item Length of the debiasing sequence (\ref{si:ablations-length-debiasing}). Longer debiasing sequences lead to improvements in most statistics.
\item Number of debiased variables being modeled (\ref{si:sec:debiasing-variables}). Debiasing 10 variables improves \ourname's ability to capture TC statistics compared to variants with 4 or 6 debiased variables. Since computational costs scale with the number of modeled variables, we leave the selection of an optimal variable set to future work.
\item Number of training steps (\ref{si:ablations-training-steps}). Additional training steps beyond 300k lead to an overestimation of the number of tropical cyclones, possibly due to overfitting.
\item Number of LENS2 ensemble members used for training (\ref{si:sec:ensemble_members}). While LENS2 contains 100 members, we train on a \rev{small} subset and evaluate on the full ensemble. Including multiple members in the training set enhances performance, though benefits saturate beyond four members.

\item \rev{The data coupling strategy used to train the debiasing stage (\ref{si:ablations_coupling}). Coupling samples with similar climatologies (e.g., samples from the same day of the year, or from adjacent years) yields more statistically accurate results.}

\item \rev{The training period for the super-resolution stage (\ref{si:ablations_years_sr}). \ourname is largely insensitive to this change, provided enough data.}

\item \rev{The length of the temporal sequence and the stitching strategy in the super-resolution stage (\ref{si:super_resolution_temporal_length}). Temporal coherence through domain decomposition improves the statistics of spatio-temporal phenomena, more so for long-lived events.}

\item \rev{Importance of residual modeling on super-resolution} (\ref{si:residual_modeling}). Modeling the fine-grained deviations from the debiasing stage leads to improved results for most variables and metrics, compared to predicting the full high-resolution fields.

\end{itemize}
Throughout the ablation studies, the evaluation period (2010 - 2019) remains unchanged \rev{to ensure consistent comparison}.

\subsection{Training period for the debiasing stage} \label{si:ablations_years}

\begin{table}[th!]
\centering
\footnotesize
\caption{Effect of training data period on the mean absolute bias, Wasserstein distance, and absolute error of the $99^{\text{th}}$ percentile for the summers (June-July-August) of 2010-2019 in CONUS. The precise definitions of the metrics are included in Section ~\ref{si:evaluation_metrics}.}
{
\setlength\tabcolsep{2.25pt}
\label{table:ablation_years}
\begin{tabular}{|lccccccc|}
\hline
\multirow{2}{*}{Variable} & 60s     & 70s   & 80s   & 90s   & 60s-90s & 70s-90s & 80s-90s \\ 
\multirow{2}{*}{}         &  \multicolumn{7}{c|}{Mean Absolute Bias, $\downarrow$}          \\
Temperature (K)           & 0.54 &	0.48 &	0.53 &	\textbf{0.39} &	0.53 &	0.48 &	0.41 \\
Wind speed (m/s)          & 0.23 &	0.24 &	0.19 &	\textbf{0.16} &	0.17 &	0.19 &	0.19 \\
Specific humidity (g/kg)  & 0.40 &	0.35 &	0.47 &	0.37 &	0.50   & 0.43 &	\textbf{0.31} \\
Sea-level pressure (Pa)   & 30.07 &	57.46 &	51.49 &	\textbf{28.33} &	43.94 &	54.71 &	39.92 \\
Relative humidity (\%)    & 2.24 &	1.88 &	2.84 &	2.03 &	2.08 &	1.93 &	\textbf{1.71} \\
Heat index (K)            & 0.59 &	0.53 &	0.55 &	\textbf{0.46} &	0.65 &	0.59 &	0.47 \\ \hline
\multirow{2}{*}{}         & \multicolumn{7}{c|}{Mean Wasserstein Distance, $\downarrow$}          \\            
Temperature (K)           & 0.61	& 0.54	& 0.59	& \textbf{0.47}	& 0.59	& 0.55	& \textbf{0.47} \\
Wind speed (m/s)          & 0.28	& 0.29	& 0.22	& 0.21	& \textbf{0.20}	& 0.22	& 0.22 \\
Specific humidity (g/kg)  & 0.48	& 0.43	& 0.51	& 0.42	& 0.53	& 0.47	& \textbf{0.36} \\
Sea-level pressure (Pa)   & 53.32	& 71.81	& 63.51	& \textbf{44.79}	& 50.85	& 64.48	& 52.09 \\
Relative humidity (\%)    & 2.62	& 2.32	& 3.1	& 2.27	& 2.41	& 2.29	& \textbf{2.1} \\
Heat index (K)            & 0.67	& 0.6   & 0.61	& \textbf{0.53}	& 0.70	& 0.65	& \textbf{0.53} \\ \hline
\multirow{2}{*}{} & \multicolumn{7}{c|}{Mean Absolute Error, $99^{\text{th}}$ $\downarrow$}         \\
Temperature (K)           & 1.02	& 0.83	& 0.87	 & \textbf{0.60}	 & 0.64	 & 0.68	 & 0.61 \\
Wind speed (m/s)          & 0.85	& 0.74	& 0.61	 & 0.56	 & \textbf{0.39} & 0.48 & 0.48 \\
Specific humidity (g/kg)  & 0.83	& 0.68	& 0.58	 & 0.41	 & \textbf{0.4}	 & 0.42	 & 0.45 \\
Sea-level pressure (Pa)   & 128.36	& 80.61	& 107.09 & 92.12 & \textbf{60.31} & 69.89 & 77.99 \\
Relative humidity (\%)    & 2.39	& 2.27	& 2.10	 & 1.99	 & 1.90	 & 1.93	 & \textbf{1.87} \\
Heat index (K)            & 1.2     & 0.96	& 0.92	 & \textbf{0.67}	 & 0.77	 & 0.81	 & 0.68 \\ \hline
\end{tabular}
}
\end{table}

\rev{We evaluate training period sensitivity using 2010–2019 CONUS summer and downscaled North Atlantic TC statistics. We consider models trained on individual decades from the 1960s to the 1990s, as well as models trained over multiple decades spanning the period 1960--2000.}

\rev{Overall, the results underscore the critical importance of training data quality and diversity. Table \ref{table:ablation_years} indicates that training exclusively on reanalysis data prior to 1979—which is significantly less constrained by satellite observations—substantially degrades model performance. This decline is particularly pronounced in wind speed and humidity extremes. These findings align with previous studies suggesting that training AI weather models on pre-1979 reanalysis data does not lead to improved skill \cite{Kochkov2024, Price2025}.}

\begin{figure}[th!]
\centering
\includegraphics[width=0.99\textwidth]{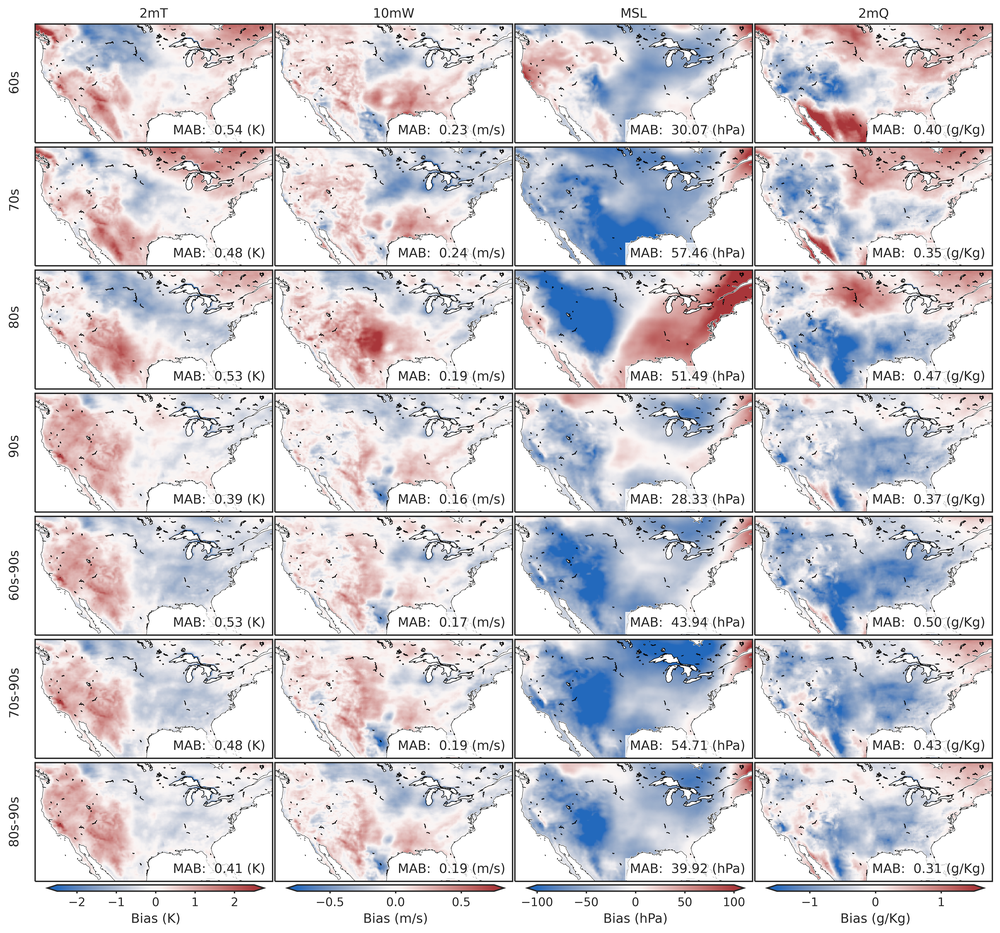}
\caption{\textbf{Summertime bias of \ourname over CONUS.} Bias of multiple fields  during summers (June-August) of 2010-2019, for \ourname models trained using data from different reference periods.
}\label{fig:si_ablation_conus_bias_period}
\end{figure}

\begin{figure}[tbh!]
\centering
\includegraphics[width=0.99\textwidth]{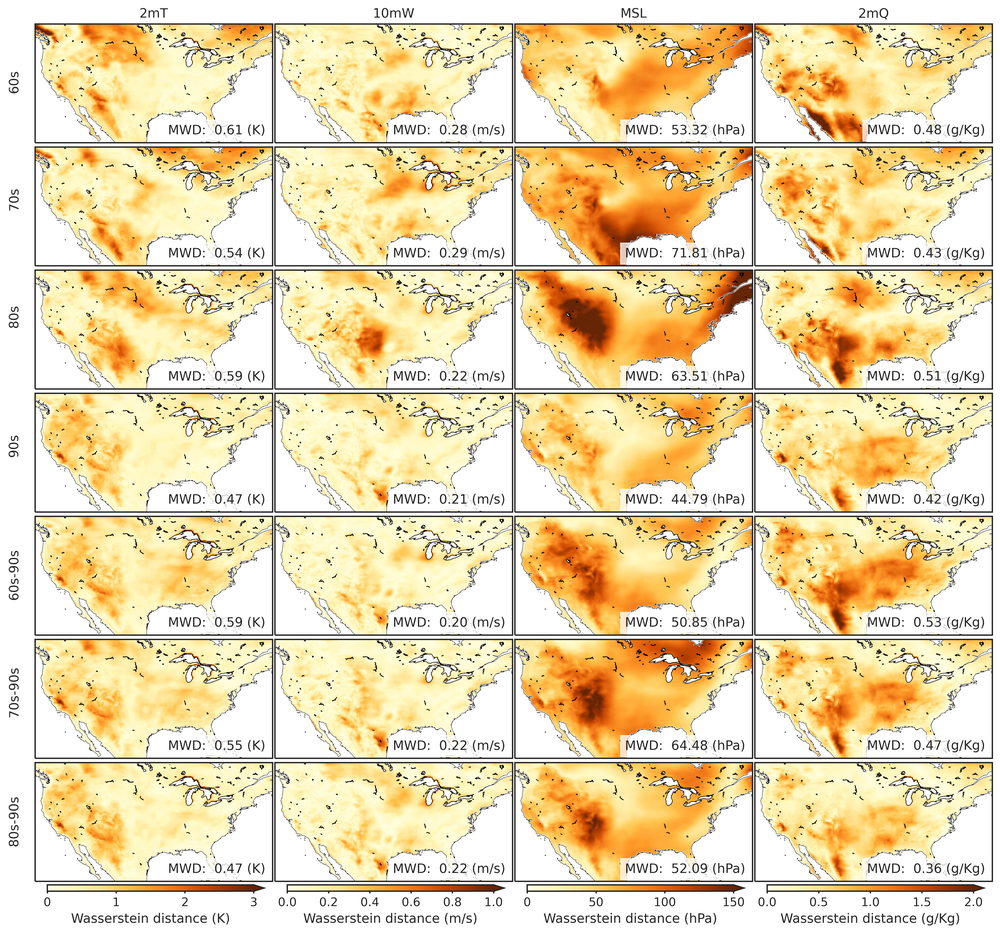}
\caption{\textbf{Summertime Wasserstein distance over CONUS.} Pointwise Wasserstein distance (see~\ref{si:wass}) during summers (June-August) of 2010-2019, for \ourname models trained using data from different reference periods. 
}\label{fig:si_ablation_conus_wass_period}
\end{figure}

\begin{figure}[th!]
\centering
\includegraphics[width=0.99\textwidth]{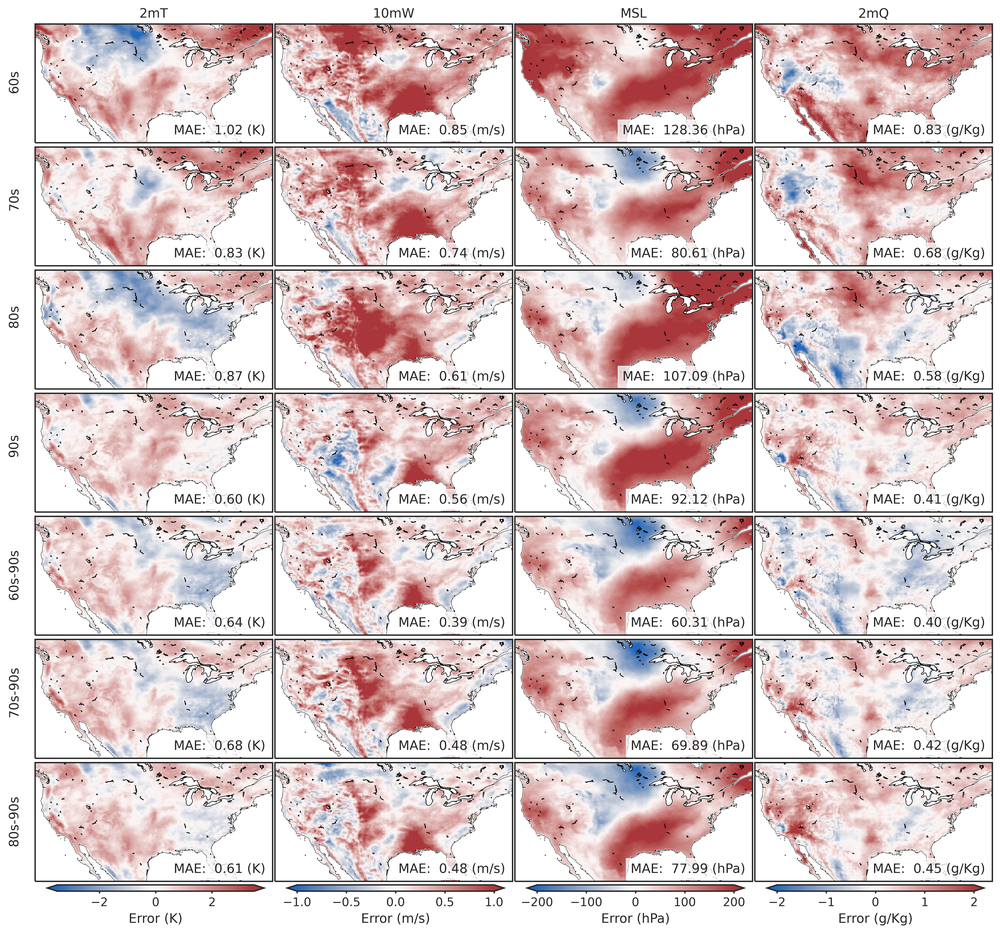}
\caption{\textbf{Summertime $99^{th}$ percentile error over CONUS}. Pointwise error in the $99^{th}$ percentile of multiple fields during summers (June-August) of 2010-2019, for \ourname models trained using data from different reference periods. 
}\label{fig:si_ablation_conus_99_period}
\end{figure}

Table \ref{table:ablation_years} also demonstrates that leveraging more training data improves the representation of extremes. \rev{However, extending the training period to include pre-1979 data leads to higher bias and Wasserstein distances, indicating a trade-off between data diversity and quality. Consequently, the 1980--1999 period yields the best aggregate performance.} These findings are further supported by Figs.~\ref{fig:si_ablation_conus_bias_period}-\ref{fig:si_ablation_conus_99_period}, which illustrate the geographical distribution of the bias, Wasserstein distance, and the $99^{\text{th}}$ percentile, respectively.

\begin{figure}[tbh!]
\centering
\includegraphics[width=0.95\textwidth]{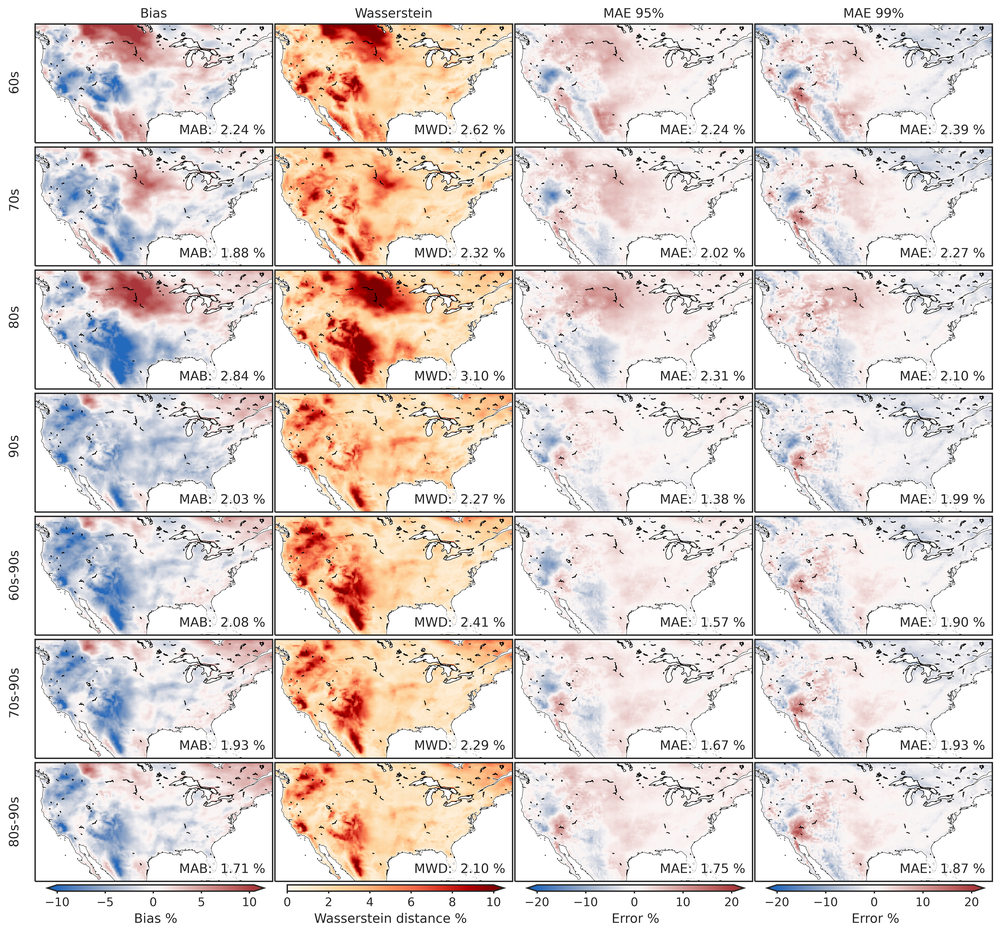}
\caption{\textbf{Metrics for relative humidity over CONUS.} Metrics for the relative humidity, a derived variable, over CONUS during the summer (June-August) for the evaluation period 2010-2019. We include bias, Wasserstein error, error of the $95^{\text{th}}$ and $99^{\text{th}}$ percentiles for \ourname trained using data from different reference periods. 
}\label{fig:si_ablation_conus_rh_period}
\end{figure}

\begin{figure}[tbh!]
\centering
\includegraphics[width=0.95\textwidth]{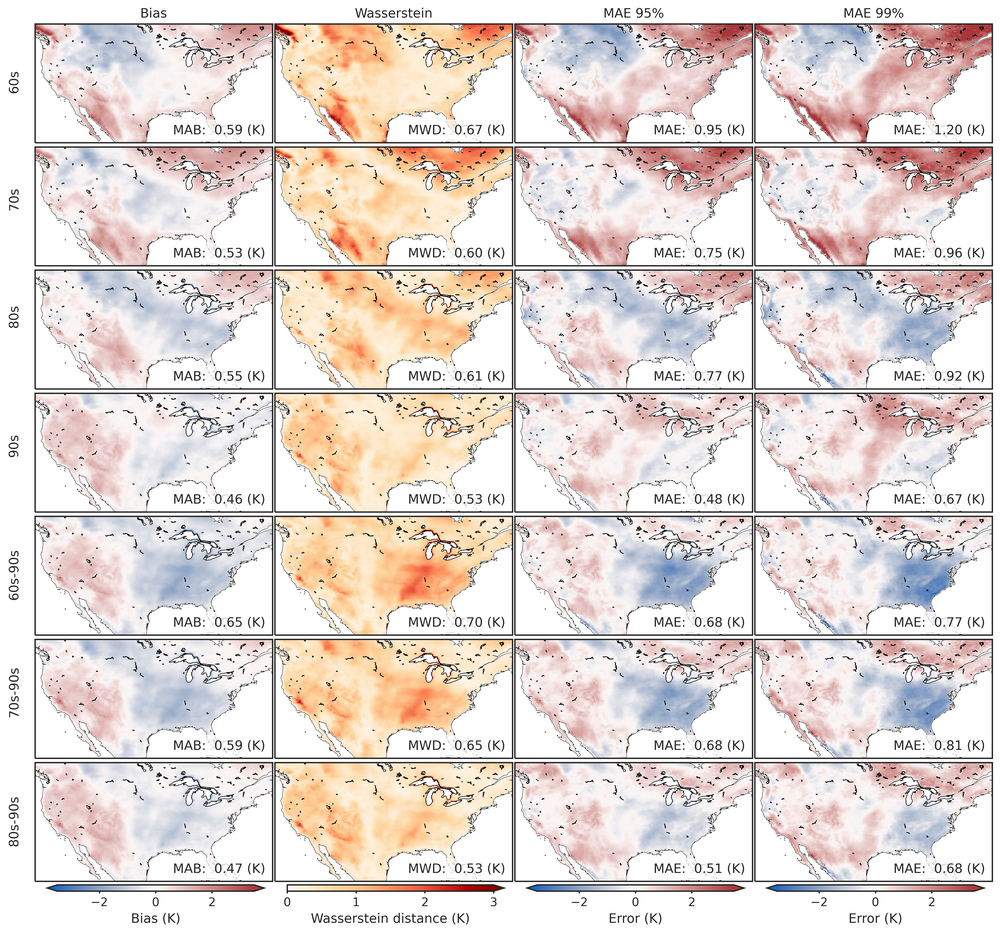}
\caption{\textbf{Metrics for the heat index over CONUS.} Metrics for the heat index, a derived variable, over CONUS during the summer (June-August) for the evaluation period 2010-2019. We include bias, Wasserstein error, error of the $95^{\text{th}}$ and $99^{\text{th}}$ percentiles for \ourname trained using data from different training periods.
}\label{fig:si_ablation_conus_hi_period}
\end{figure}

\rev{Additional results for relative humidity and heat index are shown in Figs.~\ref{fig:si_ablation_conus_rh_period} and \ref{fig:si_ablation_conus_hi_period}. Notably, Fig.~\ref{fig:si_ablation_conus_hi_period} reveals that models trained solely on pre-1979 data exhibit strong biases in heat index extremes at high latitudes; these biases are significantly mitigated in models trained on post-1979 data.}

\begin{figure}[tbh!]
\centering
\includegraphics[width=0.99\textwidth]{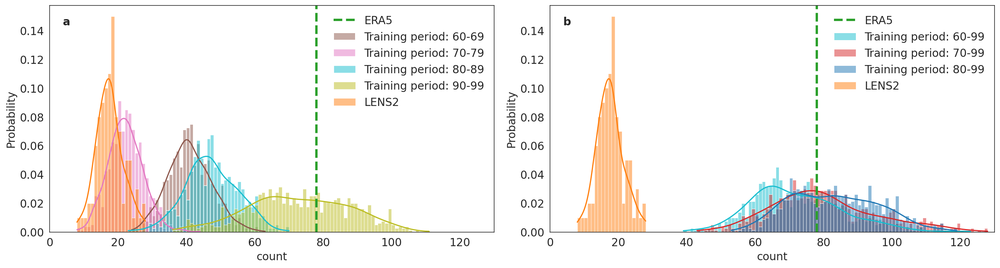}
\caption{\textbf{Distribution of TC occurrences.} Distribution of the number of TCs detected by TempestExtremes in the North Atlantic during the peak hurricane season (August–October), 2010–2019. \rev{\textbf{a}. Distribution of TC counts for different decadal training periods. \textbf{b}. Distribution of the TC counts for training data periods of varying length. TCs from all \ourname variants were calibrated separately following SI Section \ref{si:tc_calibration}}.
}\label{fig:si_ablation_tc_counts}
\end{figure}

\begin{figure}[tbh!]
\centering
\includegraphics[width=0.85\textwidth]{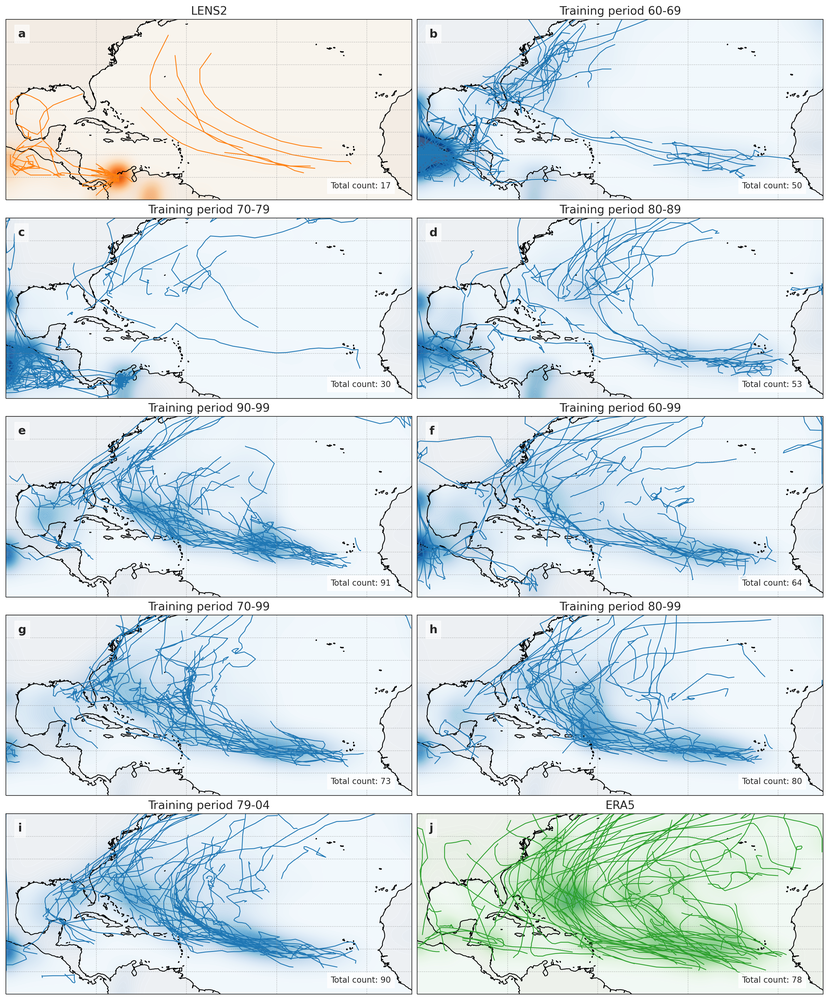}
\caption{\textbf{TC tracks and their density.} \rev{TC tracks for a single climate projection overlaying the ensemble TC track density for the peak North Atlantic hurricane season (August–October), 2010–2019. \textbf{a}. LENS2. \textbf{b}-\textbf{i}. \ourname variants trained on different periods.. \textbf{f}. ERA5 tracks and their density. TCs from all \ourname variants were calibrated following SI Section \ref{si:tc_calibration}}.
}\label{fig:si_ablation_tc_tracks}
\end{figure}

\rev{Data quality and diversity are also critical for the realistic representation of TCs. As shown by the TempestExtremes detection counts in Fig.~\ref{fig:si_ablation_tc_counts}, a single decade of training data is typically inadequate for learning accurate TC representations in the North Atlantic (Fig.~\ref{fig:si_ablation_tc_counts}a). Performance improves when training on high-activity decades like the 1990s \cite{Goldenberg_2001}, consistent with recent evidence that AI weather models only faithfully represent TCs when they are sufficiently prevalent in the training set \cite{sun_TCs_pnas_2025}. This requirement is further illustrated by the example tracks in Fig.~\ref{fig:si_ablation_tc_tracks}, which demonstrate the importance of data volume for capturing TC statistics.}

\begin{figure}[tbh!]
\centering
\includegraphics[width=0.99\textwidth]{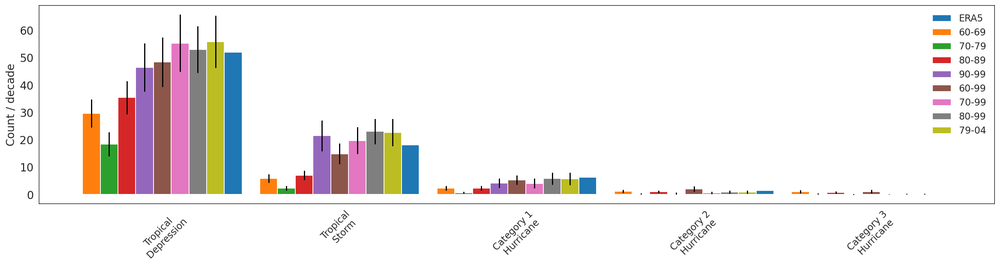}
\caption{\textbf{Saffir-Simpson scale distribution of TCs.} \rev{Distribution of the intensity of TCs detected by TempestExtremes in the North Atlantic during the peak hurricane season (August–October), 2010–2019. Results shown for GenFocal models trained on different periods, and for ERA5. Error bars denote the ensemble standard deviation. TCs from all GenFocal variants were calibrated separately following SI Section \ref{si:tc_calibration}}.
}\label{fig:si_ablation_tc_categories_periods}
\end{figure}

\rev{In contrast, training on multi-decadal datasets yields realistic TC counts and tracks for all considered training periods (Fig.~\ref{fig:si_ablation_tc_counts}b, Fig.~\ref{fig:si_ablation_tc_tracks}). However, models trained on early reanalysis data—unconstrained by satellite-based remote sensing—exhibit a slight underestimation of TCs across all categories (Fig.~\ref{fig:si_ablation_tc_categories_periods}).}

\clearpage
\subsection{Length of the debiasing sequence}
\label{si:ablations-length-debiasing}

This section shows that harnessing spatiotemporal correlations in the input data leads to a reduction in distribution matching errors, by evaluating the sensitivity of \ourname to the number of consecutive days debiased simultaneously. We retain the same architecture and number of trainable parameters as the model reported in the main text.

\begin{figure}[th!]
\centering
\includegraphics[width=0.95\textwidth]{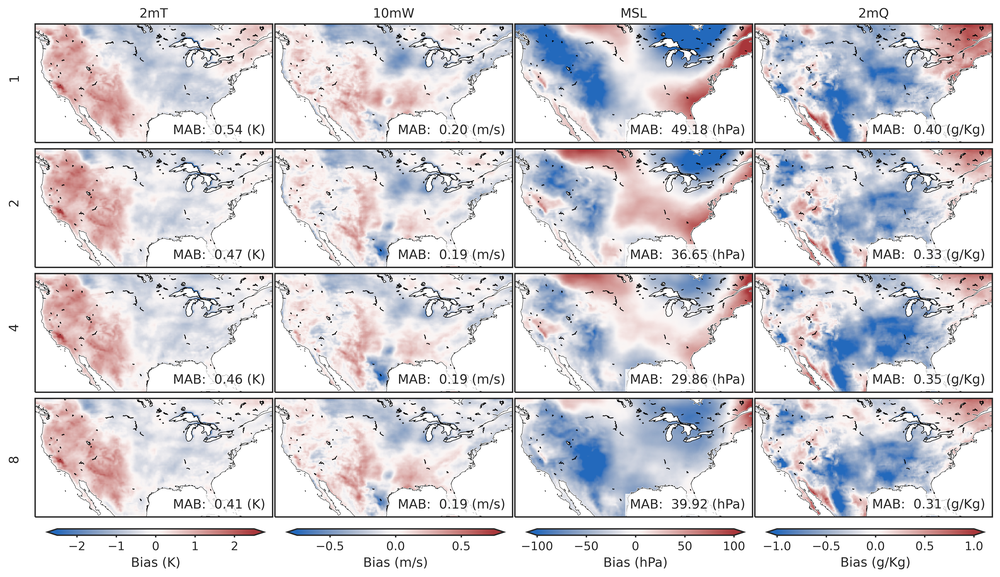}
\caption{\textbf{Downscaled variable biases.} Biases of downscaled variables, over CONUS during the summer (June-August) for the evaluation period 2010-2019 for \ourname trained with different debiasing sequence lengths. 
}\label{fig:si_ablation_conus_bias_seq_length}
\end{figure}

\begin{figure}[tbh!]
\centering
\includegraphics[width=0.95\textwidth]{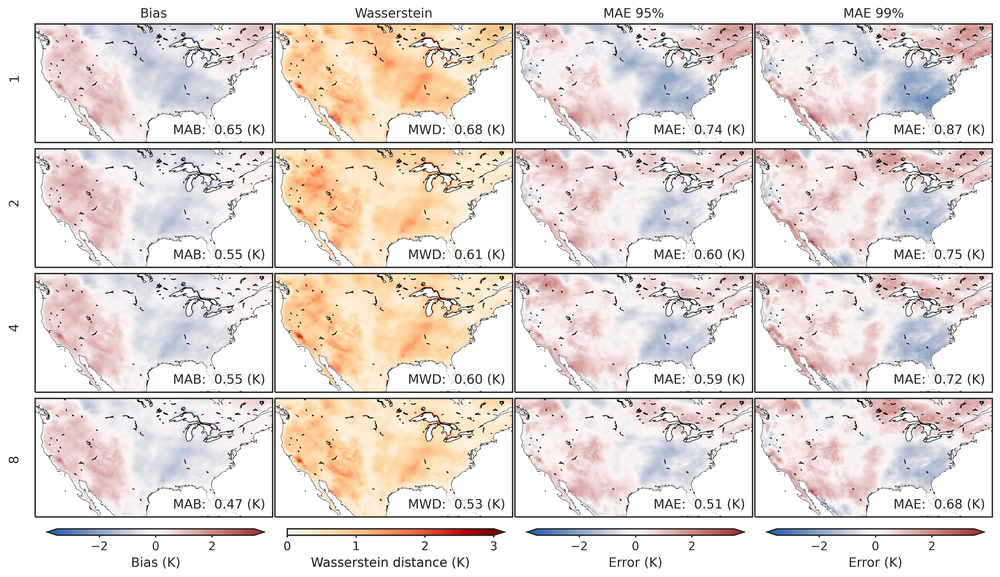}
\caption{\textbf{Spatial distribution of heat index errors.} Spatial distribution of statistical modeling errors for the heat index, one of the derived variables, over CONUS during the summer (June-August) for the evaluation period 2010-2019. We include bias, Wasserstein error, error of the $95^{\text{th}}$ and $99^{\text{th}}$ percentiles for \ourname trained with different debiasing sequence lengths. 
}\label{fig:si_ablation_conus_hi_seq_length}
\end{figure}

Table \ref{table:ablation_bias_days} summarizes the statistical errors of \ourname models with different debiasing sequence lengths. Longer debiasing sequences lead to improvements in most statistics. Fig.~\ref{fig:si_ablation_conus_bias_seq_length} shows the spatial distribution of biases for the directly modeled variables.  Fig.~\ref{fig:si_ablation_conus_hi_seq_length} shows the geographical distribution of the metrics for the heat index. In both, we observe that the geographical distribution of the errors is similar across debiasing sequence lengths, with an overall error reduction for longer sequences.

\begin{figure}[tbh!]
\centering
\includegraphics[width=0.95\textwidth]{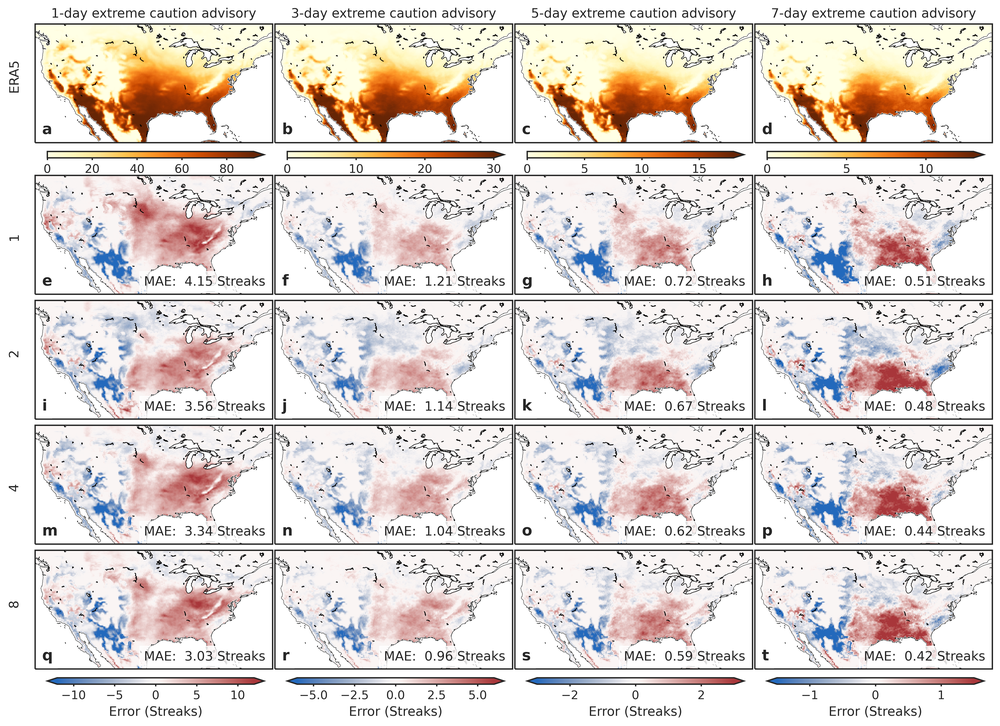}
\caption{\textbf{Spatial distribution of heat streak errors.} Spatial distribution of statistical modeling errors in the number of heat-streaks per year for extreme caution advisory considering different debiasing sequence lengths (in days). We show the ground truth (ERA5)(\textbf{a-d}), and the pointwise errors of \ourname with different lengths of the debiased sequence.
}\label{fig:si_conus_heatwaves_extreme_caution_ablation_seq_length}
\end{figure}

Fig.~\ref{fig:si_conus_heatwaves_extreme_caution_ablation_seq_length} shows the bias in the projected number of extreme caution advisory periods per year, for periods of varying length. We observe that increasing the length of the debiasing sequence uniformly decreases the bias in the number of predicted heat streaks of 1 to 7 days.

\begin{table}[tbh!]
\centering
\footnotesize
\caption{Effect of debiasing sequence length on mean absolute bias, mean Wasserstein distance, and mean absolute error in the $99^{\text{th}}$ percentile for different variables during summers of 2010-2019 (June-July-August) over CONUS. The metrics are defined in \ref{si:evaluation_metrics}.}
{
\setlength\tabcolsep{1.7pt}
\label{table:ablation_bias_days}
\begin{tabular}{|lcccc|cccc|cccc|}
\hline
\multirow{2}{*}{Variable} & \multicolumn{4}{c|}{Mean} & \multicolumn{4}{c|}{Mean} & \multicolumn{4}{c|}{Mean}\\
 & \multicolumn{4}{c|}{Absolute Bias $\downarrow$} 
 & \multicolumn{4}{c|}{Wasserstein Distance $\downarrow$}
 & \multicolumn{4}{c|}{Absolute Error, $99^{\text{th}}$ $\downarrow$}  \\
                          & 1       & 2     & 4     & 8              & 1    & 2     & 4     & 8     & 1     & 2     & 4      & 8  \\ \hline
Temperature (K)           & 0.54	& 0.47	& 0.46	& 0.41	& 0.57	& 0.52	& 0.52	& 0.47	& 0.69	& 0.66	& 0.64	& 0.61 \\
Wind speed (m/s)          & 0.2	& 0.19	& 0.19	& 0.19	& 0.22	& 0.22	& 0.22	& 0.22	& 0.41	& 0.44	& 0.46	& 0.48 \\
Specific humidity (g/kg)  & 0.4	& 0.33	& 0.35	& 0.31	& 0.43	& 0.38	& 0.4	& 0.36	& 0.44	& 0.47	& 0.46	& 0.45 \\
Sea-level pressure (Pa)   & 49.18	& 36.65 & 29.86	& 39.92	& 58.08	& 47.72	& 43.36	& 52.09	& 75.57	& 86.13	& 80.84	& 77.99 \\
Relative humidity (\%)    & 1.85	& 1.69	& 1.74	& 1.71	& 2.21	& 2.07	& 2.11	& 2.1	& 1.83	& 1.9	& 1.91	& 1.87 \\
Heat index (K)            & 0.65	& 0.55	& 0.55	& 0.47	& 0.68	& 0.61	& 0.6	& 0.53	& 0.87	& 0.75	& 0.72	& 0.68 \\ \hline
\end{tabular}
}
\end{table}

\begin{figure}[tbh!]
\centering
\includegraphics[width=0.99\textwidth]{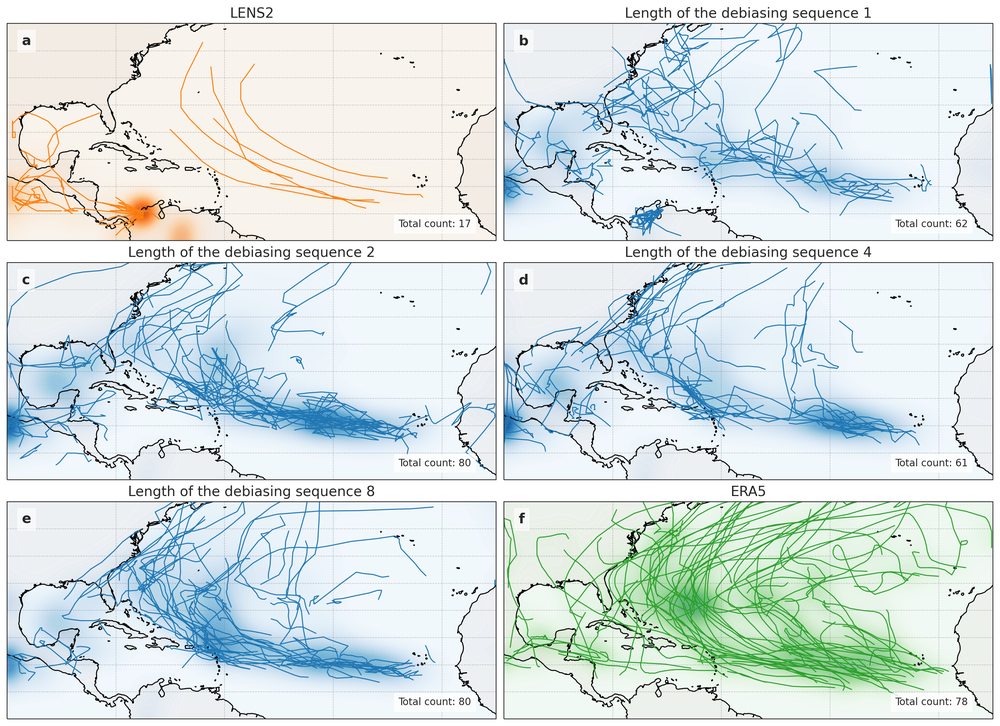}
\caption{\textbf{TC tracks and density.} Tracks and their density for a LENS2 member in the North Atlantic in the time period 2010-2019 (\textbf{a}), and for a downscaled sample from the same member generated using \ourname for different debiasing sequence length (\textbf{b}-\textbf{e}). Also shown are the tracks detected in the reference ERA5 data (\textbf{f}).
}\label{fig:si_ablation_tc_tracks_seq_length}
\end{figure}

\begin{figure}[tbh!]
\centering
\includegraphics[width=0.75\textwidth]{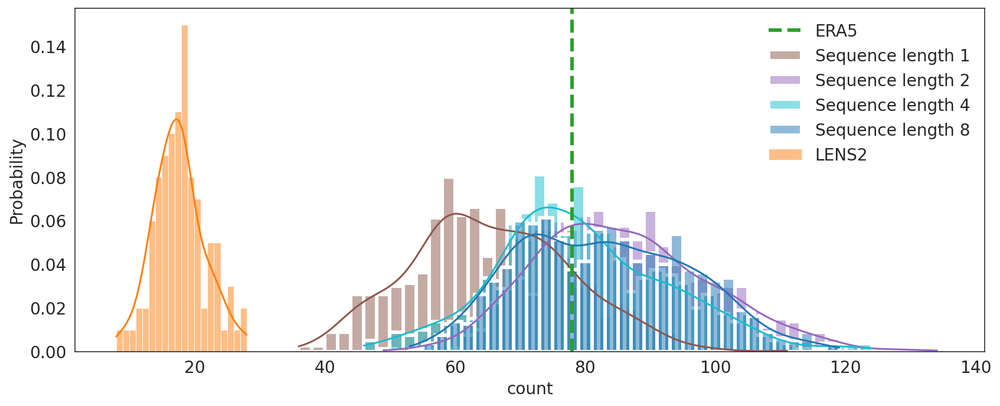}
\caption{\textbf{Distribution of TC frequency.} Distribution of the number of TCs detected by TempestExtremes in the North Atlantic  for the hurricane season August-September-October during 2010-2019, using downscaled data from \ourname models with varying debiasing sequence lengths.
}\label{fig:si_ablation_tc_counts_seq_length}
\end{figure}

Figs.~\ref{fig:si_ablation_tc_tracks_seq_length} and~\ref{fig:si_ablation_tc_counts_seq_length} further show that using longer debiasing sequences also leads to tropical cyclones with more realistic trajectories in the North Atlantic basin. Furthermore, statistics of projected TCs match the observational record better.

\clearpage
\subsection{Number of debiased variables}
\label{si:sec:debiasing-variables}

\begin{table}[tbh!]
\centering
\footnotesize
\caption{Effect of the number of debiased variables on mean absolute bias, mean Wasserstein distance, and mean absolute error in the $99^{\text{th}}$ percentile of the downscaled output for different variables. The metrics are defined in \ref{si:evaluation_metrics}.}
{
\setlength\tabcolsep{2.25pt}
\label{table:bias_and_wass_variables}
\begin{tabular}{|lccc|ccc|ccc|}
\hline
\multirow{2}{*}{Variable} &  \multicolumn{3}{c|}{Mean}
 & \multicolumn{3}{c|}{Mean}
 & \multicolumn{3}{c|}{Mean}\\
 & \multicolumn{3}{c|}{Absolute Bias $\downarrow$} 
 & \multicolumn{3}{c|}{Wasserstein Distance $\downarrow$}
 & \multicolumn{3}{c|}{Absolute Error, $99^{\text{th}}$ $\downarrow$}  \\ 
                          & 4       & 6     & 10                & 4     & 6     & 10                & 4        & 6      & 10         \\ \hline
Temperature (K)           & 0.43    & 0.47  & \textbf{0.41}      & 0.49  & 0.52  & \textbf{0.47}      & 0.64      & 0.63  & \textbf{0.61} \\
Wind speed (m/s)          & \textbf{0.16}    & 0.19  & 0.19      & \textbf{0.2}   & 0.22  & 0.22      & 0.58      & \textbf{0.44}  & 0.48 \\
Specific humidity (g/kg)  & 0.35    & 0.39  & \textbf{0.31}      & 0.41  & 0.43  & \textbf{0.36}      & 0.47      & 0.45  & \textbf{0.45} \\
Sea-level pressure (Pa)   & 36.82   & \textbf{36.67} & 39.92     & 54.14 & \textbf{50.22} & 52.09     & 117.07    & 91.22 & \textbf{77.99} \\
Relative humidity (\%)    & \textbf{1.63}    & 1.7   & 1.71      & \textbf{2.01}  & 2.1   & 2.1       & \textbf{1.84}      & 1.87  & 1.87 \\
Heat index (K)            & 0.51    & 0.58  & \textbf{0.47}      & 0.57  & 0.63  & \textbf{0.53}      & 0.71      & 0.75  & \textbf{0.68} \\   \hline
\end{tabular}
}
\end{table}

We explore the sensitivity to the number of debiased variables by considering two alternative models that use $4$ and $6$ of the variables described in \ref{si:debiasing_data_set}, respectively. The model with $4$ inputs retains the variables to be super-resolved, and the variant with $6$ input variables incorporates the geopotential height at $200$ and $500$ hPa. In all cases, the super-resolution step only takes $4$ variables as inputs.

\begin{figure}[tbh!]
\centering
\includegraphics[width=0.75\textwidth]{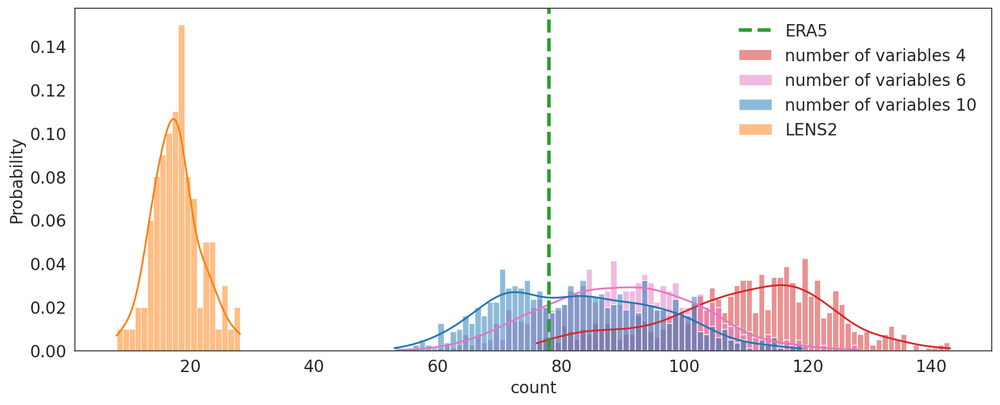}
\caption{\textbf{Distribution of TC frequency.} Distribution of the number of TCs detected by TempestExtremes in the North Atlantic for the hurricane season August-September-October during 2010-2019 for \ourname trained with different number of debiasing variables. 
}\label{fig:si_ablation_tc_counts_num_vars}
\end{figure}

\begin{figure}[tbh!]
\centering
\includegraphics[width=0.95\textwidth]{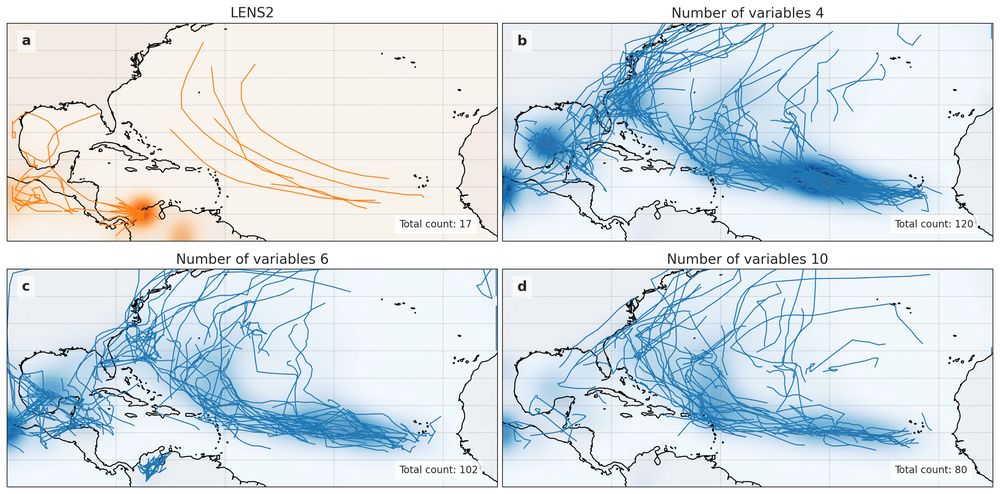}
\caption{\rev{\textbf{Tracks and their density.}} Tracks and their density for a LENS2 member in the North Atlantic in the time period 2010-2020 (\textbf{a}), one of downscaling samples from the same member generated using \ourname trained with different number of debiased variables sets (\textbf{b}-\textbf{d}).
}\label{fig:si_ablation_tc_tracks_num_vars}
\end{figure}

From Table \ref{table:bias_and_wass_variables} we can
see that increasing the number of debiased variables leads to improvements in temperature and specific humidity, but not in wind speed or relative humidity.
Figs.~\ref{fig:si_ablation_tc_counts_num_vars} and ~\ref{fig:si_ablation_tc_tracks_num_vars} show that increasing the number of debiased variables leads to more accurate TC statistics.

\clearpage
\subsection{Number of training steps}
\label{si:ablations-training-steps}

\begin{table}[t!]
\centering
\footnotesize
\caption{Effect of the number of training steps on mean absolute bias, mean Wasserstein distance, and mean absolute error in the $99^{\text{th}}$ percentile for different variables. The precise definitions of the metrics are included in \ref{si:evaluation_metrics}.}
{
\setlength\tabcolsep{1.15pt}
\label{table:bias_and_wass_train_steps}
\begin{tabular}{|lcccc|cccc|cccc|}
\hline
\multirow{3}{*}{Variable} & \multicolumn{4}{c|}{Mean}
 & \multicolumn{4}{c|}{Mean}
 & \multicolumn{4}{c|}{Mean}\\
 & \multicolumn{4}{c|}{Absolute Bias $\downarrow$} 
 & \multicolumn{4}{c|}{Wasserstein Distance $\downarrow$}
 & \multicolumn{4}{c|}{Absolute Error, $99^{\text{th}}$ $\downarrow$}  \\
                          & 300k    & 500k  & 1M  & 2M          & 300k  & 500k  & 1M     & 2M       & 300k & 500k  & 1M & 2M\\ \hline
Temperature (K)           & 0.41	& 0.41	& 0.43	& \textbf{0.4}	  & \textbf{0.47}	& 0.47	& 0.5	& 0.48	     & \textbf{0.61}	& 0.67	& 0.7	& 0.67 \\
Wind speed (m/s)          & 0.19	& 0.18	& 0.17	& \textbf{0.16}	  & 0.22	& 0.22	& \textbf{0.21}	& 0.2	     & 0.48	& \textbf{0.47}	& 0.47	& 0.52 \\
Specific humidity (g/kg)  & \textbf{0.31}	& 0.33	& 0.36	& 0.36	  & \textbf{0.36}	& 0.38	& 0.4	& 0.42	     & \textbf{0.45}	& 0.47	& 0.5	& 0.5 \\
Sea-level pressure (Pa)   & 39.92	& 37.07	& \textbf{26.6}	& 36.74	  & 52.09	& 50.83	& \textbf{43.63}	& 51.67	     & \textbf{77.99}	& 87.53	& 130.13	& 112.24 \\
Relative humidity (\%)    & \textbf{1.71}	& 1.76	& 1.79	& 1.79	  & 2.1	    & 2.12	& 2.12	& \textbf{2.09}	     & 1.87	& 1.85	& 1.78	& \textbf{1.75} \\
Heat index (K)            & \textbf{0.47}	& 0.47	& 0.51	& 0.47	  & \textbf{0.53}	& 0.54	& 0.57	& 0.56	     & \textbf{0.68}	& 0.75	& 0.77	& 0.82 \\ \hline
\end{tabular}
}
\end{table}

\begin{figure}[tbh!]
\centering
\includegraphics[width=0.95\textwidth]{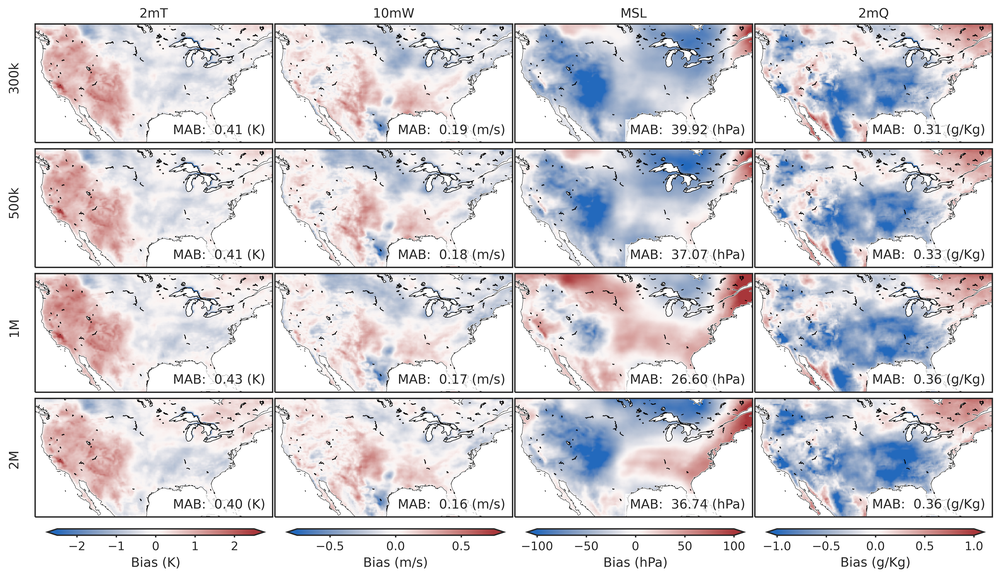}
\caption{\rev{\textbf{Spatial distribution of biases.}} 
Spatial distribution of statistical biases of the downscaled variables over CONUS during the summer (June-August) for the evaluation period 2010-2019 for \ourname trained with different number of steps. 
}\label{fig:si_ablation_conus_bias_num_steps}
\end{figure}

Here, we evaluate changes in the performance of \ourname with longer training times. Table \ref{table:bias_and_wass_train_steps} shows marginal improvements in the statistics of some downscaled fields over CONUS with increased training time, with the exception of the sea-level pressure, which benefits from longer training. At one million training steps we observe that some metrics start to deteriorate for some fields. Fig.~\ref{fig:si_ablation_conus_bias_num_steps} depicts the changes in the geographical distribution of biases with training time. We can observe that increasing the number of training steps does not change the distribution significantly, besides the sea-level pressure.

\begin{figure}[tbh!]
\centering
\includegraphics[width=0.75\textwidth]{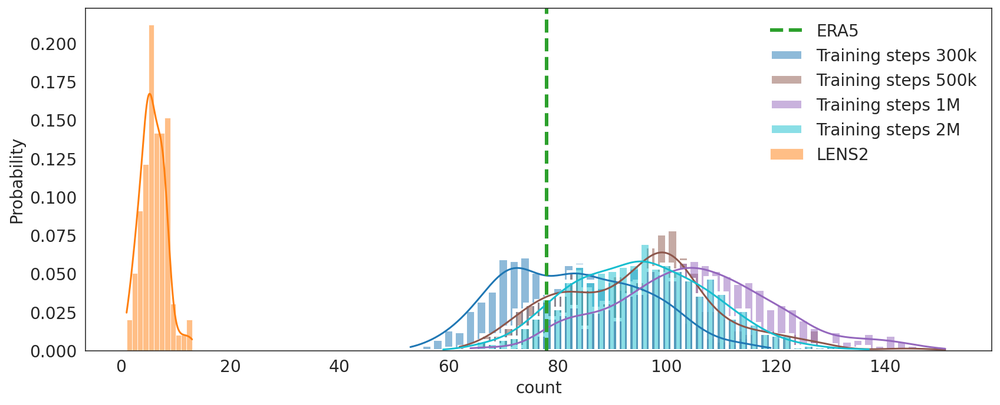}
\caption{\textbf{Distribution of TC counts.} Distribution of the number of TCs detected by TempestExtremes in the North Atlantic for the hurricane season August-September-October during 2010-2019 for \ourname trained with different number of steps. 
}\label{fig:si_ablation_tc_counts_num_steps}
\end{figure}

\begin{figure}[tbh!]
\centering
\includegraphics[width=0.99\textwidth]{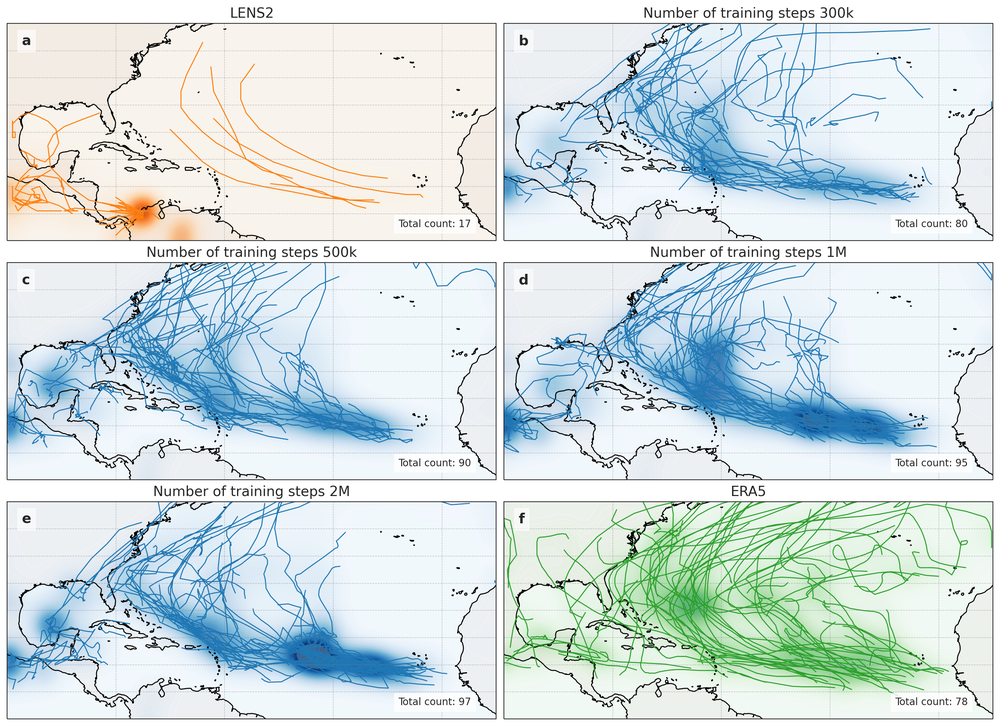}
\caption{\textbf{TC tracks and density.} TC tracks and their density for a LENS2 member in the North Atlantic in the time period 2010-2020 (\textbf{a}), one of downscaling samples from the same member generated using \ourname trained with trained with different number of steps (\textbf{b}-\textbf{e}), tracks detected using the reference ERA5 data (\textbf{f}).
}\label{fig:si_ablation_tc_tracks_num_steps}
\end{figure}

In contrast, longer training times deteriorate the ability of \ourname to represent TCs, as shown in Fig.~\ref{fig:si_ablation_tc_counts_num_steps} and Fig.~\ref{fig:si_ablation_tc_tracks_num_steps}. \ourname models trained for more than 300k steps tend to overestimate the frequency of tropical cyclones and reduce their track variability.

\clearpage
\subsection{Number of ensemble members}
\label{si:sec:ensemble_members}

\begin{table}[t!]
\centering
\footnotesize
\caption{Indices of the different LENS2 members used for training.}
{
\setlength\tabcolsep{2pt}
\label{table:indices_lens2_members}
\begin{tabular}{|c|c|c|c|}
\hline
1 member & 2 members & 4 members & 8 members \\ \hline
\texttt{cmip6\_1001\_001} & \texttt{cmip6\_1001\_001} & \texttt{cmip6\_1001\_001} & \texttt{cmip6\_1001\_001} \\
& \texttt{cmip6\_1251\_001} & \texttt{cmip6\_1251\_001} & \texttt{cmip6\_1251\_001} \\
& & \texttt{cmip6\_1301\_010} & \texttt{cmip6\_1301\_010} \\
& & \texttt{smbb 1301 020} & \texttt{smbb 1301 020} \\
& & &  \texttt{smbb\_1011\_001}\\
& & &  \texttt{smbb\_1301\_011}\\
& & &  \texttt{cmip6\_1281\_001}\\
& & &  \texttt{cmip6\_1301\_003},\\
\hline
\end{tabular}
}
\end{table}

Here we considering training the debiasing stage using $1$, $2$, $4$ and $8$ LENS ensemble members, with indices shown in Table \ref{table:indices_lens2_members}.

The impact of this change is summarized in Table \ref{table:bias_and_wass_ensemble_members}. Using more than 1 ensemble member generally improves performance. However, there is no clear improvement trend beyond using 4 members.

\begin{figure}[tbh!]
\centering
\includegraphics[width=0.95\textwidth]{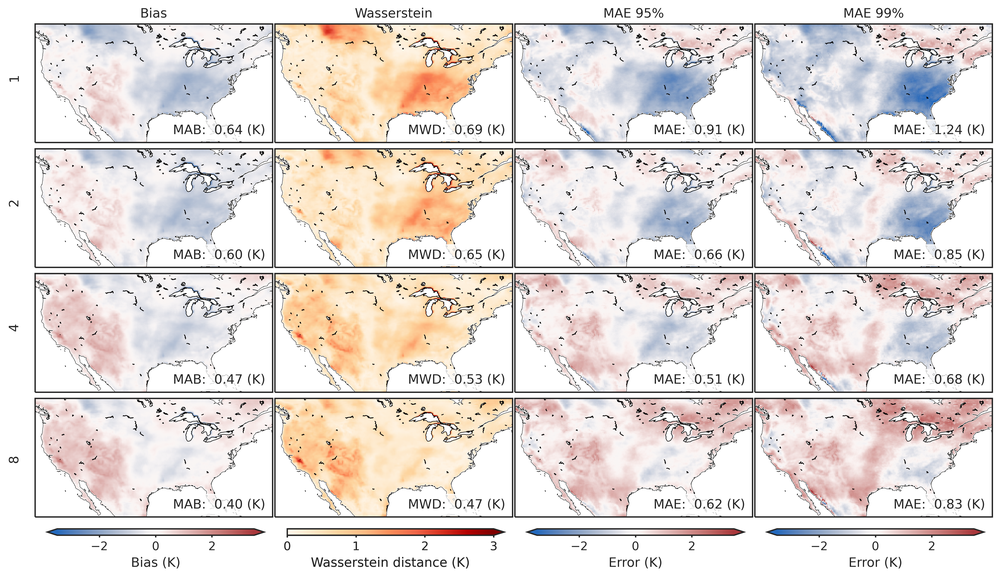}
\caption{\textbf{Metrics of heat index.} Metrics of the derived variable heat index over CONUS during the summer (June-August) for the evaluation period 2010-2019 for \ourname trained with different number of LENS2 ensemble members. 
}\label{fig:si_ablation_conus_hi_num_members}
\end{figure}

Fig.~\ref{fig:si_ablation_conus_hi_num_members} shows the impact of the number of LENS ensemble members used during training on the heat index statistics. Using more members decreases the errors on average, up to 4 ensemble members. However, this behavior is not uniform, as the Wasserstein error increases in the Rockies and in the Sierra Nevada, whereas it is reduced in the East Coast.

\begin{figure}[tbh!]
\centering
\includegraphics[width=0.95\textwidth]{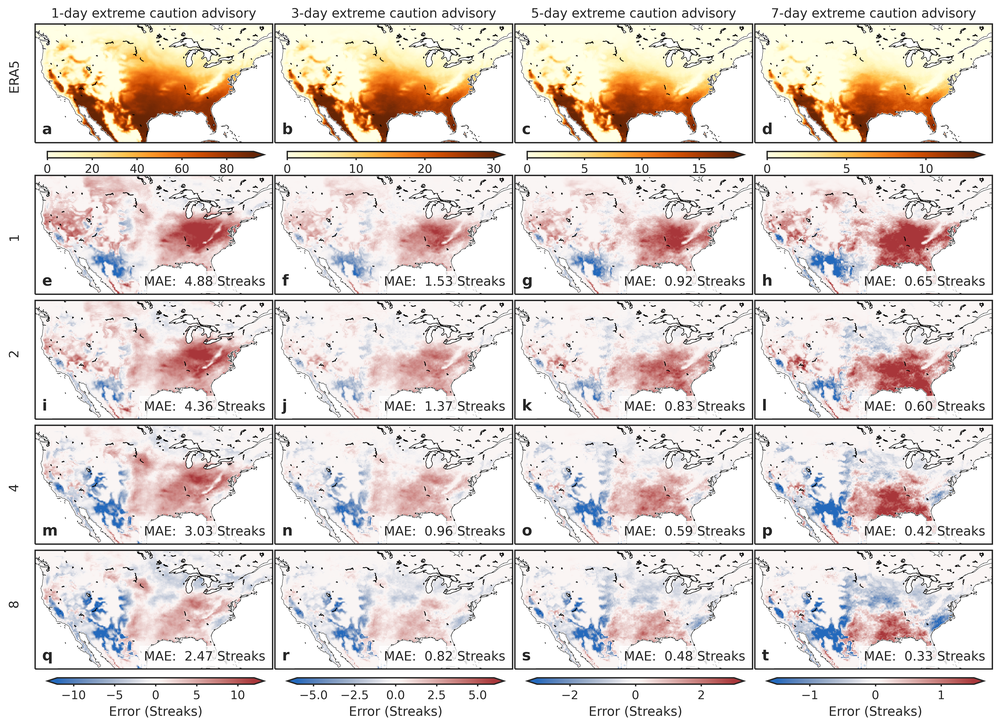}
\caption{\textbf{Bias in extreme caution advisories.} Spatial distribution of mean absolute errors in the number of extreme caution advisory streaks of different lengths per year. We show the ground truth (ERA5)(\textbf{a-d}), and the pointwise errors of \ourname trained with different number of LENS2 members.
}\label{fig:si_conus_heatwaves_extreme_caution_ablation_num_members}
\end{figure}

\rev{Fig.~\ref{fig:si_conus_heatwaves_extreme_caution_ablation_num_members} shows that leveraging multiple ensemble members is critical for the accurate prediction of rare extremes. Increasing the number of ensemble members from 1 to 8 decreases the error in extreme caution advisories by roughly half for periods ranging from 1 to 7 days.}

\begin{table}[t!]
\centering
\footnotesize
\caption{
Effect of the number of LENS2 members used during training on mean absolute bias, mean Wasserstein distance, and mean absolute error in the $99^{\text{th}}$ percentile for different variables. The precise definitions of the metrics are included in~\ref{si:evaluation_metrics}.}
{
\setlength\tabcolsep{1.2pt}
\label{table:bias_and_wass_ensemble_members}
\begin{tabular}{|lcccc|cccc|cccc|}
\hline
\multirow{2}{*}{Variable} & \multicolumn{4}{c|}{Mean}
 & \multicolumn{4}{c|}{Mean}
 & \multicolumn{4}{c|}{Mean}\\
 & \multicolumn{4}{c|}{Absolute Bias $\downarrow$} 
 & \multicolumn{4}{c|}{Wasserstein Distance $\downarrow$}
 & \multicolumn{4}{c|}{Absolute Error, $99^{\text{th}}$ $\downarrow$}  \\
                          & 1       & 2     & 4     & 8                  & 1     & 2  & 4  & 8                          & 1    & 2  & 4  & 8\\ \hline
Temperature (K)           & 0.5     & 0.47  & 0.41  & \textbf{0.39}      & 0.55  & 0.52  & 0.47  & \textbf{0.45}      & 0.84  &	0.61 &	\textbf{0.61} &	0.75 \\
Wind speed (m/s)          & \textbf{0.15}    & 0.17  & 0.19  & 0.17      & \textbf{0.17}  & 0.19  & 0.22  & 0.22      & \textbf{0.38}  &	0.42 &	0.48 &	0.52 \\
Specific humidity (g/kg)  & 0.37    & 0.36  & 0.31  & \textbf{0.27}      & 0.4   & 0.4   & 0.36  & \textbf{0.35}      & 0.57  &	\textbf{0.4} &	0.45 &	0.62 \\
Sea-level pressure (Pa)   & 41.25   & \textbf{33.99} & 39.92 & 60.55     & 46.34 & \textbf{42.14} & 52.09 & 72.93     & \textbf{62.75} &	63.35 &	77.99 &	84.25 \\
Relative humidity (\%)    & 1.78    & \textbf{1.55}  & 1.71  & 1.76      & 2.06  & \textbf{1.88}  & 2.1   & 2.16      & 2.75  &	\textbf{1.82} &	1.87 &	1.96 \\
Heat index (K)            & 0.64    & 0.6   & 0.47  & \textbf{0.4}       & 0.69  & 0.65  & 0.53  & \textbf{0.47}      & 1.24  &	0.85 &	\textbf{0.68} &	0.83 \\ \hline
\end{tabular}
}
\end{table}

\subsection{\rev{Training data coupling in the debiasing stage}} \label{si:ablations_coupling}

\rev{The choice of coupling $\pi \in \Pi(\mu_y^i, \mu_y')$ in \eqref{eq:app_reflow_loss} is critical and primarily driven by the need to align climatologies. Specifically, samples from each dataset must be paired with consistent climatological statistics to effectively compute the unbiased anomalies transported by the flow in \eqref{eq:reflow_ode}.}

\rev{In this section, we ablate the choice of training data coupling. In the baseline \ourname model, the coupling enforces strict temporal alignment to ensure that both the climatology and the specific year remain consistent. This is achieved via a map based on timestamps: we sample a date within the training period and extract the corresponding samples from both datasets (sampling across different members for the LENS2 dataset). These samples are then normalized using their respective day-of-the-year climatologies. As explained in Section \ref{si:reflow_cost}, the rectified flow model is trained to map the resulting anomalies from the climate simulation to the coarse-grained reanalysis.}

\rev{For the ablation study, we relax the coupling by allowing the timestamps of the paired samples to differ by a defined threshold. We consider three variations:
\begin{itemize}
    \item \textbf{Day shift ($\pm d$d):} The year matches, but the day-of-the-year may differ by up to $d$-number of days. In this regime, the climatologies remain relatively similar.
    \item \textbf{Year shift ($\pm y$y):} The day-of-the-year matches, but the years may differ by up to $y$-number of years.
    \item \textbf{Random coupling:} We use the tensor product of the marginals, $\pi = \mu_y^i \otimes \mu_y'$, effectively ignoring timestamps entirely.
\end{itemize}
To ensure a fair comparison, all ablation experiments use the \textit{exact} same configuration, including the learning rate schedule, number of iterations, and random seed for weight initialization.}

\rev{Table \ref{table:ablation_coupling} summarizes the evaluation of models trained with different couplings over CONUS during the summers (June--August) of the testing period, highlighting several important trends. When the year is fixed, errors increase as the daily climatologies become misaligned. Conversely, when the day-of-the-year is aligned but the years are allowed to shift, small shifts (e.g., $\pm 1$y) can slightly improve performance, likely due to the increased diversity of training samples. However, performance regresses sharply as the year window widens further. This degradation stems from the loss of year-scale alignment; because the model is trained on only 20 years of a slowly evolving distribution and must extrapolate accurately to the future, it is imperative that the network captures this time-dependent signal. Widening the coupling window blurs these subtle distributional shifts. Finally, using a random coupling (ignoring timestamps entirely) results in a sharp degradation across all metrics.}

\begin{figure}[tbh!]
\centering
\includegraphics[width=0.95\textwidth]{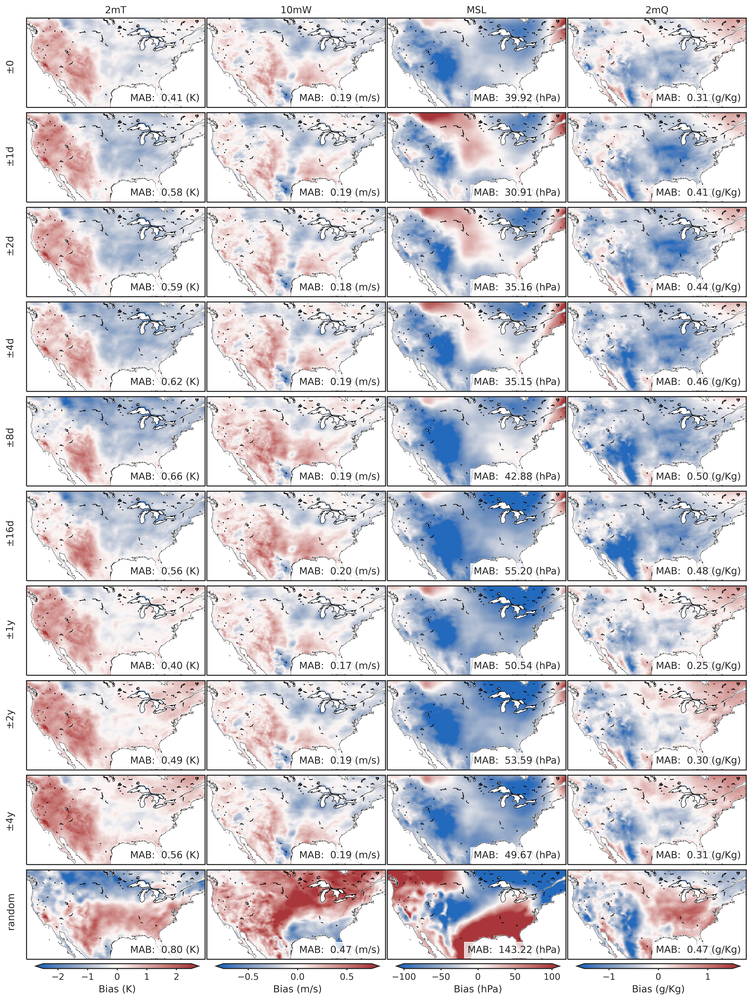}
\caption{\textbf{Spatial distribution of biases.} \rev{Spatial distribution of statistical biases of the downscaled variables over CONUS during the summer (June-August) for the evaluation period 2010-2019 for \ourname with different training data couplings.}
}\label{fig:si_conus_pixel_err_ablation_coupling}
\end{figure}

\begin{figure}[tbh!]
\centering
\includegraphics[width=0.95\textwidth]{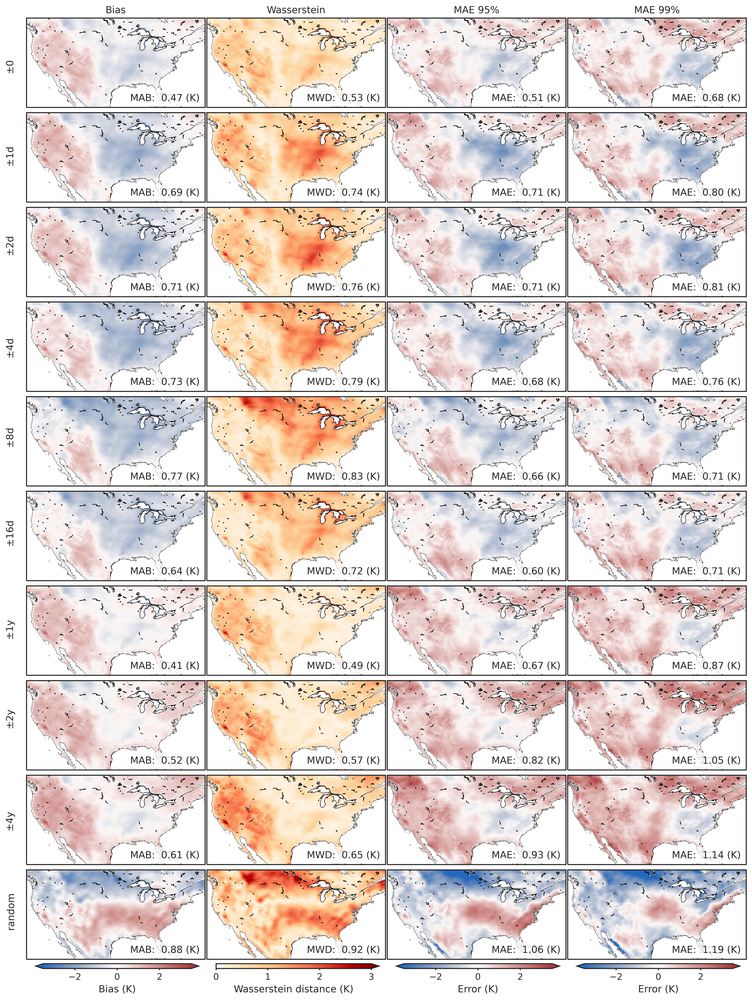}
\caption{\textbf{Spatial distribution of different errors for the heat index.} \rev{Spatial distribution of different errors for the downscaled heat index compound variables over CONUS during the summer (June-August) for the evaluation period 2010-2019 for \ourname with different training data couplings.}
}\label{fig:si_conus_hi_err_ablation_coupling}
\end{figure}

\rev{Fig.~\ref{fig:si_conus_pixel_err_ablation_coupling} illustrates the biases over CONUS during the summer months of 2010--2019. Increasing the misalignment in days does not drastically alter the geographical distribution of errors, with the exception of the mean sea-level pressure, which improves with small differences but deteriorates for differences beyond a week. When maintaining the same day-of-the-year but varying the years, the spatial pattern of biases remains roughly constant, though their magnitude fluctuates: decreasing for a one-year difference before rapidly increasing. In contrast, the random coupling yields a distinct spatial error distribution with significantly larger biases. A similar trend is observed in Fig. \ref{fig:si_conus_hi_err_ablation_coupling} for the heat index metrics, where the geographical distribution of errors shifts smoothly as the temporal difference (in days or years) increases. As with the direct variables, the errors for the random coupling are significantly larger.}

\begin{figure}[tbh!]
\centering
\includegraphics[width=0.95\textwidth]{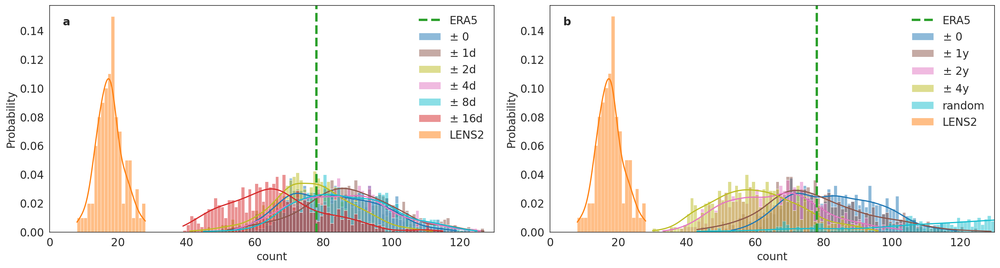}
\caption{\textbf{Distribution of TC counts for different training couplings.} \rev{Distribution of the number of TCs detected in the North Atlantic for the hurricane season August-October during 2010-2019 for \ourname with different training data couplings.} 
}\label{fig:si_ablation_tc_counts_couplings}
\end{figure}

\begin{figure}[tbh!]
\centering
\includegraphics[width=0.75\textwidth]{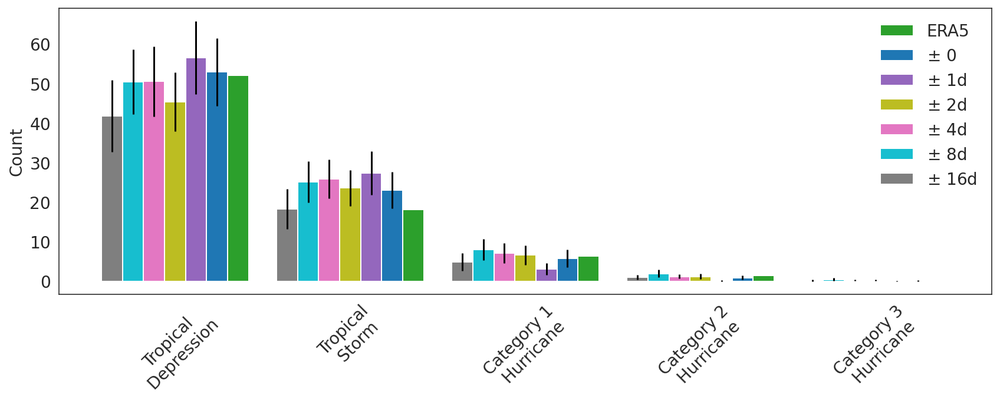}
\caption{\textbf{Saffir-Simpson scale distribution of TCs.} \rev{Distribution of the intensity of TCs detected in the North Atlantic during the peak hurricane season (August–October), 2010–2019. Results shown for GenFocal models trained with different couplings, for small differences (in days) of the time stamps. Error bars denote the ensemble standard deviation.}
}\label{fig:si_ablation_tc_categories_coupling_day}
\end{figure}

\begin{figure}[tbh!]
\centering
\includegraphics[width=0.75\textwidth]{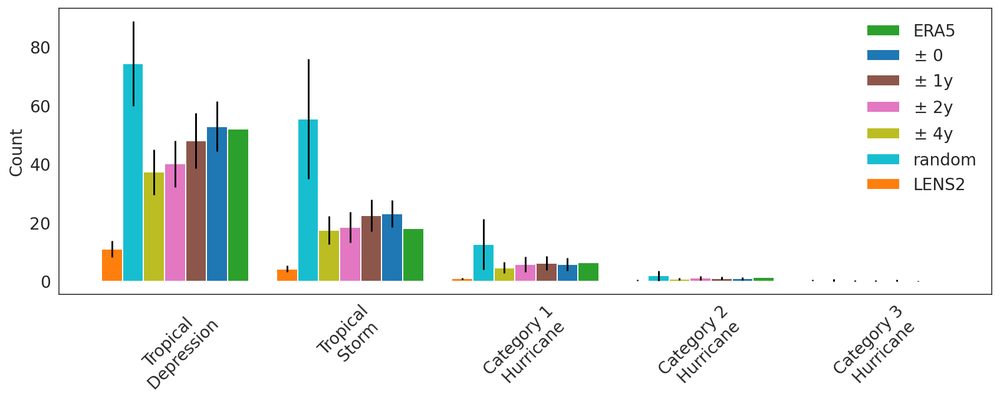}
\caption{\textbf{Saffir-Simpson scale distribution of TCs.} \rev{Distribution of the intensity of TCs detected in the North Atlantic during the peak hurricane season (August–October), 2010–2019. Results shown for GenFocal models trained with different couplings, in which the day-of-the-year remains fixed and we sample from similar years. Error bars denote the ensemble standard deviation.}
}\label{fig:si_ablation_tc_categories_coupling_years}
\end{figure}

\rev{A similar trend is observed for tropical cyclones (TCs), as shown in Fig.~\ref{fig:si_ablation_tc_counts_couplings}. Increasing the maximum day shift preserves the distribution for small shifts; however, beyond a one-week threshold, the models begin to underestimate the annual TC count. Conversely, widening the year window results in a much more rapid decline in TC frequency. In contrast, the fully random coupling severely overestimates the number of TCs. This is further corroborated by the Saffir-Simpson intensity distributions shown in Fig.~\ref{fig:si_ablation_tc_categories_coupling_day}, where the response to increasing day shifts is relatively smooth, whereas year shifts induce a much sharper transition (Fig.~\ref{fig:si_ablation_tc_categories_coupling_years}). Finally, the completely random coupling, given by the tensor product of the marginal measures, exhibits markedly anomalous behavior, consistent with its poor performance on other metrics.}

\begin{table}[th!]
\centering
\footnotesize
\caption{\rev{Effect of training data coupling on the mean absolute bias, mean Wasserstein distance, and mean absolute error of the $99^{\text{th}}$ percentile. Results are shown for various coupling strategies applied to the CONUS summer seasons (June–August) from 2010 to 2019. Metric definitions are provided in section~\ref{si:evaluation_metrics}.}}
{
\setlength\tabcolsep{2.25pt}
\label{table:ablation_coupling}
\begin{tabular}{|lcccccccccc|}
\hline
\multirow{2}{*}{Variable} & $\pm 0$     & $\pm 1$d   & $\pm 2$d  & $\pm 4$d   & $\pm 8$d  & $\pm 16$d  & $\pm 1$y & $\pm 2$y & $\pm 4$y & random    \\ 
\multirow{2}{*}{}         &  \multicolumn{10}{c|}{Mean Absolute Bias, $\downarrow$}          \\
Temperature (K)           & 0.41 &	0.58 &	0.59 &	0.62 &	0.66 &	0.56 &	\textbf{0.4} &	0.49 &	0.56 &	0.8 \\
Wind speed (m/s)          & 0.19 &	0.19 &	0.18 &	0.19 &	0.19 &	0.2 &	\textbf{0.17} &	0.19 &	0.19 &	0.47 \\
Specific humidity (g/kg)  & 0.31 &	0.41 &	0.44 &	0.46 &	0.5 &	0.48 &	\textbf{0.25} &	0.30 &	0.31 &	0.47 \\
Sea-level pressure (Pa)   & 39.92 &	\textbf{30.91} &	35.16 &	35.15 &	42.88 &	55.2 &	50.54 &	53.59 &	49.67 &	143.22 \\
Relative humidity (\%)    & \textbf{1.71} &	1.73 &	1.8 &	1.8 &	1.88 &	1.92 &	1.77 &	1.97 &	2.00 & 	2.82 \\
Heat index (K)            & 0.47 &	0.69 &	0.71 &	0.73 &	0.77 &	0.64 &	\textbf{0.41} &	0.52 &	0.61 &	0.88 \\ \hline
\multirow{2}{*}{}         & \multicolumn{10}{c|}{Mean Wasserstein Distance, $\downarrow$}          \\            
Temperature (K)           & 0.47 & 	0.63 & 	0.64 & 	0.68 & 	0.72 & 	0.63 & 	\textbf{0.46} & 	0.54 & 	0.6 & 	0.83 \\
Wind speed (m/s)          & 0.22 & 	0.23 & 	0.22 & 	0.23 & 	0.24 & 	0.25 & 	\textbf{0.21} & 	0.23 & 	0.24 & 	0.5 \\
Specific humidity (g/kg)  & 0.36 & 	0.47 & 	0.49 & 	0.52 & 	0.56 & 	0.55 & 	\textbf{0.34} & 	0.38 & 	0.39 & 	0.51 \\
Sea-level pressure (Pa)   & 52.09 & \textbf{49.53} & 51.22 & 55.44 & 65.59 & 71.86 & 61.41 & 67.08 & 60.36 & 149.25 \\
Relative humidity (\%)    & \textbf{2.1} & 	2.17 & 	2.25 & 	2.28 & 	2.37 & 	2.41 & 	2.19 & 	2.37 & 	2.38 & 	3.2 \\
Heat index (K)            & 0.53 & 	0.74 & 	0.76 & 	0.79 & 	0.83 & 	0.72 & 	\textbf{0.49} & 	0.57 & 	0.65 & 	0.92 \\ \hline
\multirow{2}{*}{} & \multicolumn{10}{c|}{Mean Absolute Error, $99^{\text{th}}$ $\downarrow$}         \\
Temperature (K)           & \textbf{0.61} &	0.71 &	0.7 &	0.68 &	0.66 &	0.68 &	0.8 &	0.95 &	1.02 &	1.03 \\
Wind speed (m/s)          & 0.48 &	0.5 &	0.47 &	0.53 &	0.60 &	0.62 &	\textbf{0.45} &	0.48 &	0.49 &	1.22 \\
Specific humidity (g/kg)  & 0.45 &	0.47 &	0.46 &	0.45 &	\textbf{0.44} &	0.46 &	0.6 &	0.68 &	0.7 &	0.56 \\
Sea-level pressure (Pa)   & 77.99 &	99.98 &	94.35 &	105.06 & 117.39 & 97.2 &	\textbf{75.73} &	83.54 &	75.96 &	103.49 \\
Relative humidity (\%)    & \textbf{1.87} &	1.94 &	1.96 &	2.07 &	2.22 &	2.21 &	1.97 &	1.99 &	1.98 &	5.31 \\
Heat index (K)            & \textbf{0.68} &	0.8 &	0.81 &	0.76 &	0.71 &	0.71 &	0.87 &	1.05 &	1.14 &	1.19 \\ \hline
\end{tabular}
}
\end{table}

\subsection{Training period for the super-resolution stage} \label{si:ablations_years_sr}

\begin{figure}[tbh!]
    \centering
    \includegraphics[width=\linewidth]{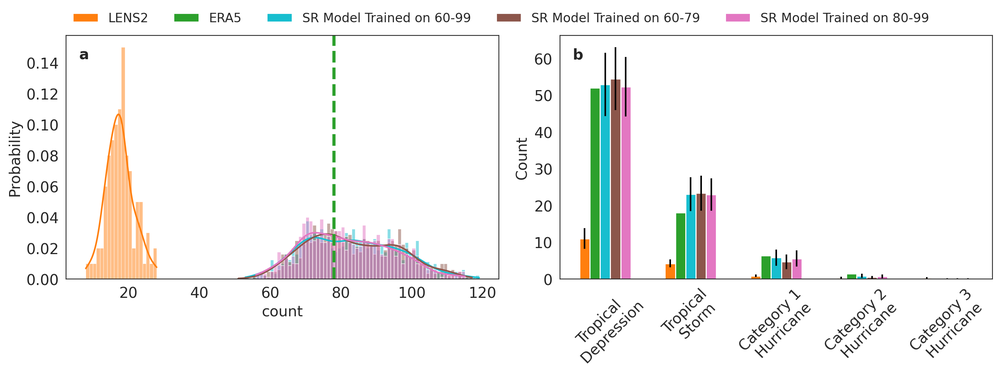}
    \caption{\rev{\textbf{TC statistics for different super-resolution model training ranges.} The same debiasing model (trained on 1980-1999) is shared between all \ourname variants.}}
    \label{fig:sr_training_range_ablation}
\end{figure}

\rev{
We observe that the performance of \ourname is insensitive to the training period of the super-resolution component. This contrasts with the debiasing step, as illustrated in Fig.~\ref{fig:sr_training_range_ablation}, which displays the TC statistics (including raw and categorized counts). Similar trends were observed across all other evaluation metrics. These results provide empirical evidence that the \ourname framework effectively decomposes the problem into stationary (super-resolution) and non-stationary components.
}

\subsection{Temporally coherent denoising in super-resolution} 
\label{si:super_resolution_temporal_length}

\rev{
The utility of the time-coherent sampling technique employed in the super-resolution stage (section~\ref{si:multi-diffusion}) is directly reflected in the temporal spectra. As shown in Table~\ref{table:stitching_tse}, this technique leads to lower spectral errors for all variables, especially for temperature and pressure.
}

\begin{table}[t!]
\caption{
\rev{\textbf{Impact of time-coherent sampling (section~~\ref{si:multi-diffusion}) on the mean temporal spectra error (TSE).} Columns represent variants with the super-resolution step trained on 7-day or 3-day samples, and time-coherent sampling enabled (+) or not (-) during inference.}
}
\label{table:stitching_tse}
\begin{tabular}{|l|ccc|ccc|}
\hline
            & 7-day + & 7-day - & \% change & 3-day + & 3-day - & \% change \\ \hline
Temperature & 0.746   & 0.780   & -4.3      & 0.728   & 0.788   & -7.6      \\
Wind speed  & 0.416   & 0.417   & -0.2      & 0.393   & 0.397   & -1.0      \\
Humidity    & 0.536   & 0.556   & -3.6      & 0.533   & 0.572   & -6.8      \\
Pressure    & 0.513   & 0.566   & -9.4      & 0.510   & 0.614   & -16.9     \\ \hline
\end{tabular}
\end{table}

\begin{figure}[tbh!]
\centering
\includegraphics[width=0.8\textwidth]{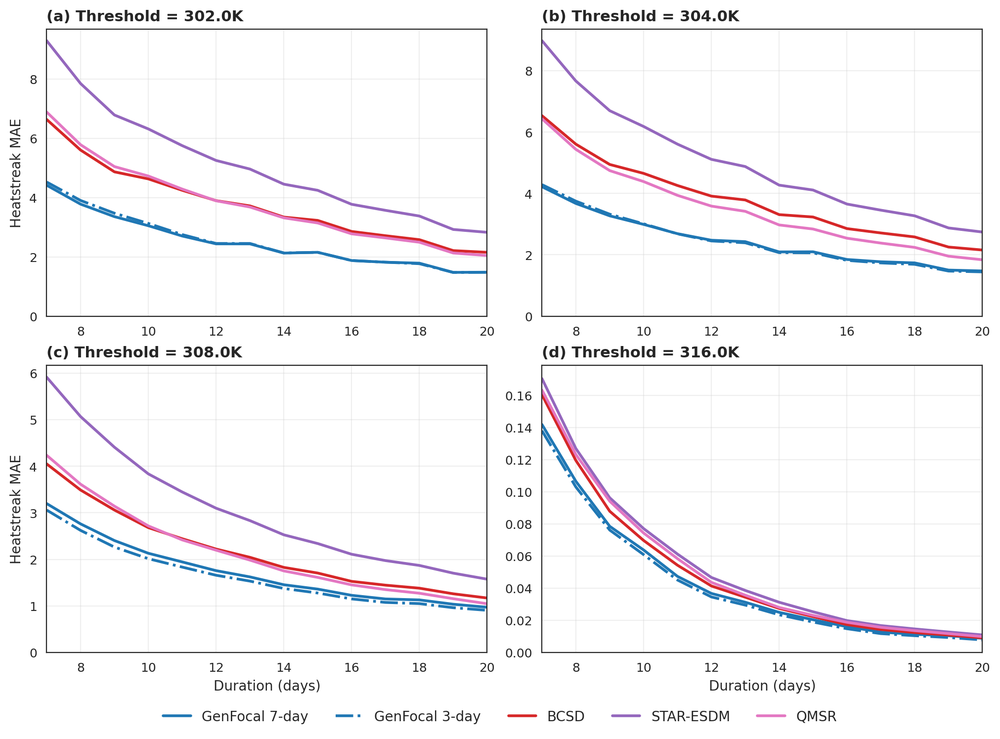}
\caption{
\rev{
\textbf{Spatially averaged bias in the number of heat streaks vs. durations for selected thresholds.} \ourname 7-day and 3-day represent variants whose super-resolution step is trained on samples of different lengths, with time-coherent sampling (1-day overlap) applied at inference time.
}
}
\label{fig:si_conus_extended_heat}
\end{figure}

\rev{
This technique allows \ourname to better capture statistics of events with durations more than the training sample length. 
Fig.~\ref{fig:si_conus_extended_heat} presents the spatially averaged bias for streak counts over 1 to 3 weeks.
We observe that across a wide range of thresholds and durations, \ourname has better statistics estimation compared to BCSD and STAR-ESDM. Furthermore, the results remain insensitive to the duration of the training samples for the super-resolution step (shown for 3-day and 7-day variants), demonstrating the robustness of our time-coherent sampling technique and the scalability of \ourname in capturing the statistics of persistent, long-lasting events.
}

\begin{figure}[tbh!]
\centering
\includegraphics[width=0.95\textwidth]{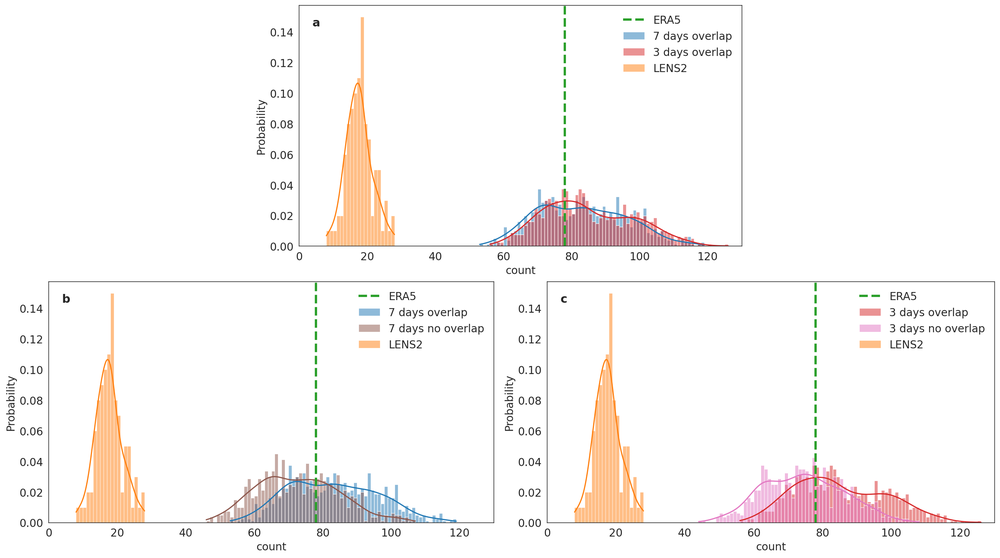}
\caption{\textbf{Distribution of North Atlantic TC counts for super-resolution models with varying training sample lengths and time-coherent sampling.} \rev{
a. super-resolution models trained with \emph{7-day vs. 3-day samples}; sampled with time coherence. b. super-resolution model trained with 7-day samples, \emph{with vs. without time-coherent sampling}. c. super-resolution model trained with 3-day samples, \emph{with vs. without time-coherent sampling}.
} 
}\label{fig:si_ablation_tc_counts_stitiching}
\end{figure}

\rev{
Fig.~\ref{fig:si_ablation_tc_counts_stitiching} demonstrates that time-coherent sampling results in improved TC statistics. Notably, it leads similar distributions regardless of window size (Fig.~\ref{fig:si_ablation_tc_counts_stitiching}\textbf{a}).
Figs.~\ref{fig:si_ablation_tc_counts_stitiching}\textbf{b} and \ref{fig:si_ablation_tc_counts_stitiching}\textbf{c} reveal a tendency to underestimate TCs when the technique is absent. 
}

\subsection{Residual modeling in super-resolution}
\label{si:residual_modeling}

\rev{
Table~\ref{table:pixel_error_residual} presents the ablation between the main \ourname model against one trained to directly predict the output, i.e. without applying the residual formulation (reference equation). We observe that modeling the residual between high-resolution target and the interpolated low-resolution input leads to significant improvements in the distribution of temperature-related variables, as well as the extreme percentiles for all variables except pressure.
}

\begin{table}[t!]
\centering
\footnotesize
\caption{
\rev{
Effect residual modeling on mean absolute bias, mean Wasserstein distance, and mean absolute error in the $99^{\text{th}}$ percentile for different variables. The precise definitions of the metrics are included in \ref{si:evaluation_metrics}.
}
}
{
\setlength\tabcolsep{3pt}
\label{table:pixel_error_residual}
\begin{tabular}{|lcc|cc|cc|}
\hline
\multirow{2}{*}{Variable} & \multicolumn{2}{c|}{Mean}
 & \multicolumn{2}{c|}{Mean}
 & \multicolumn{2}{c|}{Mean}\\
 & \multicolumn{2}{c|}{Absolute Bias $\downarrow$} 
 & \multicolumn{2}{c|}{Wasserstein Distance $\downarrow$}
 & \multicolumn{2}{c|}{Absolute Error, $99^{\text{th}}$ $\downarrow$}  \\
                          & residual & direct & residual  & direct   & residual & direct \\ \hline
Temperature (K)           & \textbf{0.41}	& 0.73	  & \textbf{0.47}	& 0.75	     & \textbf{0.61}	& 1.16 \\
Wind speed (m/s)          & 0.19	& 0.19	  & 0.22	& 0.22	     & \textbf{0.48}	& 0.52 \\
Specific humidity (g/kg)  & 0.31	& \textbf{0.29}	  & \textbf{0.36}	& 0.37	     & \textbf{0.45}	& 0.75 \\
Sea-level pressure (Pa)   & \textbf{39.92}	& 46.61	  & \textbf{52.09}	& 56.96	     & 77.99& \textbf{73.96} \\
Relative humidity (\%)    & \textbf{1.71}	& 1.98	  & \textbf{2.10}    & 2.16	     & \textbf{1.87}	& 2.08 \\
Heat index (K)            & \textbf{0.47}	& 0.79	  & \textbf{0.53}	& 0.82	     & \textbf{0.68}	& 1.37	\\ \hline
\end{tabular}
}
\end{table}

\clearpage
\section{Additional studies} \label{si:additional_studies}

\rev{We include additional studies for environmental risks during the Northern Hemisphere winter (December, January and February) over the evaluation period. During these months, the combination of cold temperatures and strong winds can pose human safety hazards, such as hypothermia and frostbite \cite{wind_chill}.  Persistent freezing events can cause significant damage to agricultural crops and infrastructure.}

\rev{To this end, we evaluate the ability of \ourname to capture winter extremes by analyzing the tail dependence of high near-surface winds and low near-surface temperatures at a fixed time of day (12Z). We also assess the statistics of multi-day streaks of freezing daily minimum temperatures and windchill temperatures as projected by \ourname, comparing them against the ERA5 reanalysis.}

\rev{Consistent with everywhere else, we perform the model selection from the same pool of models trained for studying heat streaks in Northern Hemisphere reported in the main text and section~\ref{si:conus}. To save compute, we did not perform full end-to-end model selection. Using the validation data in the winters from the period of 2000-2009, we chose the model with the lowest 2m temperature bias, after the debiasing step at the coarse-level, namely without going through the super-resolution stage. We then applied the model to the test data from the period of 2010-2019, including both debiasing and super-resolution. We report metrics not only on single variables but also on derived as well as compound ones.}

\subsection{Definition of Events}

\paragraph{Wind Chill Temperature} 
\rev{The wind chill temperature (WCT) is a derived meteorological index that models the rate of heat loss from exposed human skin under combined wind and temperature conditions, and adopted by the National Weather Service (NWS) and Environment Canada \cite{wind_chill}. The WCT is defined as}

\begin{equation}
    WCT = 13.12 + 0.6215 T_{air} - 11.37 V^{0.16} + 0.3965 T_{air} V^{0.16}
    \label{eq:wct}
\end{equation}
\rev{where $T_{air}$ is the air temperature in degrees Celsius ($^{\circ}\mathrm{C}$) and $V$ is the wind speed at \ 10{meter} elevation in kilometers per hour ($\mathrm{km/h}$). }

\rev{Because WCT depends on the joint distribution of temperature and wind speed, accurate modeling requires a method that captures their inter-variable correlations.}

\rev{The WCT is a critical physiological metric for public safety. It informs safety thresholds for outdoor operations, with $-27^{\circ}\mathrm{C}$ ($246.15\,\mathrm{K}$) being the limit below which frostbite can occur.}

\subsection{Pixel-wise statistics}

\rev{We first evaluate the capability of \ourname to reconstruct marginal distributions of key variables. Table \ref{table:winter_bias_and_wass} presents the statistical modeling errors for different variables during the winter months. Here \ourname outperforms the baselines in most of the metrics. While BCSD and STAR-ESDM perform adequately on simple variables like wind speed, \ourname demonstrates superior performance on derived variables like the wind chill temperature.}

\begin{table}[tbh!]
\footnotesize
\centering
\caption{\rev{Statistical modeling errors of directly and derived downscaled variables in marginal distributions for winters (December-January-February) in CONUS (2010-2019). \ourname consistently outperforms baselines in capturing the distribution shapes (Wasserstein) and extremes ($99^{\text{th}}$ percentile MAE). Best values are in bold font. The precise definitions of the metrics are included in \ref{si:evaluation_metrics}}}
\label{table:winter_bias_and_wass}
\setlength\tabcolsep{4pt}
\begin{tabular}{lccc}
\toprule
\multirow{2}{*}{Variable} \hspace{2cm}& \ourname & BCSD  & STAR-ESDM   \\ 
                          & \multicolumn{3}{c}{Mean Absolute Bias $\downarrow$}  \\ 
\midrule
Temperature (K)           & \textbf{0.44}   & 0.54  & 0.66  \\
Wind speed (m/s)          & 0.18            & \textbf{0.12}  & 0.15  \\
Specific humidity (g/kg)  & \textbf{0.15}   & 0.17  & 0.26  \\
Sea-level pressure (Pa)   & 135.22          & \textbf{84.2}  & 90.21 \\ 
Windchill temperature (K) & \textbf{0.53}   & 0.70  & 0.86  \\ 

\midrule
                          & \multicolumn{3}{c}{Mean Wasserstein Distance $\downarrow$} \\ 
\midrule
Temperature (K)           & \textbf{0.54}   & 0.66  & 0.74      \\
Wind speed (m/s)          & \textbf{0.20}   & 0.22  & 0.21      \\
Specific humidity (g/kg)  & \textbf{0.19}   & 0.21  & 0.29      \\
Sea-level pressure (Pa)   & 138.80          & \textbf{88.65} & 93.82 \\
Windchill temperature (K) & \textbf{0.66}   & 0.84  & 0.95  \\ 

\midrule
                          & \multicolumn{3}{c}{Mean Absolute Error, $99^{\text{th}}$ \% $\downarrow$} \\ 
\midrule
Temperature (K)           & \textbf{1.08}   & 1.16  & 1.19       \\
Wind speed (m/s)          & \textbf{0.36}   & 0.50  & 0.47       \\
Specific humidity (g/kg)  & 0.38            & \textbf{0.33}  & 0.42       \\
Sea-level pressure (Pa)   & 124.20          & \textbf{110.82} & 123.94     \\
Windchill temperature (K) & \textbf{1.33}   & 1.37  &  1.44 \\ 
\bottomrule
\end{tabular}
\end{table}

\subsection{Multi-day frostbite episodes}

\rev{Beyond instantaneous statistics, temporal coherence is vital for assessing prolonged risk. We define a ``frostbite episode'' as consecutive days where the wind chill temperature remains below the critical threshold of $-27^{\circ}\mathrm{C}$ ($246.15\,\mathrm{K}$). Capturing the risk of these episodes is essential for public health warnings.}

\rev{Table \ref{table:winter_frostbite_errors} summarizes the performance across three metrics: Mean Absolute Bias (accuracy of count), Mean Continuous Ranked Probability Score (CRPS, probabilistic accuracy), and the Spread-Skill Ratio (SSR). An SSR of 1.0 indicates perfect uncertainty calibration.}

\rev{The results show that \ourname significantly outperforms BCSD across all durations. Notably, as the threshold duration increases (from 1 to 7 days), \ourname maintains a high SSR (0.84 for 7 days) compared to the baseline (0.74). This suggests that \ourname is not only more accurate but also more reliable in quantifying the uncertainty of extreme, multi-day cold events.}

\begin{table}[tbh!]
\centering
\caption{\rev{Statistical errors for the number of multi-day frostbite episodes duration evaluation period (December-January-February, CONUS 2010-2019) using mean absolution bias, mean continuous ranked probability score (CRPS), and mean spread-skill ratio (SSR). Best values are in bold font.}}
\label{table:winter_frostbite_errors}
\setlength\tabcolsep{6pt}
\begin{tabular}{lccc}
\toprule
\multirow{2}{*}{Duration } \hspace{1cm} & \ourname  & BCSD  & STAR-ESDM   \\ 
                         & \multicolumn{3}{c}{Mean Absolute Bias $\downarrow$}  \\
\midrule
1 day         & \textbf{1.48}    & 1.89    & 1.91  \\
3 days      & \textbf{0.41}    & 0.51    & 0.52  \\
5 days      & \textbf{0.20}    & 0.26    & 0.26  \\
7 days      & \textbf{0.12}    & 0.15    & 0.15  \\ 

\midrule
           & \multicolumn{3}{c}{Mean CRPS $\downarrow$} \\ 
\midrule
1 day     & \textbf{0.71}    & 1.30    & 1.34      \\
3 days     & \textbf{0.21}    & 0.35    & 0.37      \\
5 days     & \textbf{0.11}    & 0.18    & 0.19      \\
7 days     & \textbf{0.06}    & 0.10    & 0.11 \\

\midrule
           & \multicolumn{3}{c}{Mean SSR ($\text{Ideal} \approx 1.0$) $\uparrow$} \\ 
\midrule
1  day    & \textbf{0.79}    & 0.68    & 0.67      \\
3  days    & \textbf{0.79}    & 0.70    & 0.69      \\
5  days    & \textbf{0.82}    & 0.74    & 0.73      \\
7  days    & \textbf{0.84}    & 0.74    & 0.73      \\
\bottomrule
\end{tabular}
\end{table}

\begin{figure}[tbh!]
\centering
\includegraphics[width=0.95\textwidth]{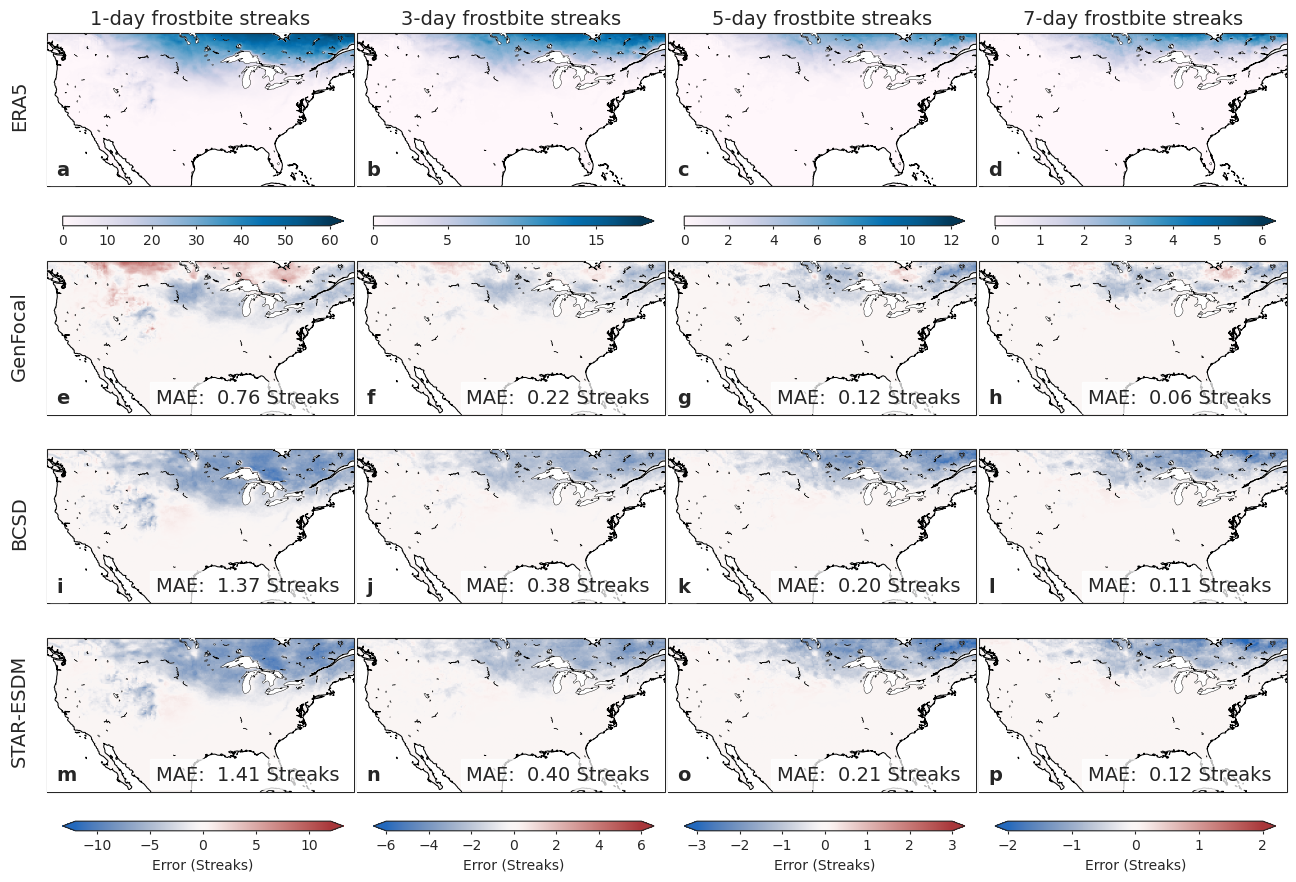}
\caption{\textbf{Frostbite streak statistics in CONUS winter (December-January-February) season, lasting [1, 3, 5, 7] days.} 
\rev{
a-d. ERA5 reanalysis. e-h. local bias of \ourname. i-l. local bias of BCSD. and m-p. local bias of STAR-ESDM.
} 
}\label{fig:si_conus_frostbite_streaks}
\end{figure}

\subsection{Tail Dependencies}

\begin{figure}[tbh!]
\centering
\includegraphics[width=0.95\textwidth]{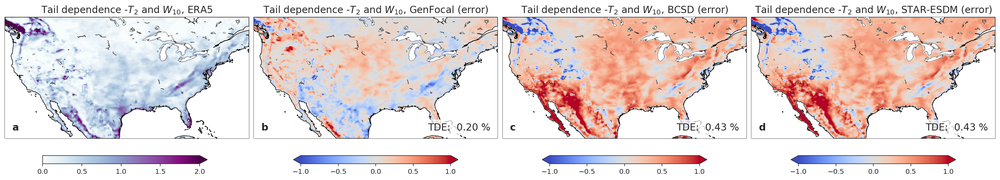}
\caption{\textbf{Tail dependence of low surface temperature and high wind speed in CONUS winter (December-January-February) season. }\rev{
a. ERA5 reanalysis. b. \ourname c. BCSD and d. STAR-ESDM.
} 
}\label{fig:si_conus_winter_tails}
\end{figure}

\rev{Fig.~\ref{fig:si_conus_winter_tails} shows that \ourname captures the co-occurrence of cold and windy extremes better than the BCSD and STAR-ESDM baselines. In particular, the improvement is most prominent across the Western and Southern United States, where the baselines tend to overestimate and \ourname reflects statistics much closer to that of the ERA5 reanalysis.}

\clearpage

\clearpage
\putbib
\end{bibunit}

\end{document}